\PassOptionsToPackage{table,x11names}{xcolor}
\documentclass{naturep}
\usepackage{graphicx}
\usepackage{pdfpages}
\usepackage{caption}
\usepackage[T1]{fontenc}
\usepackage[utf8]{inputenc}  
\usepackage{pifont}

\usepackage{mathptmx}
\usepackage{afterpage}
\usepackage{cuted} 
\usepackage{booktabs}
\usepackage{longtable}
\usepackage{amssymb}
\usepackage{amsmath}
\usepackage{graphicx}
\usepackage{algorithmicx}
\usepackage{algorithm}
\usepackage{adjustbox}
\usepackage{graphicx}
\usepackage{hyperref}
\usepackage{stfloats} 
\usepackage[noend]{algpseudocode}
\usepackage{bbm}
\usepackage{lineno}
\usepackage[margin=0.6in]{geometry}
\usepackage{ragged2e}
\usepackage{longtable}
\usepackage{xurl}     
\usepackage{hyperref} 
\usepackage{caption}
\hypersetup{breaklinks=true,hidelinks}
\usepackage{anyfontsize}
\usepackage{setspace}
\usepackage{float}
\usepackage{svg}
\usepackage{booktabs}
\usepackage{multirow}
\usepackage{blindtext}
\usepackage{tabularx}
\usepackage{caption}
\usepackage{subcaption}
\usepackage{enumitem, kantlipsum}
\usepackage{xspace}
\usepackage{pifont}

\usepackage{graphicx}
\usepackage{amsmath}
\usepackage{amssymb}
\usepackage{booktabs}
\usepackage{float}
\usepackage{placeins}
\usepackage{docmute}
\usepackage{xcolor}

\usepackage{makecell}
\usepackage{bm}
\usepackage{chngcntr}

\definecolor{maroon}{cmyk}{0,0.87,0.68,0.32}
\definecolor{gray}{rgb}{0.3,0.3,0.3}

\newcommand\Heading[1]{
  \noindent\textbf{\Large{#1}}
}

\newcommand\heading[1]{
  \noindent\textbf{\large{#1}}
}

\usepackage[capitalize,noabbrev]{cleveref}
\usepackage{appendix}
\usepackage{cleveref}

\crefname{figure}{Fig.}{Figs.}
\crefname{suppfigure}{Supplementary Fig.}{Supplementary Figs.}

\crefname{section}{Section}{Sections}
\crefname{theorem}{Theorem}{Theorems}
\crefname{lemma}{Lemma}{Lemmas}
\crefname{equation}{Equation}{Equations}
\crefname{proposition}{Proposition}{Propositions}
\crefname{claim}{Claim}{Claims}
\crefname{appendix}{Appendix}{Appendices}
\crefname{algorithm}{Algorithm}{Algorithms}
\crefname{figure}{Figure}{Figs}
\crefname{table}{Table}{Tables}
\crefname{remark}{Remark}{Remarks}
\crefname{definition}{Definition}{Definitions}
\crefname{equation}{Equation}{Equations}
\crefname{corollary}{Corollary}{Corollaries}
\crefname{section}{Method}{Methods}

\DeclareMathAlphabet{\mathcal}{OMS}{cmsy}{m}{n}

\usepackage{cite}

\title{\begin{flushleft}{\begin{spacing}{1}nnMIL: a generalizable multiple instance learning framework for computational pathology
    \end{spacing}}\end{flushleft}}

\makeatletter
\let\saved@includegraphics\includegraphics
\AtBeginDocument{\let\includegraphics\saved@includegraphics}
\renewenvironment*{figure}{\@float{figure}}{\end@float}
\makeatother

\begin{document}
\maketitle\vspace{-20mm}
\begin{spacing}{1.4}
\noindent 
Xiangde Luo$^{1}$, 
Jinxi Xiang$^{1}$,
Yuanfeng Ji$^{1}$,
Ruijiang Li$^{\textbf{*}1,2}$
\end{spacing}

\vspace{-7mm}
\begin{spacing}{1.4}
\begin{affiliations}
\item Department of Radiation Oncology, Stanford University School of Medicine, Stanford, CA, USA.
\item Stanford Institute for Human-Centered Artificial Intelligence, Stanford, CA, USA
  \\\textbf{*} Correspondence to Ruijiang Li (\textbf{\texttt{rli2@stanford.edu}}).
\end{affiliations}
\end{spacing}
\noindent\textbf{First Version:} 18 November 2025; \textbf{Second Version:} 1 July 2026.
\vspace{-3mm}
\begin{spacing}{1.2}

\end{spacing}

\clearpage
\Heading{Abstract}
\begin{spacing}{1.4}

\noindent
Computational pathology holds substantial promise for improving diagnosis and guiding treatment decisions. Recent pathology foundation models enable the extraction of rich patch-level representations from large-scale whole-slide images (WSIs), but current approaches for aggregating these features into slide-level predictions remain constrained by design limitations that hinder generalizability and reliability. Here we present \textbf{nnMIL}, a simple yet broadly applicable multiple-instance learning framework that connects patch-level foundation models to robust slide-level clinical prediction. nnMIL introduces random sampling at both the patch and feature levels, enabling large-batch optimization, task-aware sampling strategies, and efficient and scalable training across datasets and model architectures. A lightweight aggregator performs sliding-window inference to generate ensemble slide-level predictions and supports principled uncertainty estimation. Across 40,000 WSIs encompassing 35 clinical tasks and four pathology foundation models, nnMIL consistently outperformed existing MIL methods for disease diagnosis, histologic subtyping, molecular biomarker detection, and pan-cancer prognosis prediction. It further demonstrated strong cross-model generalization, reliable uncertainty quantification, and robust survival stratification in multiple external cohorts. In conclusion, nnMIL offers a practical and generalizable solution for translating pathology foundation models into clinically meaningful predictions, advancing the development and deployment of reliable AI systems in real-world settings.

\end{spacing}

\newpage

\begin{spacing}{1.35}
\Heading{Introduction}

\noindent
Histopathology is the gold standard for disease diagnosis and provides valuable information for informing prognosis and treatment decisions\cite{bera2019artificial,yates2025new,bhinder2021artificial,van2021deep}. Recent advances in computational pathology and the advent of pathology foundation models have enabled powerful patch-level feature representations learned from large-scale whole-slide images (WSIs)\cite{xu2024whole,wang2024pathology,xiang2025vision,hoptimus0,chen2024towards,lu2024visual,vorontsov2024foundation,ma2024towards,huang2023visual}. These foundation models have shown strong diagnostic performance and generalization across diverse tissue types\cite{wang2025foundation,campanella2025real,kondepudi2025foundation}. However, translating patch-level representations into slide-level predictions remains challenging, particularly when clinical deployment requires both performance and reliability\cite{shao2025mil}.

Multiple instance learning (MIL) has become the standard paradigm for aggregating patch-level representations into a slide-level prediction, providing a crucial link between foundation model embeddings and downstream clinical applications\cite{ilse2018attention,lu2021data}. Because the number of patches is highly variable across WSIs, existing MIL methods are limited to a batch size of one for training\cite{ilse2018attention,lu2021data,zhang2022dtfd,li2021dual,xiang2023exploring,shao2021transmil,li2024dynamic}. This constraint limits the use of advanced optimization strategies and impedes model convergence \cite{neidlinger2024benchmarking}. Although recent attention- and transformer-based variants of MIL have shown modest performance gains, they largely focus on architectural refinements for specific applications rather than addressing the fundamental issues of generalizability and robustness\cite{ma2025pathbench,neidlinger2024benchmarking,el2025whole,shao2025mil,chen2022pan}. 

Beyond architectural considerations, progress in MIL has also been shaped by training procedures. Multiple studies have reported substantial performance variability even using the same MIL architecture\cite{neidlinger2024benchmarking,ma2025pathbench,vaidya2025molecular,ding2024multimodal,shao2025mil}. In general machine learning, batch construction, sampling strategy, loss formulation, and optimization heuristics all have a significant impact on model performance, and in some cases more so than architectural choices themselves\cite{he2019bag,isensee2021nnu}. Despite their importance, these components have received little systematic analysis, making their true impact difficult to assess and hindering the development of a unified, generalizable training recipe for MIL. 

Here, we propose a training-centric perspective that focuses on efficient training in the foundation-model era, rather than MIL network design. We systematically investigate MIL training strategies and distill them into nnMIL, a simple yet generalizable framework that unifies patch-level foundation-model embeddings with stable and scalable slide-level learning. nnMIL resolves the long-standing limitation of single-batch training by introducing patch sampling, which converts variable-length bags into fixed-length sub-bags and enables large-batch optimization using balanced mini-batches. For slide-level aggregation, nnMIL employs a sliding-window scheme that integrates predictions from multiple overlapping sub-sampled embeddings, effectively functioning as an ensemble and providing principled uncertainty estimates for model outputs\cite{lakshminarayanan2017simple,gal2016dropout}. Furthermore, nnMIL offers a plug-and-play performance enhancement that is compatible with most existing MIL networks, underscoring its extensibility and value for future methodological developments.

Comprehensive evaluation and benchmarking across 35 clinically relevant tasks involving nearly 40,000 WSIs demonstrate that nnMIL consistently outperforms existing MIL methods in disease diagnosis and subtyping, molecular biomarker detection, and pan-cancer prognosis prediction, regardless of the pathology foundation models used for feature extraction. Beyond its strong performance, nnMIL provides quantification of model confidence by computing uncertainty scores for model predictions, which substantially improves accuracy among low-uncertainty cases while flagging the most uncertain cases for expert review in disease diagnosis. For prognosis prediction, the uncertainty scores can be used to further refine survival stratification within the high-risk population. By systematically optimizing the configuration of multiple instance learning, nnMIL establishes a versatile framework for solving clinically relevant tasks in computational pathology.

\clearpage

\Heading{Results}

\heading{Overview of nnMIL}

\noindent
We developed nnMIL, a simple yet generalizable framework for multiple instance learning (MIL) designed to bridge the gap between patch-level foundation models and slide-level clinical prediction (\textbf{\cref{fig:figure1}}). The framework operates on patch features extracted from various pathology foundation models (such as GigaPath\cite{xu2024whole}, UNI\cite{chen2024towards},  H-Optimus-0 (H0)\cite{hoptimus0} or Virchow2\cite{zimmermann2024virchow2}) and trains a slide-level aggregator for task-specific prediction. To address a long-standing problem in MIL, namely that training is constrained by a batch size of one due to varying bag sizes (i.e., number of patches) across WSIs, we redesigned the training pipeline to enable efficient large-batch optimization. By introducing patch sampling, nnMIL samples fixed-length sub-bags from variable-length bags, thereby supporting substantially larger and better-balanced mini-batches. Furthermore, a task-specific batch sampler was developed to improve data utilization across different tasks, collectively enhancing training efficiency, stability, and overall performance. Finally, we designed the slide-level aggregator such that it preserves the semantic integrity of pretrained features by learning from a subset of embeddings and predicts across the entire feature embedding space via a sliding-window inference scheme. This produces an ensemble of predictions, allowing estimation of the uncertainty in model outputs. 

We evaluated and benchmarked nnMIL across 35 computational pathology tasks using nearly 40,000 unique WSIs and features extracted by four pathology foundation models. The results demonstrate that nnMIL consistently outperforms existing MIL methods across all foundation models\cite{ilse2018attention,zhang2022dtfd,li2021dual,xiang2023exploring,shao2021transmil,li2024dynamic}. Furthermore, nnMIL yields model uncertainty estimates that align well with actual reliability, enhancing the clinical utility of slide-level predictions.

\begin{figure*}[hptb]
\centering
\includegraphics[width=1.00\textwidth]{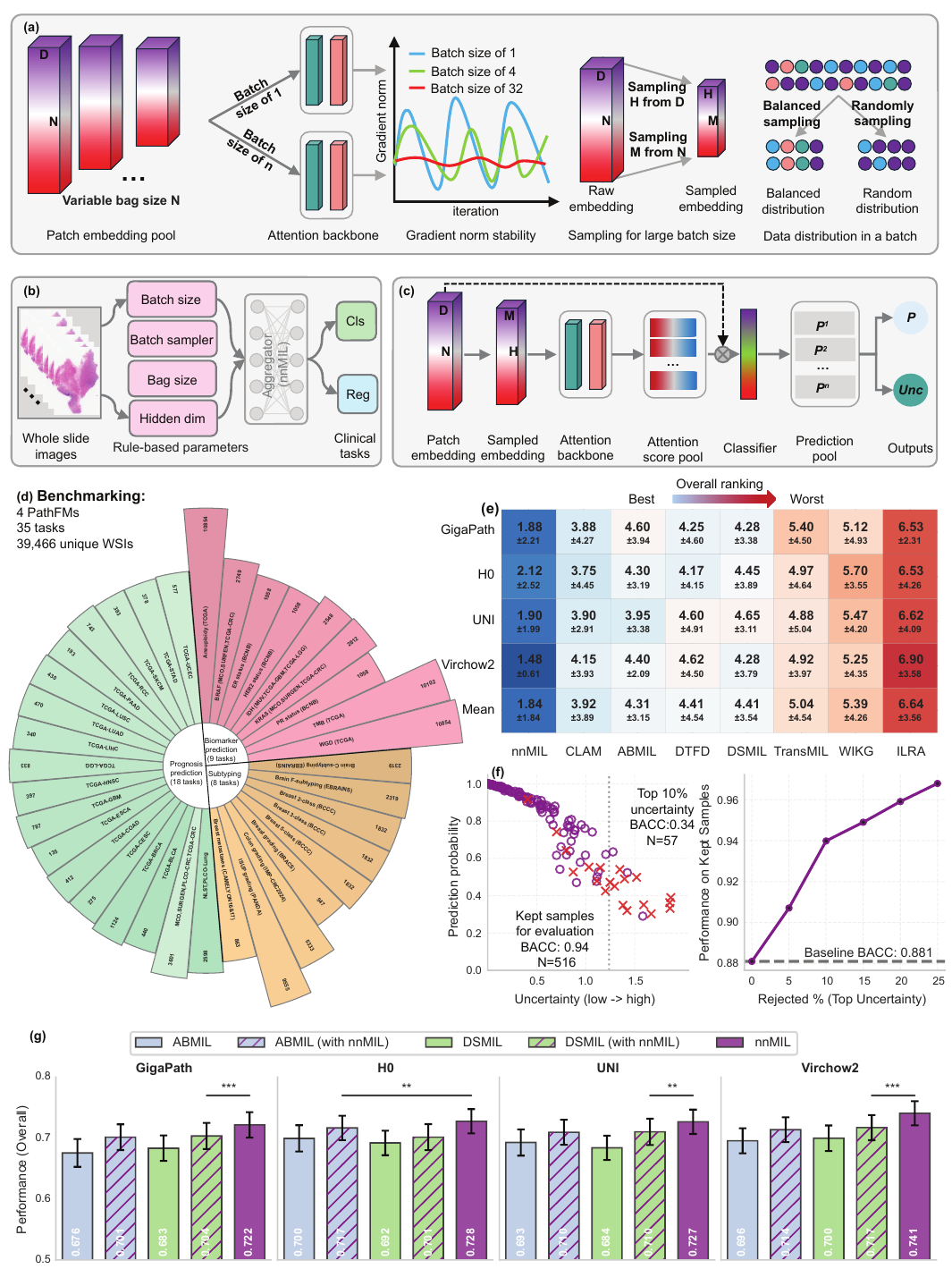}
\end{figure*}
\begin{figure*}
\caption{\textbf{Overview of nnMIL framework and evaluation.} nnMIL is a generalizable multiple instance learning framework that shows superior performance across 35 slide-level computational pathology tasks. \textbf{(a), }The motivation of \texttt{nnMIL} is to enable MIL training in a \emph{simple yet generalizable} manner. We realize this idea by leveraging large-batch optimization to enhance both robustness and effectiveness of model training~\cite{goyal2017accurate,mandt2017stochastic}. Specifically, we introduce stochastic sampling at both the patch level (randomly sampling \textbf{M} out of \textbf{N} patches) and the feature level (randomly sampling \textbf{H} out of \textbf{D} embedding dimensions), combined with task-aware batch construction tailored for MIL optimization. \textbf{(b)} A rule-based parameterization strategy, determined according to dataset characteristics, is further employed to simplify and streamline both training and inference. \textbf{(c), }The simplified attention-based aggregator for slide-level prediction and uncertainty estimation. \textbf{(d), }Overview of the evaluation datasets encompassing 35 slide-level clinical tasks across three major categories: disease classification and subtyping (8 tasks), molecular biomarker detection (9 tasks), and prognosis prediction (18 tasks). \textbf{(e), }Overall ranking scores for MIL methods across all 35 slide-level clinical tasks, four pathology foundation models (GigaPath\cite{xu2024whole}, H0\cite{hoptimus0}, UNI\cite{chen2024towards}, and Virchow2\cite{zimmermann2024virchow2}) and seven existing MIL methods (CLAM\cite{lu2021data}, DTFD\cite{zhang2022dtfd}, DSMIL\cite{li2021dual}, ILRA\cite{xiang2023exploring}, TransMIL\cite{shao2021transmil}, WIKG\cite{li2024dynamic} and ABMIL\cite{ilse2018attention}). \textbf{(f), }Reliability analysis of model predictions based on uncertainty scores (\textit{\textbf{Unc}}) in the left panel shows that removing cases with the highest uncertainty scores could improve model performance on disease diagnosis (BCCC 3 cls) in the right panel. Purple open circles indicate correctly classified cases, orange crosses denote misclassified. \textbf{(g), }Comparison among conventional ABMIL and DSMIL trained with their default settings following the original works, ABMIL and DSMIL trained with nnMIL strategies, and the complete nnMIL framework across four pathology foundation models. {Bars show mean values and error bars denote the standard error of the mean across all 35 tasks (40 cohorts). Statistical significance was determined using a two-sided Wilcoxon signed-rank test, where ** indicates $P$ < 0.01, and *** indicates $P$ < 0.001.} WSI: whole-slide image; PathFMs: pathology foundation models; \textit{Cls}: classification; \textit{Reg}: regression; \textit{\textbf{P}}: prediction; \textit{\textbf{Unc}}: uncertainty; BACC: balanced accuracy.
}
\label{fig:figure1}
\end{figure*}

\heading{Disease diagnosis and subtyping}

\noindent
Accurate disease classification and subtyping are fundamental tasks in computational pathology, providing crucial information for diagnosis, treatment selection, and prognosis assessment. We conducted benchmark experiments on eight challenging disease classification and subtyping tasks, including skin cancer subtyping (2 cls, 3 cls, and 5 cls, cls means classes) using the Basal Cell Carcinoma Classification (BCCC) dataset\cite{yacob2023weakly} (1,832 samples), breast cancer subtyping (7 cls) using the BReAst Carcinoma Subtyping (BRACS) dataset (547 samples)\cite{brancati2022bracs}, brain tumor subtyping (12 cls and 30 cls) using the EBRAINS dataset (2,319 samples)\cite{roetzer2022digital}, colorectal cancer diagnosis using IMP-CRC2024 (5,333 samples)\cite{neto2024interpretable} and Gleason grading of prostate cancer using PANDA (9,555 samples)\cite{bulten2022artificial}. For a fair comparison, we used the official or widely adopted data splits for model training, validation and testing.

Averaged over eight subtyping tasks, nnMIL achieved the best overall performance among all evaluated MIL methods (\textbf{\cref{fig:figure2} (a)}, \textbf{\hyperref[edfig:1]{Extended Data Fig. 1} (a)}, and \textbf{\hyperref[tab:diagnosis_gigapath]{Supplementary Tables 1--4}}). Compared with the second-best method, ABMIL\cite{ilse2018attention}, nnMIL improved performance by 2.6-3.8\% across four pathology foundation models. The performance gain reached up to 6.1\% for GigaPath and 7.3\% for UNI compared with ILRA, respectively. In addition, nnMIL exhibited a marked reduction in performance variance ($P$ < 0.001), indicating greater stability across different cancer types. The overall performance of nnMIL was highly consistent across different pathology foundation models, falling within a narrow range of 80.7\% and 82.0\%. This suggests that nnMIL's advantage is largely model-agnostic and not tied to any specific feature extractor.

Among all MIL methods, nnMIL ranked first on five tasks, second-best or comparable across the others, reflecting strong performance in diagnosing diverse diseases (\textbf{\cref{fig:figure2} (b)-(i)}). Compared with the second-best method ABMIL, nnMIL achieved its largest gain on the EBRAINS Fine classification task with a 10.5\% relative improvement in balanced accuracy (BA): 0.724 \textit{vs} 0.656, $P$ < 0.001, while showing a slight 1.5\% decrease on the PANDA dataset (Kappa: 0.924 \textit{vs} 0.938, $P$ < 0.001). In the smallest subtyping dataset, BRACS, nnMIL achieved a BA of 0.444, representing a 7.2\% relative improvement over DTFD with a BA of 0.414.

To assess the reliability of slide-level predictions, we further analyzed the relationship between model uncertainty and classification performance by conducting a selective prediction experiment through sample rejection (\textbf{\cref{fig:figure2} (j)-(m)} and \textbf{\hyperref[edfig:2]{Extended Data Fig. 2} (a)}). As slides with the highest uncertainty scores were progressively excluded, nnMIL's performance on the retained slides increased steadily across various tasks, suggesting that uncertainty estimates are well aligned with prediction reliability. For instance, excluding just 10\% of the slides with the highest uncertainty scores achieved substantial error reduction by as much as 60\% from 5.0\% to 2.0\% in the IMP-CRC2024 dataset. These results indicate that the uncertainty scores estimated by nnMIL can be used to identify complex cases with low-confidence model predictions, enabling selective triage for pathologist review and thereby further improving overall diagnostic performance.

\begin{figure*}[hptb]
\centering
\includegraphics[width=1.00\textwidth]{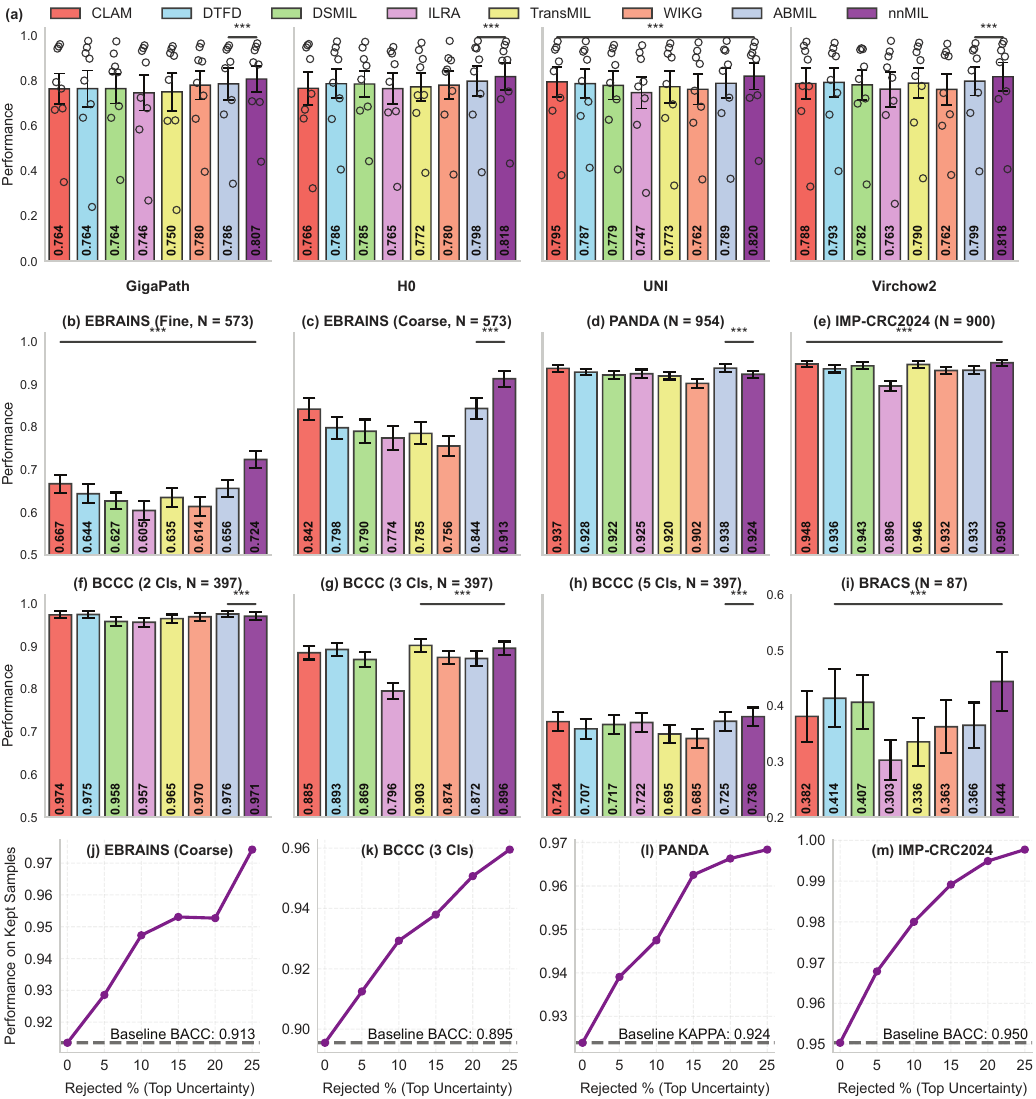}
\end{figure*}
\begin{figure*}
\caption{\textbf{Disease classification and subtyping.} \textbf{(a), }Comparisons of MIL methods for performance on 8 disease classification and subtyping tasks across four pathology foundation models (GigaPath, H0, UNI and Virchow2). {Bars show mean values and error bars denote the standard error of the mean across all 8 tasks, each dot represents one task.} nnMIL was compared with seven widely used MIL methods including CLAM, DTFD, DSMIL, ILRA, TransMIL, WIKG and ABMIL. \textbf{(b)-(i), }Comparison across eight individual tasks using UNI as the feature extractor, including EBRAINS (30 cls Fine and 12 cls Coarse tasks, where cls means classes), PANDA (7 cls), IMP-CRC2024 (3 cls), BCCC (2-, 3-, and 5-cls tasks), and BRACS (7 cls). Performance was evaluated using balanced accuracy (BACC), except for PANDA, which was assessed with Cohen's Kappa. {In \textbf{(b)-(i)}, bars represent mean values and error bars represent standard deviation, derived from 1,000 bootstrap replicates on each independent test set.} {Statistical significance was determined using a two-sided Wilcoxon signed-rank test (\textbf{(a)--(i)}), where * indicates $P$ < 0.05, ** indicates $P$ < 0.01, and *** indicates $P$ < 0.001.} \textbf{(j)-(m), }Performance analysis after excluding cases with the highest uncertainty scores.
}
\label{fig:figure2}
\end{figure*}

\heading{Molecular biomarker detection}

\noindent
Accurate prediction of protein expression or gene mutation from routine histopathology slides can accelerate treatment decisions and reduce unnecessary testing\cite{el2025whole}. In this study, we evaluated the ability of nnMIL and other MIL methods to predict molecular phenotypes from whole-slide histopathology images. Specifically, we conducted experiments on nine clinically relevant molecular biomarkers across 12 datasets, including three protein expression biomarkers (ER, PR, and HER2), three gene mutations (\textit{BRAF}, \textit{KRAS}, and \textit{IDH}), and three pan-cancer genomic biomarkers (whole-genome doubling (WGD), tumor mutational burden (TMB), and aneuploidy). These biomarkers were derived from large-scale cohorts, including an early breast cancer core needle biopsy dataset (BCNB, 1,058 samples) containing PR, ER, and HER2 statuses\cite{xu2021predicting}; three colorectal cancer cohorts for \textit{BRAF} and \textit{KRAS} mutations (SURGEN\cite{myles2025surgen} as the development set, $n=799$; MCO\cite{ward2015mco,jonnagaddala2016integration} and TCGA-CRC\cite{weinstein2013cancer} as external evaluation sets, $n=1,492$ and $n=521$); three brain tumor cohorts for \textit{IDH} mutation (MUV as the development set\cite{roetzer2022digital}, $n=872$; TCGA-GBM and TCGA-LGG as external evaluation sets\cite{weinstein2013cancer}, $n=834$ and $n=842$); and the pan-cancer TCGA dataset ($n=10,854$) annotated with WGD, TMB, and aneuploidy statuses\cite{taylor2018genomic}. These evaluation datasets and tasks provide a comprehensive benchmark for evaluating MIL methods for slide-level molecular prediction. All models were trained with identical slide-level labels and compared across four pathology foundation models (Virchow2, GigaPath, UNI, and H0) to ensure fair comparison.

Averaged across all nine biomarkers, nnMIL consistently outperformed other MIL methods across four pathology foundation models (GigaPath, H0, UNI, and Virchow2). Specifically, nnMIL with Virchow2 as the feature extractor achieved the highest performance with a mean performance of 0.794, surpassing ABMIL and DSMIL\cite{li2021dual} by 3.3--4.1\% ($P$~<~0.001; \textbf{\cref{fig:figure3} (a)}, \textbf{\hyperref[edfig:1]{Extended Data Fig. 1} (b)}, and \textbf{\hyperref[tab:biomarker_gigapath]{Supplementary Tables 5--8}}). To further assess performance for individual biomarkers, we compared all MIL methods using Virchow2 as the feature extractor (\textbf{\cref{fig:figure3} (b)-(m)}). For protein expression prediction, nnMIL significantly outperformed all competing approaches on ER, HER2, and PR in the BCNB dataset ($P$ < 0.001). For gene mutation prediction (\textit{BRAF}, \textit{KRAS} and \textit{IDH}), nnMIL ranked first in four of the six mutation prediction settings. In addition, nnMIL outperformed all comparison methods in WGD and TMB prediction, achieving high AUCs of 0.845 and 0.876, respectively, and also attained the second-best performance on aneuploidy regression (Pearson's r = 0.630 $vs$ 0.637 for TransMIL\cite{shao2021transmil}). Collectively, these results highlight nnMIL's ability to generalize across both expression- and genome-level biomarkers, linking histomorphological features to underlying molecular alterations.

To further evaluate the reliability of biomarker prediction, we investigated the relationship between model uncertainty and prediction accuracy across all biomarkers (\textbf{\cref{fig:figure3} (n)-(q)} and \textbf{\hyperref[edfig:2]{Extended Data Fig. 2} (b)}). As slides with the highest uncertainty scores obtained from nnMIL were progressively excluded, performance generally improved, indicating that uncertainty estimates were closely aligned with prediction reliability. Removing the top 25\% most-uncertain samples increased the AUC from 0.905 to 0.940 for ER prediction, from 0.831 to 0.873 for BRAF, and from 0.863 to 0.916 for IDH, while the pan-cancer TMB prediction improved from 0.877 to 0.954. These results demonstrate that nnMIL can provide high-confidence biomarker predictions for a substantial fraction of samples while identifying low-confidence cases that may benefit from further molecular testing.

\begin{figure*}[hptb]
\centering
\includegraphics[width=1.00\textwidth]{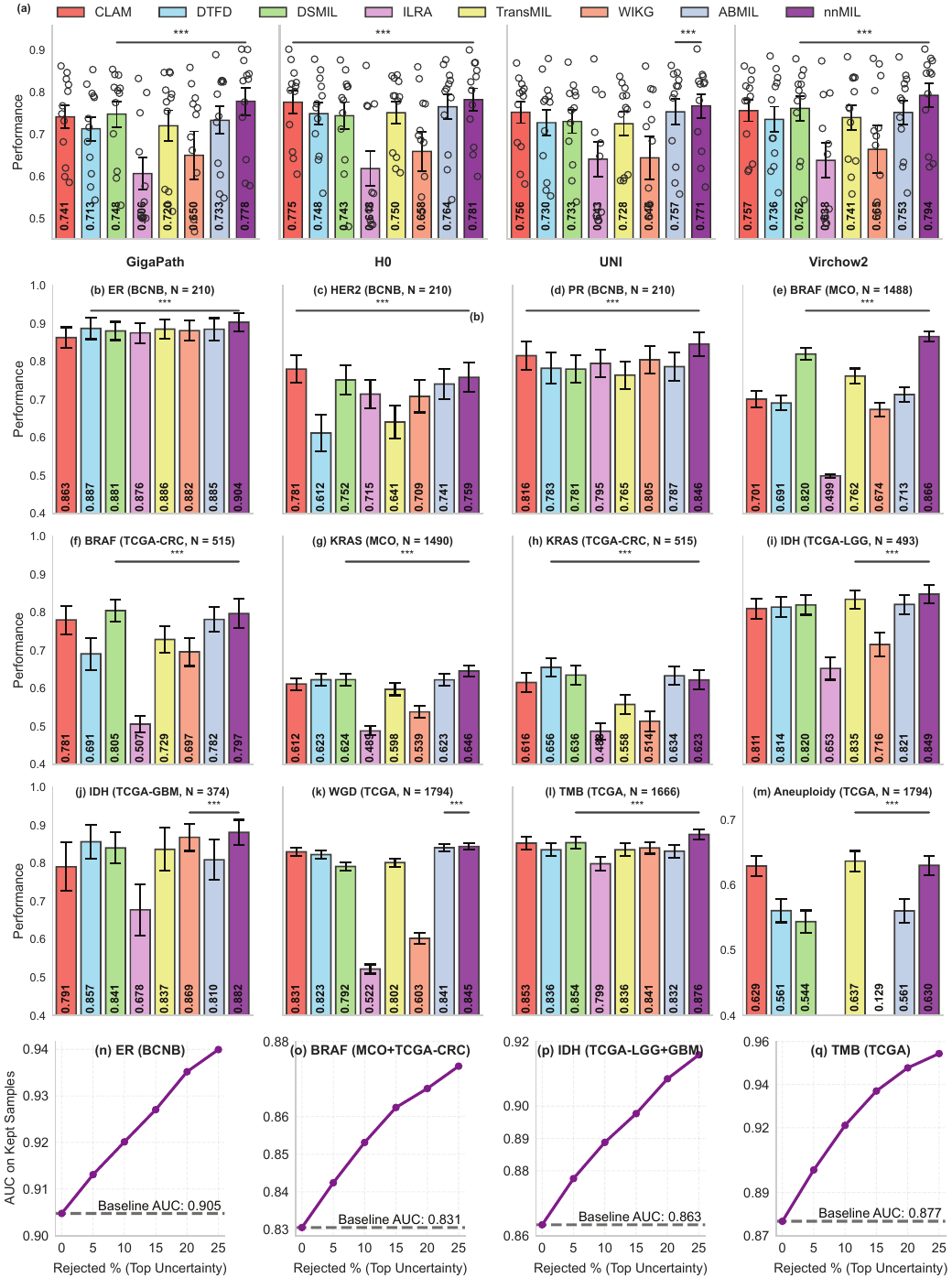}
\end{figure*}
\begin{figure*}
\caption{\textbf{Molecular biomarker detection.} \textbf{(a), }Average performance of nnMIL and comparative methods across 12 datasets representing 9 molecular biomarkers evaluated on four pathology foundation models (GigaPath, H0, UNI, and Virchow2). nnMIL was benchmarked against seven widely used MIL methods, including {CLAM}, DTFD, DSMIL, ILRA, TransMIL, WIKG, and ABMIL. {Bars show mean values and error bars denote the standard error of the mean across all 12 cohorts, each dot represents one cohort.}
\textbf{(b)--(m), }Comparisons across 12 individual datasets using Virchow2 as the feature extractor, covering three breast cancer biomarkers (ER, HER2, and PR) from the BCNB cohort; two colorectal cancer biomarkers (\textit{BRAF} and \textit{KRAS}) from the TCGA-CRC and MCO cohorts; one brain tumor biomarker (\textit{IDH}) from the TCGA-LGG and TCGA-GBM cohorts; and three pan-cancer genomic biomarkers (WGD, TMB, and Aneuploidy) from TCGA. Performance was evaluated using the area under the curve (AUC), except for Aneuploidy, which was a regression task assessed using the Pearson correlation coefficient. {In \textbf{(b)--(m)}, bars represent mean values and error bars represent standard deviation, derived from 1,000 bootstrap replicates on each independent test set.} {Statistical significance was determined using a two-sided Wilcoxon signed-rank test (\textbf{(a)--(m)}), where * indicates $P$ < 0.05, ** indicates $P$ < 0.01, and *** indicates $P$ < 0.001.} 
\textbf{(n)--(q), }Performance analysis after excluding samples with the highest uncertainty scores from nnMIL.
}
\label{fig:figure3}
\end{figure*}

\heading{Pan-cancer prognosis prediction}

\noindent
Accurate prediction of survival outcomes from routine whole-slide pathology images can enable personalized treatment. To evaluate the prognostic capability of nnMIL, we curated a pan-cancer cohort of 6,602 patients from The Cancer Genome Atlas (TCGA), comprising 7,927 diagnostic hematoxylin and eosin (H\&E) whole-slide images with corresponding follow-up data\cite{weinstein2013cancer}. Following previous work\cite{xiang2025vision}, we trained and evaluated nnMIL on each cancer type using five-fold cross-validation, with disease-specific survival as the clinical endpoint. 

Across all 16 cancer types, nnMIL consistently outperformed existing multiple instance learning approaches (\textbf{\cref{fig:figure4} (a)}, \textbf{\hyperref[edfig:1]{Extended Data Fig. 1} (c)}, and \textbf{\hyperref[tab:prognosis_gigapath]{Supplementary Tables 9--12}}). When trained with Virchow2 as the feature extractor, nnMIL achieved the highest mean C-Index of 0.670, significantly outperforming the second-best method, DSMIL\cite{li2021dual}, which achieved a mean C-Index of 0.626 ($P$~<~0.001).
nnMIL maintained stable prognostic performance across all pathology foundation models with a mean C-Index of 0.656, 0.651, and 0.641 for UNI, H0, and GigaPath, respectively. In comparison, the classical ABMIL method showed consistently lower performance, with mean C-Index values of 0.604, 0.611, 0.586, and 0.615 on UNI, H0, GigaPath, and Virchow2, respectively. Overall, nnMIL improved the mean C-Index by 4.2--7.0\% compared with the second-best MIL method under each foundation model, demonstrating superior survival prediction in pan-cancer settings.

To further assess the prognostic relevance of nnMIL, we investigated survival stratification using Kaplan--Meier (KM) curves based on predictions from nnMIL trained on Virchow2 (\textbf{\cref{fig:figure4} (b)}). For each of the 16 cancer types, patients were stratified into high- and low-risk groups based on the median predicted risk score. The resulting KM curves showed clear and consistent separation between risk groups, with log-rank $P$~<~0.05 in 14 of 16 cohorts (LUSC: $P$ = 0.073, PAAD: $P$ = 0.089). The survival stratification was particularly pronounced in BRCA, CESC, COADREAD, RCC, and UCEC, with hazard ratios greater than 3.0.

Finally, we performed a more detailed comparison of survival prediction performance within individual cancers (\textbf{\cref{fig:figure4} (c)}). nnMIL achieved the highest C-Index in 15 out of 16 cancer types, demonstrating consistently superior performance over the two other top-performing MIL methods. The most notable improvements were observed in BRCA, LGG, ESCA, and LIHC cancers, where nnMIL exceeded the second-best method by 6--15\% in C-Index.  In contrast, ABMIL and DSMIL generally yielded lower and more variable results across cohorts. Collectively, these analyses demonstrate that nnMIL provides more accurate and robust survival prediction across a wide range of tumor types. 

\begin{figure*}[hptb]
\centering
\includegraphics[width=1.00\textwidth]{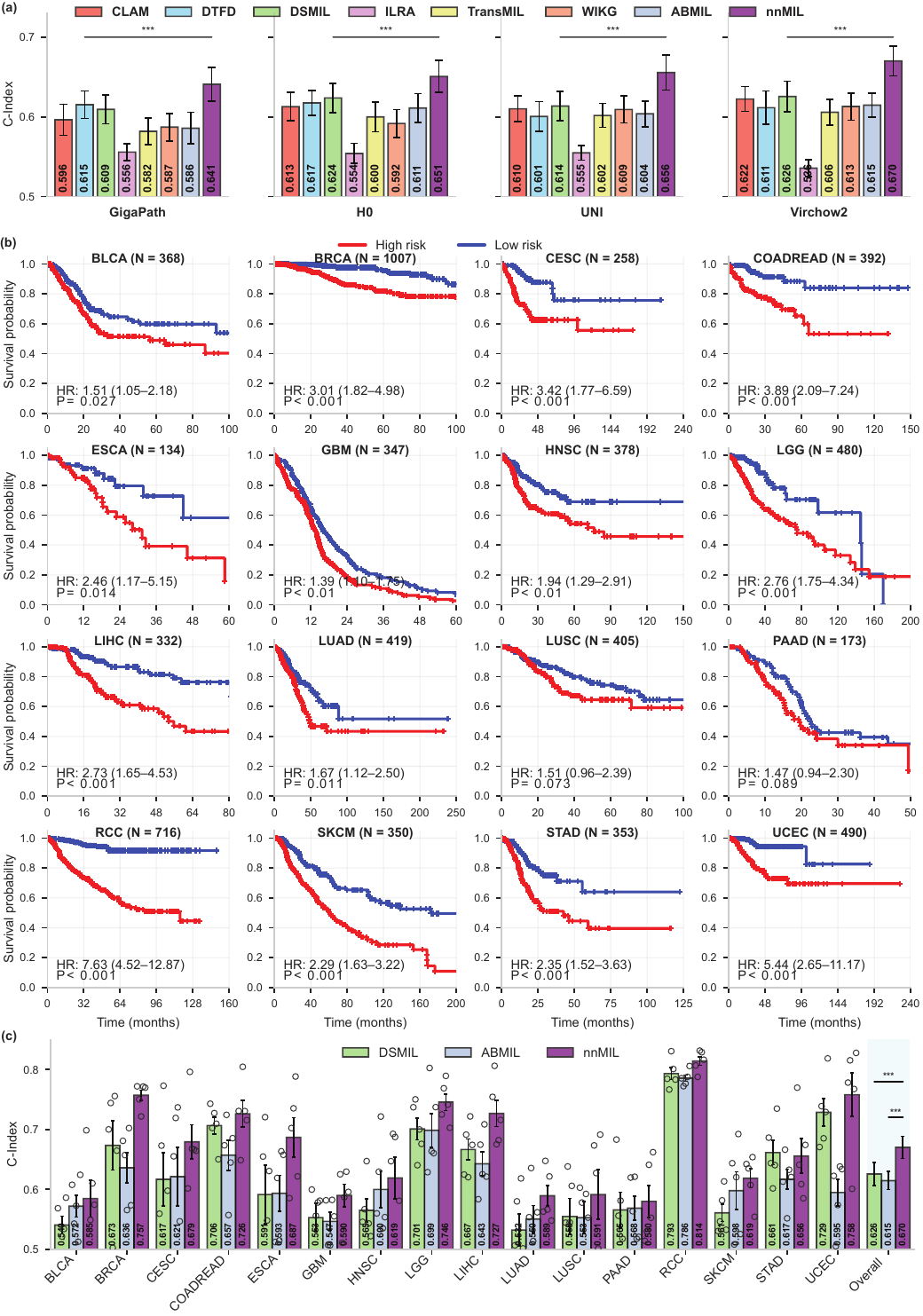}
\end{figure*}
\begin{figure*}
\caption{\textbf{Pan-cancer prognosis prediction.}
\textbf{(a), }Average performance of nnMIL and comparative methods across 16 cancer types evaluated on four pathology foundation models (GigaPath, H0, UNI, and Virchow2). nnMIL was benchmarked against seven widely used MIL methods, including CLAM, DTFD, DSMIL, ILRA, TransMIL, WIKG, and ABMIL. {Bars show mean values and error bars denote the standard error of the mean across all 16 tasks.} 
\textbf{(b), }Kaplan--Meier survival plots of nnMIL across 16 cancer types using Virchow2 as the feature extractor, including bladder urothelial carcinoma (BLCA), breast invasive carcinoma (BRCA), cervical squamous cell carcinoma and endocervical adenocarcinoma (CESC), colorectal and rectal adenocarcinoma (COADREAD), esophageal carcinoma (ESCA), glioblastoma multiforme (GBM), head and neck squamous cell carcinoma (HNSC), low-grade glioma (LGG), liver hepatocellular carcinoma (LIHC), lung adenocarcinoma (LUAD), lung squamous cell carcinoma (LUSC), pancreatic adenocarcinoma (PAAD), renal cell carcinoma (RCC), skin cutaneous melanoma (SKCM), stomach adenocarcinoma (STAD), and uterine corpus endometrial carcinoma (UCEC). Survival differences between high- and low-risk groups were assessed using a two-sided log-rank test.
\textbf{(c), }Prognostic performance in each cancer type for the top three methods on Virchow2. {In \textbf{(c)}, bars represent mean values and error bars represent standard error of the mean based on five-fold cross-validation, each dot represents one fold. Statistical comparisons were performed using a two-sided Wilcoxon signed-rank test (\textbf{(a)} and \textbf{(c)}), where * indicates $P$ < 0.05, ** indicates $P$ < 0.01, and *** indicates $P$ < 0.001. }
}
\label{fig:figure4}
\end{figure*}

\heading{Generalizability across institutions and clinical tasks}

\noindent
Variations in tissue fixation, staining protocols, and slide digitization can lead to substantial heterogeneity in WSI appearance across institutions. Such differences can introduce significant distribution shifts that challenge the generalizability of computational pathology models. Therefore, it is essential to rigorously evaluate model robustness and effectiveness across multiple independent external cohorts to assess its ability to generalize under diverse real-world clinical conditions. 

We first evaluated the cross-institutional generalization ability of nnMIL for breast cancer micro-metastasis detection using the CAMELYON16 and CAMELYON17 datasets\cite{bejnordi2017diagnostic,bandi2018detection}. Specifically, we considered a three-class classification setting (normal tissue, micro-metastasis, and macro-metastasis) based on the multi-institution CAMELYON benchmarks. CAMELYON16 was used for model development, comprising 399 WSIs collected from Radboud University Medical Center in Nijmegen (RUMC) and University Medical Center Utrecht (UMCU), while CAMELYON17 was reserved for external evaluation with 464 WSIs from five independent institutions: RUMC, Canisius-Wilhelmina Hospital in Nijmegen (CWZ), UMCU, Rijnstate Hospital in Arnhem (RST), and the Laboratory of Pathology East-Netherlands in Hengelo (LPON). For fair comparison, we followed the widely adopted data splits and evaluation protocol in prior studies, reporting WSI-level accuracy for each institution\cite{ling2025comprehensive,cai2025attrimil}. The results show that nnMIL consistently outperformed other MIL methods and demonstrated strong generalization across institutions for disease diagnosis (\textbf{\hyperref[tab:camelyon_results_acc]{Supplementary Tables 13 and 14}}).

Survival prediction is inherently more challenging than disease classification, as it requires learning prognostic cues that are often subtle and confounded by clinical risk factors. Here, we collected four colorectal cancer cohorts, including PLCO (with 1279 WSIs from 661 patients)\cite{gohagan2000prostate}, TCGA-CRC (with 596 WSIs from 567 patients)\cite{weinstein2013cancer}, MCO (with 1301 WSIs from 1278 patients)\cite{ward2015mco,jonnagaddala2016integration} and SURGEN (with 425 WSIs from 425 patients)\cite{myles2025surgen} and two non-small cell lung cancer cohorts, comprising PLCO (with 1455 WSIs from 470 patients)\cite{gohagan2000prostate} and NLST (with 1143 WSIs from 414 patients)\cite{national2011national}. PLCO is a multi-institutional study with sufficient follow-up, and therefore was used for model development in both cancers. All MIL methods were evaluated for disease-free survival prediction within each disease in the independent external cohorts.

Across four independent external cohorts, nnMIL achieved the highest or near-highest concordance indices across pathology foundation models, outperforming other MIL methods such as ABMIL, CLAM, DSMIL, TransMIL, and DTFD (\textbf{\cref{fig:figure5} (a)}, \textbf{\hyperref[edfig:1]{Extended Data Fig. 1} (d)}, and \textbf{\hyperref[tab:generalization_prognosis_gigapath]{Supplementary Tables 15--18}}). The performance gains were consistent across foundation models, with nnMIL surpassing the second-best method by 2--9\% in C-Index ($P$ < 0.001), indicating that its advantage is not dependent on a specific pretrained representation. Collectively, these results demonstrate that nnMIL delivers stronger survival discrimination than existing MIL approaches when evaluated on independent external cohorts.

We then performed a detailed per-cohort evaluation of survival prediction performance using Virchow2 as the feature extractor (\textbf{\cref{fig:figure5} (b)}). Across all four external cohorts, nnMIL achieved the highest C-Index, with performance gains of 3.3-7.2\% ($P$ < 0.001) compared with the second-best methods in each dataset (CLAM, DTFD, and WIKG). Notably, nnMIL maintained similarly strong performance across both disease types. In Kaplan--Meier analyses (\textbf{\cref{fig:figure5} (c)}), predicted risk scores by nnMIL significantly stratified patients into distinct prognostic groups based on a median-risk threshold (all with log-rank $P$ < 0.0001). Clear separation between high- and low-risk groups was observed across all external cohorts, with hazard ratios of 2.95 (95\% CI: 1.79--4.86), 4.77 (95\% CI: 3.64--6.26), and 4.38 (95\% CI: 2.69--7.12) for the three CRC cohorts (TCGA-CRC, MCO, and SURGEN) and 2.70 (95\%~CI: 1.96--3.73) for the lung cohort (NLST). These results indicate that nnMIL achieves consistent survival stratification across external datasets and disease types. To validate the clinical interpretability of the model, we overlay representative attention heatmaps on the corresponding whole-slide images. These visualizations show that the high-risk regions identified by nnMIL predominantly localize to tumor cell--rich areas and dense fibroblastic stroma, consistent with features assessed by pathologists during routine diagnostic procedures (\textbf{\hyperref[edfig:3]{Extended Data Fig. 3}}).

We further examined whether the nnMIL risk score retained prognostic significance after controlling for available clinical covariates, including age, sex, grade, and stage (\textbf{\cref{fig:figure5} (d), \hyperref[tab:lung_patient_characteristics]{Supplementary Tables 19 and 20}}). In all external cohorts, the predicted risk scores by nnMIL remained a statistically significant predictor of disease-free survival,  independent of conventional risk factors. Specifically, the hazard ratio estimates were 1.75 (95\% CI: 1.19--2.58) for TCGA-CRC, 1.88 (95\% CI: 1.60--2.22) for MCO, 1.91 (95\% CI: 1.50--2.43) for SURGEN, and 1.40 (95\% CI: 1.18--1.67) for NLST, with all $P$ < 0.01. Collectively, these findings confirm that the nnMIL risk score provides independent prognostic value across both colorectal and lung cancer cohorts.

We also investigated whether the model uncertainty scores generated by nnMIL could provide additional prognostic value beyond the risk score itself (\textbf{\cref{fig:figure5} (e)}). After stratifying patients by the nnMIL-derived risk score, the high-risk group was further subdivided according to estimated uncertainty into high-risk + low-uncertainty and high-risk + high-uncertainty subgroups (the hazard ratios in high-risk groups were 1.9, 2.3, 2.6 and 1.9 for TCGA-CRC, MCO, SURGEN and NLST, respectively, all $P$ < 0.05). Across all external cohorts, the Kaplan--Meier curves showed a consistent pattern, with high-risk + high-uncertainty patients exhibiting the poorest survival, high-risk + low-uncertainty showing intermediate outcomes, and the low-risk group achieving the best prognosis. Consistently, scatter plots of predicted risk versus uncertainty (\textbf{\hyperref[edfig:4]{Extended Data Fig. 4}}) demonstrated a positive association, indicating that uncertainty scores capture intrinsic heterogeneity within the high-risk population and complement risk-based prediction. Interestingly, the uncertainty scores did not further stratify patients among the low-risk population. Overall, these results suggest that integrating uncertainty estimates with risk scores could provide more refined prognostic stratification and highlight a subset of high-risk patients with high variability in model predictions and adverse outcomes.

Beyond across-institution evaluation, we further examined the generalization ability of nnMIL under a challenging clinical multi-task prediction setting. We constructed a multi-task dataset from three colorectal cancer cohorts, comprising tumor grading, molecular biomarker detection, and prognosis prediction, with SURGEN serving as the development cohort and MCO and TCGA-CRC as external evaluation cohorts. In this setting, we trained a unified multi-task model with a shared backbone and four task-specific heads to jointly predict all tasks within a single framework. Under identical training conditions, nnMIL in the multi-task setting consistently outperformed both multi-task ABMIL and nnMIL trained with single-task objectives in terms of overall performance across all cohorts. These results demonstrate that nnMIL maintains robust and competitive performance in clinically relevant multi-task scenarios encompassing tumor grading, molecular biomarker detection, and prognosis prediction (\textbf{\hyperref[tab:singletask_vs_multitask]{Supplementary Table 21}}).

\begin{figure*}[hptb]
\centering
\includegraphics[width=1.00\textwidth]{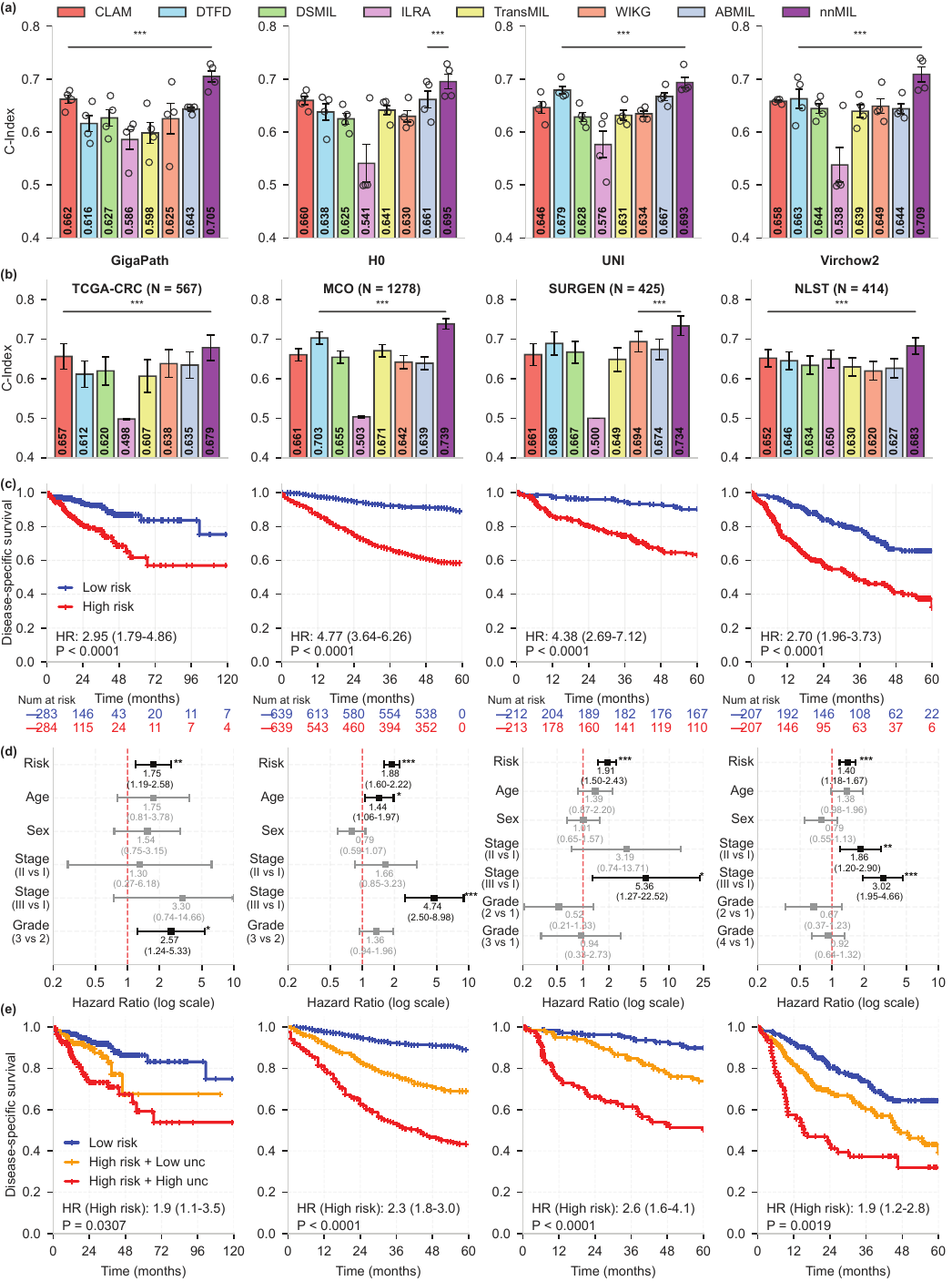}
\end{figure*}
\begin{figure*}
\caption{\textbf{Generalizability of prognostic models in external validation}.
\textbf{(a), }Average performance of nnMIL and comparative methods across four independent external cohorts using four pathology foundation models (GigaPath, H0, UNI, and Virchow2). {Bars show mean values and error bars denote the standard error of the mean across all four cohorts, each dot represents one cohort.} \textbf{(b), }Per-cohort survival prediction performance using Virchow2 features across three colorectal cancer cohorts (TCGA-CRC, MCO, and SURGEN) and one lung cancer cohort (NLST). {In \textbf{(b)}, bars represent mean values and error bars represent standard deviation, derived from 1,000 bootstrap replicates on each independent test set. Statistical comparisons were performed using a two-sided Wilcoxon signed-rank test (\textbf{a} and \textbf{b}), where * indicates $P$ < 0.05, ** indicates $P$ < 0.01, and *** indicates $P$ < 0.001.} \textbf{(c), }Kaplan--Meier survival analyses based on predicted risk score in each external cohort, showing clear separation between high- and low-risk groups.
\textbf{(d), }Multivariate Cox regression analyses controlling for clinical covariates, confirming that nnMIL provides independent prognostic value beyond conventional risk factors.
\textbf{(e), } Kaplan--Meier survival analyses after subdividing the high-risk group by estimated uncertainty scores in each external cohort. Statistical significance for survival differences between high- and low-risk groups was assessed using a two-sided log-rank test.
}
\label{fig:figure5}
\end{figure*}

\heading{Ablation experiments}

\noindent
We first conducted comprehensive ablation studies by progressively adding the major components of the nnMIL framework across four pathology foundation models (Virchow2, GigaPath, UNI, and H0; \textbf{\hyperref[edfig:5]{Extended Data Fig. 5} (a)}). In detail, we progressively added gradient accumulation (effective batch size = 32), patch sampling, and feature sampling, starting from a baseline model trained with a batch size of 1. Our results show that MIL methods benefit from larger effective batch sizes, even when larger batches are achieved solely through gradient accumulation. In addition to the effect of larger batch size, the patch sampling and feature sampling components also contributed to the final performance. Moreover, to enable a more comprehensive and fair comparison, we also applied gradient accumulation to simulate larger batch sizes for other MIL methods during training and evaluation, instead of using their widely adopted default settings with a batch size of one (\textbf{\hyperref[tab:mil_batch1_vs_grad_accum_gigapath]{Supplementary Tables 22--25}}). Similar trends were observed across MIL methods, with larger batch sizes generally leading to improved performance. Overall, the proposed nnMIL method consistently outperformed the modified versions of these baselines. Taken together, these results demonstrate that the performance gains arise from the integrated framework rather than from a single component.

Then, we conducted comprehensive comparisons among the original ABMIL and DSMIL architectures, their modified versions trained with the nnMIL strategies, and the complete nnMIL framework across four pathology foundation models (Virchow2, GigaPath, UNI, and H0, \textbf{\cref{fig:figure1} (g)} and \textbf{\hyperref[tab:mil_comparison_combined]{Supplementary Table 26}}). With these simple modifications, the nnMIL training strategy already enhances the stability and generalization of ABMIL and DSMIL across datasets and feature representations, suggesting that the proposed training strategy alone can lead to promising performance gains. Building on this foundation, the full nnMIL, combining the modified MIL architecture and the rule-based configuration strategy, achieves strong cross-model consistency and the best overall results across 35 benchmarks covering three major categories of tasks: disease diagnosis, biomarker detection, and prognosis prediction. Across all four foundation models, nnMIL consistently yields higher performance and substantially lower performance variance than existing ABMIL and DSMIL approaches, demonstrating that its streamlined architecture and rule-based training design jointly enable scalable, stable, and model-agnostic slide-level prediction.

We further evaluated the robustness of nnMIL for hyperparameter choices, focusing on hidden dimension, bag size, inference stride, dropout ratio, and mini-batch sampling strategy. Across all tasks and datasets, nnMIL exhibited consistently stable behavior under broad parameter variations ({\textbf{\hyperref[edfig:5]{Extended Data Fig. 5} (b)}}). First, the hidden dimension showed a near-flat performance curve, with a broad optimum centered around 256. This flatness indicates that nnMIL does not depend on a finely tuned embedding size; instead, a wide range of moderate values (128--512) achieves nearly identical performance. In practice, selecting 256 serves as a strong default, preserving sufficient representational capacity while controlling memory usage and reducing the need for extensive hyperparameter search. Similarly, model performance remains stable across different bag sizes, with only a mild saturation trend, where performance plateaus beyond the median bag length. This suggests that nnMIL effectively captures essential slide-level information without requiring excessively large bags. Then, inference stride primarily affects computational cost rather than predictive performance. Varying the stride altered efficiency but produced negligible performance differences, enabling users to flexibly adjust inference speed without substantially compromising performance. Finally, the mini-batch sampling strategy also had a noticeable positive impact on model performance in certain tasks, including disease diagnosis and biomarker detection. As shown in {\textbf{\hyperref[edfig:5]{Extended Data Fig. 5} (c)}}, task-aware sampling outperforms random sampling across multiple settings, while preserving low variance. This demonstrates that nnMIL benefits from informed sampling without being highly sensitive to the exact sampling configuration, further supporting its robustness in practical large-scale training settings.

In addition, although performance varied modestly across datasets, nnMIL remained robust to key training hyperparameters, including dropout ratio and random seed, with near-flat performance curves across a broad operating range (\textbf{\hyperref[edfig:5]{Extended Data Fig. 5} (b)}). At the same time, we observed a localized performance dip at intermediate feature subspace dimensions (e.g., 384--640) for the BRAF (MCO) task when using GigaPath features, followed by a consistent recovery at higher dimensions (e.g., 768). It may be caused by the combined effects of stochastic feature subspace sampling and the redundant, anisotropic structure of high-dimensional foundation model embeddings. At intermediate subspace sizes, randomly sampled feature subsets may provide incomplete coverage of task-relevant embedding directions, whereas larger subspaces introduce sufficient redundancy to stabilize attention-based aggregation and recover performance. Together, these results indicate that nnMIL delivers stable and reliable performance across a wide range of hyperparameter configurations, supporting its robustness and practical scalability for large-scale computational pathology.

\Heading{Discussion}

\noindent
In this study, we proposed a training-centric perspective that emphasizes efficient training rather than MIL network design in the foundation-model era. Following this perspective, we introduce nnMIL, a simple yet generalizable framework for multiple instance learning (MIL) that bridges patch-level pathology foundation models with slide-level clinical prediction. Comprehensive evaluations across diverse clinical applications, including disease diagnosis and subtyping, molecular biomarker detection, and pan-cancer prognosis prediction, demonstrate that nnMIL consistently outperforms existing MIL methods while maintaining cross-model consistency and reliability.

A central challenge in MIL for computational pathology is the highly variable number of patches per slide, which has traditionally restricted model training to a batch size of one and limited the use of advanced optimization strategies. nnMIL addresses this issue by converting variable-length bags into fixed-length sampled bags, thereby enabling large and class-balanced mini-batches. Inspired by the rule-based design philosophy of nnU-Net\cite{isensee2021nnu} for image segmentation, nnMIL follows a similar principle of distilling domain knowledge into explicit heuristic rules that guide model configuration, for example, determining an appropriate bag size from the dataset-level distribution. The use of larger batch sizes, suitable batch samplers, and well-chosen bag sizes generally leads to better performance. This design stabilizes gradients, improves convergence, and supports the integration of task-specific batch samplers to alleviate class imbalance\cite{hoffer2017train,yang2020rethinking}. Furthermore, nnMIL adopts a moderate dimension of hidden features to balance model capacity and regularization, which preserves the semantic features of pretrained foundation models without requiring additional projection layers as in existing MIL methods\cite{luo2025ensemble}. While patch-level sampling is introduced to enable stable large-batch optimization in nnMIL, for tasks relying on rare events, it might pose a risk of missing informative patches during training; however, this issue is mitigated at inference by aggregating predictions over all patches in the whole slide. To explicitly test this, we assessed nnMIL for detecting breast cancer micro-metastasis based on independent training and validation datasets CAMELYON16 and CAMELYON17. Our results show that nnMIL achieves competitive performance on this task, indicating that the training-time sampling strategy does not compromise detection of rare pathological patterns\cite{bejnordi2017diagnostic,bandi2018detection} (\textbf{\hyperref[tab:camelyon_results_acc]{Supplementary Table 13}}).

Across multiple slide-level tasks, including disease subtyping, biomarker detection, and prognosis prediction, nnMIL maintained strong generalization and consistently superior performance across cancer types, regardless of the pathology foundation models used for feature extraction. For molecular biomarker detection, models trained on one colorectal or brain tumor cohort generalized effectively to independent cohorts with distinct demographic and imaging characteristics. For prognosis prediction, nnMIL outperformed existing MIL methods in pan-cancer settings. Furthermore, nnMIL demonstrated strong cross-institutional generalization across multiple clinical tasks and, beyond its performance on individual tasks, maintains robust performance in multi-task settings encompassing tumor grading, molecular biomarker detection, and prognosis prediction. Together, these results underscore nnMIL's potential as an effective computational pathology tool in a wide range of clinical applications under real-world data variability.

One important advantage of nnMIL is the ability to estimate uncertainty for slide-level prediction. Traditional MIL methods generally do not provide calibrated confidence estimates for slide-level predictions, which may hinder their utility for clinical decision support. nnMIL addresses this issue by computing uncertainty estimates through ensemble predictions. We show that model uncertainty estimates align closely with empirical prediction reliability, as excluding samples with the highest uncertainty scores substantially improves performance on the remaining samples\cite{de2018clinically,campanella2025real}. This approach could help identify cases that may benefit from additional pathological review or molecular testing, enabling more targeted expert review and follow-up testing. Furthermore, we show that patients with higher uncertainty scores, i.e., higher variability in ensemble predictions, tend to have worse prognosis, supporting the possibility that prediction uncertainty may capture aspects of intra-tumor morphological heterogeneity associated with aggressive biology and poor outcomes\cite{sali2024morphological}.

Although nnMIL does not introduce a new network architecture or loss function, we demonstrate that systematically optimizing the training configuration of multiple instance learning, such as enabling larger batch sizes and principled sampling strategies, can substantially improve the performance of slide-level tasks in computational pathology. In this regard, the core design principle of nnMIL is general and can be applied to other attention-based MIL methods, as well as combined with task-specific loss functions (such as AUC loss and Kappa loss~\cite{de2018weighted,yuan2021large}) that are otherwise infeasible under a batch size of one. We further show that incorporating the nnMIL training strategy into existing MIL frameworks, including ABMIL and DSMIL, consistently enhances their performance, and that nnMIL can be readily applied to aggregate features from other patch-level foundation models for slide-level prediction, without being restricted to a specific feature representation (e.g., CONCH~\cite{lu2024visual}, \textbf{\hyperref[tab:conch_v1_5_mil_comparison]{Supplementary Table 27}}). However, as a slide-level aggregation method, the performance of nnMIL remains bounded by the ability of the pathology feature extractor; moreover, it does not explicitly model the spatial organization of the tumor microenvironment, and our evaluation is conducted primarily on publicly available, retrospective benchmark datasets. In future work, it will be important to address these limitations to improve technical performance and enhance translational impact, for instance, through joint optimization of the feature extractor and slide aggregator, spatially aware feature aggregation, integration of complementary modalities such as spatial transcriptomics\cite{shulman2026ai} and proteomics\cite{li2026ai,li2026cellular}, as well as prospective multi-center validation with rigorous calibration and uncertainty quantification.

In conclusion, we introduce a training-centric perspective and present nnMIL, a framework for multiple instance learning in computational pathology. By systematically optimizing the model configuration, nnMIL offers a simple and practical solution for making generalizable and uncertainty-aware predictions in clinically relevant tasks, paving the way for developing and deploying reliable AI models in real-world settings. 

\end{spacing}

\setcounter{figure}{0}
\setcounter{table}{0}

\begin{spacing}{1.35}
\Heading{Methods}\phantomsection\label{sec:methods}

\heading{Design principles}

\noindent
nnMIL is a rule-based, generalizable multiple instance learning (MIL) framework that connects patch-level foundation-model representations with robust slide-level clinical predictions. Its design principles are: (i) \emph{normalize the training interface} by converting variable-length bags into fixed-length sequences to enable large and balanced mini-batches;  (ii) \emph{regularize attention and preserve semantics} by computing attention in random feature subspaces during training while performing aggregation directly in the original embedding space; and (iii) \emph{stabilize and quantify predictions} via subspace-ensemble inference that naturally estimates the uncertainty of slide-level predictions. Following a similar philosophy to nnU-Net (no-new U-Net)\cite{isensee2021nnu} for image segmentation, we name our multiple instance learning approach nnMIL (no-new MIL).

\heading{Architecture}

\noindent
The nnMIL architecture adopts a modified attention-based aggregator\cite{ilse2018attention} that transforms variable-length bags of patch embeddings into fixed-size slide-level representations. Given a WSI containing $N$ patches, each represented by a $D$-dimensional feature vector $\mathbf{x}_i \in \mathbb{R}^D$ extracted from a pathology foundation model, nnMIL computes attention weights in a randomly sampled feature subspace during training. Specifically, a subset of feature dimensions $S \subset \{1,\ldots,D\}$ with $|S|=H$ is randomly sampled, where $H$ is the hidden dimension. The attention weight for patch $i$ is computed as:
\begin{equation}
\alpha_i =
\frac{
\exp\left(
\mathbf{w}^\top \left[
\tanh(V_S \mathbf{x}_{i,S}) \odot \sigma(U_S \mathbf{x}_{i,S})
\right]
\right)
}{
\sum_{j=1}^{N}
\exp\left(
\mathbf{w}^\top \left[
\tanh(V_S \mathbf{x}_{j,S}) \odot \sigma(U_S \mathbf{x}_{j,S})
\right]
\right)
}
\end{equation}
where $\mathbf{x}_{i,S} \in \mathbb{R}^{H}$ denotes the sub-vector of $\mathbf{x}_i$ indexed by $S$. $V_S, U_S \in \mathbb{R}^{H \times H}$ denote the submatrices obtained by selecting the columns indexed by $S$ from the full learnable projection matrices $V, U \in \mathbb{R}^{H \times D}$. $\mathbf{w} \in \mathbb{R}^{H}$ is a scoring vector, $\odot$ denotes element-wise multiplication, and $\tanh$ and $\sigma$ represent the hyperbolic tangent and sigmoid activation functions, respectively.

This gated attention formulation combines nonlinear transformations with multiplicative gating, enabling the model to learn discriminative patch-level importance scores. Importantly, although the attention weights are computed in a sampled feature subspace, slide-level aggregation is performed directly in the original foundation-model embedding space:
\begin{equation}
\mathbf{h} = \sum_{i=1}^{N} \alpha_i \mathbf{x}_i
\end{equation}
where $\mathbf{h} \in \mathbb{R}^{D}$ denotes the slide-level representation. The final prediction is produced by a linear prediction head:
\begin{equation}
\hat{y} = f_{\mathrm{head}}(\mathbf{h})
\end{equation}
where $f_{\mathrm{head}}$ denotes a linear projection layer that outputs class logits for classification tasks, continuous predictions for regression tasks, or risk scores for survival analysis. By default, nnMIL uses a hidden dimension of $H=256$ and a dropout rate of 0.25. These design choices balance model capacity and regularization while keeping the overall architecture minimal and readily transferable across diverse pathology foundation models.

\noindent
Compared with conventional attention-based MIL architectures such as ABMIL\cite{ilse2018attention}, nnMIL has two major distinctions. First, instead of computing attention from the full feature space in a single deterministic pass, nnMIL computes attention in randomly sampled feature subspaces during training. This stochastic subspace sampling acts as an implicit regularizer, stabilizing optimization and improving generalization without increasing model complexity. During inference, nnMIL replaces random sampling with a deterministic subspace-ensemble strategy: the full feature space is divided into multiple overlapping $H$-dimensional subspaces, and each subspace produces an independent slide-level prediction. The final prediction is obtained by aggregating predictions across subspaces, yielding an ensemble-like effect and enabling uncertainty estimation without requiring multiple trained models. Second, although attention is computed in sampled feature subspaces, aggregation is performed directly on the original $D$-dimensional foundation-model embeddings rather than on intermediate projected features. This design preserves the pretrained embedding space and minimizes unnecessary transformations of foundation-model representations.

\heading{Rule-based parameter design}

\noindent
For each dataset, nnMIL first computes a compact data fingerprint that summarizes key dataset characteristics, including the per-slide patch-count distribution (median, interquartile range, and 5th/95th percentiles), magnification, embedding dimension $D$, class prevalence for classification tasks, target range for regression tasks, and event/censoring rates with follow-up time distributions for survival analysis. Given this fingerprint, nnMIL derives a rule-based parameter configuration, including the fixed bag size $M$, attention hidden dimension $H$ (default 256), dropout rate (0.25), task-aware batch sampler and batch size, optimizer and learning-rate scheduler, and inference stride for feature-subspace coverage. To enable efficient large-batch training while handling variable-length bags, nnMIL introduces a two-level sampling strategy that converts variable-length bags into fixed-size sampled bags and computes attention in sampled feature subspaces.

\noindent \textbf{Bag size during bag-level sampling.} 
During training, each WSI is represented by a fixed bag size $M$ determined from the dataset-level distribution of patch counts. Specifically, $M$ is set to half of the median number of patches per slide in the training set. For WSIs with more than $M$ patches, a random subset of $M$ patches is selected; for WSIs with fewer than $M$ patches, zero-padding is applied. This ensures that all samples within a mini-batch have a uniform length, enabling efficient batch processing and stable gradient computation.

\noindent \textbf{Feature-subspace sampling for attention.} 
During the forward pass, the attention mechanism operates on a randomly sampled subset of $H$ dimensions from the full $D$-dimensional feature space. This feature-subspace sampling serves two purposes. First, it reduces the computational cost of attention computation during training while preserving access to the full foundation-model embedding space during aggregation. Second, it acts as a form of regularization, encouraging the model to learn attention patterns that are not overly dependent on specific feature dimensions.

\noindent \textbf{Task-aware mini-batch sampler.} 
To address class imbalance and improve data utilization across heterogeneous tasks, nnMIL employs task-aware batch samplers that encourage balanced representation within each mini-batch. For classification, the \texttt{BalancedBatchSampler} enforces approximately equal numbers of samples from each class and distributes any remainder in a round-robin manner. For regression, the \texttt{RegressionBatchSampler} divides the target variable into bins and samples across bins to improve coverage of the target range. For survival analysis, the sampler balances event status and follow-up time distributions across mini-batches. These task-aware samplers allow nnMIL to train with substantially larger effective batch sizes, typically 32 or higher, compared with conventional MIL methods that are commonly trained with a batch size of one. This leads to smoother gradient updates, better data efficiency, and improved generalization.

\noindent \textbf{Stride for sliding-window inference.} 
During inference, nnMIL uses a sliding-window strategy to comprehensively cover the feature space. The full $D$-dimensional feature space is covered by multiple overlapping chunks of size $H$, with each chunk processed independently. By default, the stride is set to $H/4$. Given stride $S$, the number of chunks is:
\begin{equation}
K = \left\lceil \frac{D-H}{S} \right\rceil + 1
\end{equation}
Predictions from all chunks are then aggregated to produce the final slide-level prediction, and the variability across chunks is used to estimate prediction uncertainty.

\heading{Uncertainty estimation}

\noindent
nnMIL provides uncertainty estimates for slide-level prediction through a sliding-window subspace-ensemble inference scheme\cite{lakshminarayanan2017simple,gal2016dropout}. During evaluation, the full feature space is covered by $K$ overlapping chunks of size $H$ with a default stride of $H/4$, where $H=256$. Each chunk is processed independently by the attention mechanism and prediction head, producing a set of chunk-level predictions $\{\hat{y}_1, \hat{y}_2, \ldots, \hat{y}_K\}$.

\noindent \textbf{Classification tasks.} 
For classification tasks, the final prediction is computed as the mean of the $K$ logits:
\begin{equation}
\bar{y} = \frac{1}{K} \sum_{k=1}^{K} \hat{y}_k
\end{equation}
The probability distribution for each chunk is obtained by applying the softmax function, $p_k = \mathrm{softmax}(\hat{y}_k)$, and the average probability distribution is:
\begin{equation}
\bar{p} = \frac{1}{K} \sum_{k=1}^{K} p_k
\end{equation}
The total predictive uncertainty is quantified as the entropy of the average prediction:
\begin{equation}
H_{\mathrm{total}} = H(\bar{p}) = -\sum_{c=1}^{C} \bar{p}_c \log(\bar{p}_c)
\end{equation}
where $C$ is the number of classes and $\bar{p}_c$ is the average predicted probability for class $c$. The aleatoric uncertainty is estimated as the average entropy across chunk-level predictions:
\begin{equation}
H_{\mathrm{aleatoric}} = \frac{1}{K} \sum_{k=1}^{K} H(p_k)
= -\frac{1}{K} \sum_{k=1}^{K} \sum_{c=1}^{C} p_{k,c} \log(p_{k,c})
\end{equation}
where $p_{k,c}$ is the predicted probability of class $c$ from chunk $k$. The epistemic uncertainty is estimated by the mutual information (MI), defined as the difference between total and aleatoric uncertainty:
\begin{equation}
\mathrm{MI} = H_{\mathrm{total}} - H_{\mathrm{aleatoric}}
= H(\bar{p}) - \frac{1}{K} \sum_{k=1}^{K} H(p_k)
\end{equation}
The mutual information captures the disagreement among predictions from different feature subspaces, with higher values indicating greater epistemic uncertainty. For binary and multi-class classification tasks, both MI and $H_{\mathrm{aleatoric}}$ can be computed and used for selective prediction or model interpretation. For simplicity, we used $H_{\mathrm{aleatoric}}$ as the uncertainty score for slide-level interpretability.

\noindent \textbf{Prognosis prediction tasks.} 
For survival analysis using Cox proportional hazards models, each feature subspace produces a risk score $\eta_k$. The final risk score can be obtained either by averaging the $K$ risk scores or by using log-mean-exp aggregation across the $K$ chunks:
\begin{equation}
\eta = \log\left(\frac{1}{K} \sum_{k=1}^{K} \exp(\eta_k)\right)
\end{equation}
where $\eta_k$ is the risk score, or log hazard ratio, predicted from chunk $k$. In practice, these two aggregation strategies yielded similar results. The primary uncertainty measure for prognosis prediction is the variance of risk scores across chunks:
\begin{equation}
\mathrm{Var}_{\mathrm{risk}} = \frac{1}{K} \sum_{k=1}^{K} (\eta_k - \bar{\eta})^2
\end{equation}
where $\bar{\eta} = \frac{1}{K} \sum_{k=1}^{K} \eta_k$ is the mean risk score. This variance directly measures the variability of risk predictions across feature subspaces and serves as an epistemic uncertainty estimate for survival prediction.

For clinical interpretability, we also computed uncertainty in predicted survival probabilities\cite{cox1972regression,kvamme2021continuous}. Given the baseline survival function $S_0(t)$ estimated from the training data, the survival probability at time $t$ for chunk $k$ is:
\begin{equation}
S_k(t) = S_0(t)^{\exp(\eta_k)}
\end{equation}
The uncertainty in survival probability prediction is then quantified as the standard deviation across chunks:
\begin{equation}
\mathrm{Unc}_{S(t)} = \sqrt{\frac{1}{K} \sum_{k=1}^{K} (S_k(t) - \bar{S}(t))^2}
\end{equation}
where $\bar{S}(t) = \frac{1}{K} \sum_{k=1}^{K} S_k(t)$ is the mean predicted survival probability. When patient-level predictions were aggregated from multiple WSIs, we adjusted the survival-probability uncertainty by the square root of the number of WSIs:
\begin{equation}
\mathrm{Unc}_{S(t)}^{\mathrm{adj}} = \frac{\mathrm{Unc}_{S(t)}}{\sqrt{n_{\mathrm{WSI}}}}
\end{equation}
This adjustment improves comparability of uncertainty estimates across patients with different numbers of WSIs. The risk-score variance $\mathrm{Var}_{\mathrm{risk}}$ measures prediction variability in the risk-score space, whereas $\mathrm{Unc}_{S(t)}$ provides a clinically interpretable measure of uncertainty in terms of survival probability. In this study, we used $\mathrm{Unc}_{S(t)}$ at the median follow-up time point of each dataset for model interpretation.

\heading{Training protocol}

\noindent
All models were trained using the AdamW optimizer\cite{loshchilov2017decoupled} with a learning rate of $3 \times 10^{-4}$ ($1 \times 10^{-4}$ for prognosis tasks) and weight decay of $10^{-4}$. The learning rate was scheduled using a cosine annealing schedule with a warmup\cite{loshchilov2016sgdr} period of 5 epochs. Models were trained for up to 100 epochs with early stopping\cite{prechelt2002early} based on validation performance (patience of 10 epochs), and we employed the latest epoch checkpoint for final evaluation. The default batch size was 32 for all tasks. We used cross-entropy loss for classification tasks, Cox loss for prognosis prediction, and mean squared error loss for regression tasks. All experiments used a fixed random seed of 42 for reproducibility. All parameters were automatically determined from the dataset and task. This rule-based configuration follows the design philosophy of nnU-Net\cite{isensee2021nnu}, where domain knowledge is distilled into explicit heuristic rules that guide model configuration. 

\heading{Feature extraction}

\noindent
All preprocessing and patch feature extraction were performed within the CLAM toolbox\cite{lu2021data}, which provides an integrated pipeline for tissue segmentation, patch sampling, and model-specific preprocessing. For each whole-slide image (WSI), tissue regions were automatically identified using Otsu thresholding followed by morphological closing and hole filling to remove artifacts and background. Non-overlapping tiles were then extracted from the detected tissue regions at 40$\times$, 20$\times$, or 10$\times$ magnification (\verb|~|0.25, \verb|~|0.50, or \verb|~|1.0 $\mu$m/pixel), corresponding to raw patch sizes of 1024$\times$1024, 512$\times$512, or 256$\times$256 pixels, respectively. Each patch was resized according to the input resolution required by the downstream foundation model and subsequently passed through the model's feature extractor to obtain patch-level embeddings. Four pathology foundation models were used: GigaPath\cite{xu2024whole} (1,536-dimensional embeddings), H0\cite{hoptimus0} (1,536-dimensional), UNI\cite{chen2024towards} (1,024-dimensional), and Virchow2\cite{zimmermann2024virchow2} (2,560-dimensional). The resulting embeddings and corresponding patch coordinates were stored in HDF5 format for downstream multiple instance learning.

\heading{Evaluation protocol}

\noindent
For disease classification and biomarker prediction tasks, models were evaluated using official data splits or widely used splits\cite{shao2025mil} from the respective datasets. Performance was assessed using balanced accuracy (BACC), area under the receiver operating characteristic curve (AUC), and Cohen's kappa coefficient, as appropriate for each task. For survival analysis, models were evaluated using 5-fold cross-validation or independent evaluation with patient-level stratification to ensure that all slides from the same patient were assigned to the same fold\cite{xiang2025vision}. Performance was assessed using the concordance index (C-Index) for survival discrimination. Kaplan-Meier curves and log-rank tests were used to assess risk stratification. For tasks with multiple evaluation sets (e.g., external validation cohorts), models were trained on development sets and evaluated independently on external test sets.

\heading{Baseline MIL methods}

\noindent
nnMIL was compared against seven existing MIL methods: ABMIL (Attention-based MIL)\cite{ilse2018attention}, CLAM (Clustering-constrained Attention Multiple Instance Learning)\cite{lu2021data}, DSMIL (Dual-stream MIL)\cite{li2021dual}, TransMIL (Transformer-based MIL)\cite{shao2021transmil}, DTFD-MIL (Double Tier Feature Distillation MIL)\cite{zhang2022dtfd}, WIKG-MIL (WSI is Knowledge Graph MIL)\cite{li2024dynamic}, and ILRA-MIL (Iterative low-rank attention)\cite{xiang2023exploring}. All methods were implemented based on publicly available codebases and trained with identical hyperparameters and data splits to ensure fair comparison\cite{ling2024agent}. For those baseline methods that typically use a batch size of one, we reported the results based on the default settings and further evaluated a modified version of the baselines with a simulated batch size of 32 (batch size of 1 and gradient accumulation with 32 steps). For ABMIL (with nnMIL) and DSMIL (with nnMIL) that support larger batch sizes, we used batch size 32 to match nnMIL's training configuration.

\heading{Benchmark datasets}

\noindent
We comprehensively evaluated and benchmarked nnMIL using over 35 datasets and nearly 40,000 unique whole-slide images (WSIs) across three major clinical tasks: disease classification and subtyping, molecular biomarker detection, and prognosis prediction.

\noindent
\textbf{\underline{BCCC} (disease subtyping).} This dataset comprises 1,832 WSIs from excision specimens of cutaneous basal cell carcinomas collected at the Department of Pathology, Sahlgrenska University Hospital\cite{yacob2023weakly}. 
It includes three classification tasks with binary, three-class, and five-class labels. 
The official split provides 1,435 WSIs (with 200 slides held out for validation) for training and 397 WSIs for testing. 
One low-quality WSI was excluded from the training set due to poor image quality. 

\noindent
\textbf{\underline{BRACS} (disease subtyping).} This breast carcinoma subtyping dataset contains seven categories, including normal tissue and six lesion subtypes: pathological benign, usual ductal hyperplasia, flat epithelial atypia, atypical ductal hyperplasia, ductal carcinoma in situ, and invasive carcinoma\cite{brancati2022bracs}. 
A total of 547 WSIs are divided into training, validation, and testing sets containing 395, 65, and 87 slides, respectively. 

\noindent
\textbf{\underline{EBRAINS} (disease subtyping).} This is a brain tumor subtyping benchmark dataset for computational pathology model evaluation\cite{roetzer2022digital}. 
It contains 2,319 WSIs annotated with both coarse-grained (12-class) and fine-grained (30-class) subtype labels. 
Following established works\cite{chen2024towards,ding2024multimodal,vaidya2025molecular}, the dataset is divided into 1,151 training, 595 validation, and 573 testing slides. 

\noindent
\textbf{\underline{IMP-CRC2024} (disease subtyping).} This dataset includes 5,333 H\&E-stained formalin-fixed paraffin-embedded (FFPE) colorectal biopsy and polypectomy WSIs obtained from the digital archive of the IMP Diagnostics laboratory, Portugal\cite{neto2024interpretable}. 
It provides three tumor grading categories: non-neoplastic lesions, low-grade lesions, and high-grade lesions. 
We followed the official split, using 4,433 slides for model development (with 887 slides held out for validation) and 900 slides for testing. 

\noindent
\textbf{\underline{PANDA} (disease subtyping).} This large-scale prostate cancer grading dataset contains 9,555 WSIs with seven ISUP grade categories from the Radboud University Medical Center and the Karolinska Institute\cite{bulten2022artificial}. 
For training and evaluation, we followed the widely adopted split comprising 7,647, 954, and 954 slides for training, validation, and testing, respectively.

\noindent
\textbf{\underline{BCNB} (biomarker detection).} This dataset consists of 1,058 H\&E-stained FFPE WSIs from core-needle biopsies of early breast cancer\cite{xu2021predicting}. 
Each case is annotated with the protein expression status for three established biomarkers: ER (WT: 227, MUT: 831, where WT means negative status and MUT is positive status), PR (WT: 268, MUT: 790), and HER2 (WT: 781, MUT: 277). 
The dataset was split in a label-stratified manner into training, validation, and testing sets with a 70:10:20 ratio, resulting in 740, 106, and 212 slides, respectively.

\noindent
\textbf{\underline{\textit{BRAF} and \textit{KRAS} in CRC} (biomarker detection).} To further evaluate the generalizability of nnMIL in clinically actionable biomarker prediction, we focused on two key biomarkers, \textit{BRAF} and \textit{KRAS} mutations, across four large colorectal cancer (CRC) cohorts. 
These included the Molecular and Cellular Oncology (MCO)\cite{ward2015mco,jonnagaddala2016integration} cohort (1,492 samples), the SURGEN\cite{myles2025surgen} cohort (two sub-cohorts with 799 samples), and the TCGA-CRC\cite{weinstein2013cancer} cohort (521 samples). Sample sizes vary slightly across biomarkers due to missing molecular labels or quality-control exclusions. The SURGEN dataset (primary CRC) represents a recently released real-world cohort and was used as the internal development set. 
Trained models were externally validated on the MCO (archival resection cases from 1994--2010) and TCGA-CRC (primary CRC with genomic characterization) cohorts.

\noindent
\textbf{\underline{\textit{IDH} in brain tumor} (biomarker detection).} To further assess the generalizability of nnMIL in clinically actionable biomarker prediction for brain tumors, we focused on the isocitrate dehydrogenase (IDH) mutation status across different tumor grades using three independent cohorts. 
These included the Medical University of Vienna (MUV-IDH)\cite{roetzer2022digital} dataset (872 WSIs from 794 patients), TCGA-LGG\cite{weinstein2013cancer} (842 WSIs from 493 patients), and TCGA-GBM\cite{weinstein2013cancer} (834 WSIs from 374 patients), comprising a total of 2,548 WSIs from 1,661 patients. 
We trained nnMIL on the MUV-IDH dataset and externally evaluated its performance on the TCGA-LGG and TCGA-GBM cohorts.

\noindent
\textbf{\underline{Aneuploidy regression in TCGA} (biomarker detection).} We applied nnMIL to predict the continuous Aneuploidy Score from routine histopathology images. 
Model performance was evaluated using the TCGA pan-cancer dataset\cite{taylor2018genomic}, which included 10,854 WSIs from 8,819 patients. 
Patients were split into training, validation, and testing sets following a 7:1:2 ratio. 
For this regression task, Pearson correlation was used as the primary evaluation metric.

\noindent
\textbf{\underline{Whole-genome doubling in TCGA} (biomarker detection).} We also applied nnMIL to classify whole-genome doubling (WGD) status from routine histopathology images\cite{taylor2018genomic}. 
Performance was evaluated using the TCGA pan-cancer dataset with the same data split as the Aneuploidy regression task. 
Consistent with prior studies\cite{taylor2018genomic}, we simplified this task to a binary classification of WGD-positive (samples with one or more WGD events) versus WGD-negative (no WGD).

\noindent
\textbf{\underline{Tumor mutational burden in TCGA} (biomarker detection).} We further used nnMIL to classify tumor mutational burden (TMB) from routine histopathology images\cite{taylor2018genomic}. 
Following clinical practice, we formulated this task as a binary prediction using the clinically established threshold of 10 mutations per megabase (mut/Mb). Performance was evaluated using the TCGA pan-cancer dataset with the same data split as the Aneuploidy regression task. 

\noindent
\textbf{\underline{Pan-cancer prognosis in TCGA} (prognosis prediction).} To evaluate the generalizability of nnMIL in survival prediction across diverse cancer types, we constructed a pan-cancer prognosis cohort using The Cancer Genome Atlas (TCGA) collection\cite{weinstein2013cancer}. This cohort includes 16 major cancer types: bladder urothelial carcinoma (BLCA, 368 patients, 117 events), breast invasive carcinoma (BRCA, 1,007 patients, 76 events), cervical squamous cell carcinoma and endocervical adenocarcinoma (CESC, 258 patients, 47 events), colorectal and rectal adenocarcinoma (COADREAD, 392 patients, 56 events), esophageal carcinoma (ESCA, 134 patients, 34 events), glioblastoma multiforme (GBM, 347 patients, 287 events), head and neck squamous cell carcinoma (HNSC, 378 patients, 103 events), low-grade glioma (LGG, 480 patients, 103 events), liver hepatocellular carcinoma (LIHC, 332 patients, 68 events), lung adenocarcinoma (LUAD, 419 patients, 101 events), lung squamous cell carcinoma (LUSC, 405 patients, 77 events), pancreatic adenocarcinoma (PAAD, 173 patients, 79 events), renal cell carcinoma (RCC, 716 patients, 131 events), skin cutaneous melanoma (SKCM, 350 patients, 144 events), stomach adenocarcinoma (STAD, 353 patients, 92 events), and uterine corpus endometrial carcinoma (UCEC, 490 patients, 52 events). After quality control and removal of duplicate or low-quality slides, the dataset comprised 7,927 WSIs from 6,602 unique patients with available DSS time and censoring information. The dataset was organized into five-fold cross-validation splits with patient-level stratification. In each fold, 20\% of patients were held out as the test set, and the remaining 80\% was further split into training (87.5\%, i.e., 70\% of the total) and validation (12.5\%, i.e., 10\% of the total), yielding a 7:1:2 train:validation:test ratio. Survival labels were defined using disease-specific survival time and censoring indicators, and model performance was evaluated using the concordance index (C-Index) and Kaplan--Meier risk stratification.

\noindent
\textbf{\underline{Prognostic generalization in colorectal cancer} (prognosis prediction).} To evaluate prognostic generalization across external cohorts, we collected four colorectal cancer cohorts with disease-specific survival (DSS) endpoints. 
The PLCO (Prostate, Lung, Colorectal and Ovarian Cancer Screening Trial)\cite{gohagan2000prostate} colorectal cohort served as the development set, comprising 1,279 WSIs from 661 patients with 151 events (22\% event rate). 
The PLCO dataset was divided into training (1,009 WSIs from 528 patients) and validation (270 WSIs from 133 patients) sets. 
Three independent external cohorts were used for evaluation: MCO ($n{=}1{,}278$ patients with 1,301 WSIs, 318 events)\cite{ward2015mco,jonnagaddala2016integration}, TCGA-CRC ($n{=}567$ patients with 596 WSIs, 75 events)\cite{weinstein2013cancer}, and SURGEN ($n{=}425$ patients with 425 WSIs, 95 events)\cite{myles2025surgen}. 
All models were trained on the PLCO training set and evaluated independently on each external cohort to assess cross-institutional generalizability. 
Survival labels were defined using observed DSS time and censoring indicators, and model performance was evaluated using the concordance index (C-Index), Kaplan--Meier risk stratification, and multivariable Cox regression adjusting for clinical covariates including age, sex, grade, and stage.

\noindent
\textbf{\underline{Prognostic generalization in non-small cell lung cancer} (prognosis prediction).} To evaluate prognostic generalization for lung cancer, we collected two lung cancer cohorts with DSS endpoints. 
The PLCO lung cohort\cite{gohagan2000prostate} served as the development set, comprising 1,455 WSIs from 470 patients with 231 progression events (49.1\% event rate). 
The dataset was divided into training (1,156 WSIs from 376 patients) and validation (299 WSIs from 94 patients) subsets. 
The NLST (National Lung Screening Trial)\cite{national2011national} cohort was used as an independent external test set, comprising 1,143 WSIs from 414 patients with 169 progression events (40.8\% event rate). 
All models were trained on the PLCO training set and independently evaluated on NLST to assess cross-institutional generalizability. 
Survival labels were defined using observed DSS time and censoring indicators, and performance was assessed using the concordance index (C-Index), Kaplan--Meier risk stratification, and multivariable Cox regression analysis adjusting for clinical covariates.

\noindent
\textbf{\underline{CAMELYON16\&17} (metastasis detection).} To evaluate the generalizability of breast cancer metastasis detection across multiple institutions, we collected the multi-institutional CAMELYON16 and CAMELYON17 datasets\cite{bejnordi2017diagnostic,bandi2018detection} with three classes (normal, micro and macro metastasis), where CAMELYON16 was used for model development with 399 WSIs from Radboud University Medical Center in Nijmegen (RUMC) and University Medical Center Utrecht (UMCU), and CAMELYON17 was employed for evaluation with 464 WSIs from  Radboud University Medical Center in Nijmegen (RUMC), Canisius-Wilhelmina Hospital in Nijmegen (CWZ), University Medical Center Utrecht (UMCU), Rijnstate Hospital in Arnhem (RST), and Laboratory of Pathology East-Netherlands in Hengelo (LPON). For fair comparison, we followed the widely-used data split and evaluation metric (WSI-level accuracy in different institutions) in existing works\cite{ling2025comprehensive,cai2025attrimil}.

\heading{Statistical analysis}

\noindent
All performance metrics were computed using standard implementations from the scikit-learn (version 1.0+) and lifelines (version 0.27+) libraries. For cross-validation experiments, metrics were computed per fold and then averaged across all folds, with the standard error of the mean (SEM) reported. For each independent test set, metrics were computed using 1,000 bootstrap iterations, and reported as mean values with standard deviation. For overall comparisons across multiple tasks or cohorts, metrics were reported as mean values with the standard error of the mean (SEM) across all tasks or cohorts. Statistical comparisons between methods were performed using two-sided Wilcoxon signed-rank tests.

\heading{Computing hardware and software}

\noindent
All experiments and analyses were conducted using Python (version 3.10.12) and PyTorch (version 2.6.0, CUDA 12.4) (\url{https://pytorch.org}), together with other open-source packages as required. 
Whole-slide image (WSI) processing was performed using OpenSlide (version 4.3.1), openslide-python (version 1.4.2), and CLAM\cite{lu2021data}, along with their respective dependencies. 
For linear probe evaluation, statistical analysis, and visualization, we used several public machine-learning libraries, including Scikit-learn (version 1.6.1), Scikit-survival (version 0.24.0), and Matplotlib (version 3.10.0). 
Four publicly available tile-level pathology foundation models were employed for feature extraction: UNI (\url{https://huggingface.co/MahmoodLab/UNI}), GigaPath (\url{https://huggingface.co/prov-gigapath/prov-gigapath}), Virchow2 (\url{https://huggingface.co/paige-ai/Virchow2}), and H-optimus-0 (\url{https://huggingface.co/bioptimus/H-optimus-0}). 
All training and inference experiments were executed on a high-performance computing cluster equipped with 8~$\times$~48~GB NVIDIA L40 GPUs. 
Training time varied according to dataset size and model complexity, ranging from approximately 0.5~hours for small datasets (for example, BRACS with 547 WSIs) to 5~hours for large-scale datasets (for example, TCGA pan-cancer with 10,854 WSIs). We further assessed the total inference time per whole-slide image (WSI), measured in seconds, defined as the combined time for patch-level feature extraction and slide-level MIL inference. All experiments were performed on the EBRAINS test cohort (573 WSIs) under identical hardware settings using an NVIDIA L40S GPU cluster with solid-state drive storage. Despite slight variations in slide aggregation across MIL methods, the total inference time for all methods is nearly identical, as the vast majority of the time is spent on patch-level feature extraction (see \textbf{\hyperref[tab:total_inference_time]{Supplementary Table 28}}).

\heading{Data availability}

\noindent
All histopathology images and clinical annotations used for model development and evaluation are publicly available from the following sources: 
BCCC (\url{https://datahub.aida.scilifelab.se/10.23698/aida/bccc}), 
BRACS (\url{https://www.bracs.icar.cnr.it/}), 
EBRAINS (\url{https://doi.org/10.25493/WQ48-ZGX}), 
IMP-CRC2024 (\url{https://rdm.inesctec.pt/dataset/nis-2023-008}), 
PANDA (\url{https://www.kaggle.com/c/prostate-cancer-grade-assessment/data}), 
BCNB (\url{https://bupt-ai-cz.github.io/BCNB}), 
MCO (\url{https://www.sredhconsortium.org/sredh-datasets/mco-study-whole-slide-image-dataset}), 
SURGEN (\url{https://www.ebi.ac.uk/biostudies/studies/S-BIAD1285}), 
PLCO (\url{https://cdas.cancer.gov/plco/}), 
NLST (\url{https://www.cancerimagingarchive.net/collection/nlst/}),  
TCGA (\url{https://portal.gdc.cancer.gov}), and CAMELYON16\&17 (\url{https://camelyon17.grand-challenge.org/}). 
All datasets are accessible to the research community under their respective data use agreements or institutional licenses.

\heading{Code availability}

\noindent
The implementation of this project, including training, inference, and evaluation pipelines as well as a usage tutorial, is publicly available at ({\url{https://github.com/Luoxd1996/nnMIL}}).

\heading{Competing interests}

\noindent The authors declare no competing interests.

\heading{Funding}

\noindent This study was supported by the Himalaya Foundation Faculty Scholarship (to R.L.).

\heading{Acknowledgements}

\noindent We acknowledge the BCCC\cite{yacob2023weakly}, BCNB\cite{xu2021predicting}, BRACS\cite{brancati2022bracs}, EBRAINS\cite{roetzer2022digital}, IMP-CRC2024\cite{neto2022imil4path,neto2024interpretable}, PANDA\cite{bulten2022artificial}, MCO\cite{ward2015mco,jonnagaddala2016integration}, SURGEN\cite{myles2025surgen}, NLST\cite{national2011national}, TCGA\cite{weinstein2013cancer},  CAMELYON16\cite{bejnordi2017diagnostic}, and CAMELYON17\cite{bandi2018detection} consortia for making their datasets publicly available. We further thank the research team\cite{taylor2018genomic} for providing aneuploidy scores and curating the whole-genome doubling and tumor mutational burden annotations.

\heading{Author Contributions}

\noindent {X.L. conceived and designed the study, developed the model, curated the datasets, conducted all experiments, performed statistical analysis, created visualizations, and drafted the manuscript. J.X. and Y.J. contributed to the conception of the study and provided advice on experimental design, visualization, and manuscript writing. R.L. contributed to the conception and design of the study, interpreted the results, revised the manuscript, acquired funding, and supervised the project. All authors reviewed and approved the final manuscript.}

\end{spacing}

\clearpage
\begin{nolinenumbers}
\Heading{References}

\vspace{2mm}

\begin{spacing}{0.9}
\bibliographystyle{naturemag}
\bibliography{citations}
\end{spacing}
\end{nolinenumbers}
\clearpage

\Heading{Extended Data}
\vspace{2mm}
\begin{figure}[hptb]
\centering
\phantomsection\label{edfig:1}
\includegraphics[width=\textwidth,height=0.72\textheight,keepaspectratio]{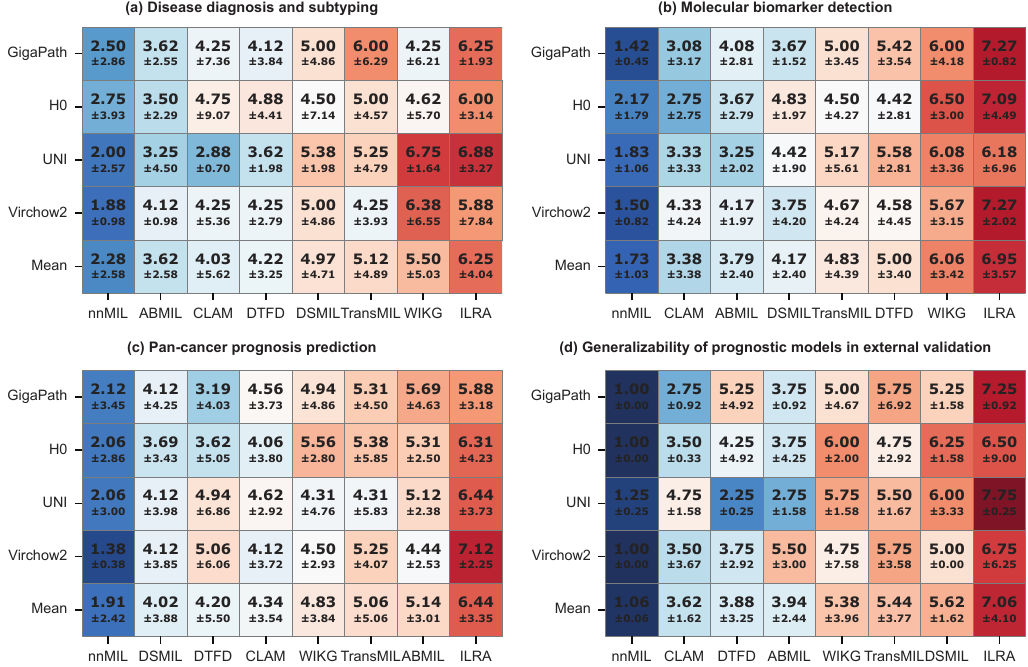}
\caption*{\textbf{Extended Data Figure 1 | Detailed ranking scores across disease diagnosis and subtyping, molecular biomarker detection, pan-cancer prognosis prediction and generalizability of prognostic models in four external cohorts.} \textbf{a,} Disease diagnosis and subtyping. \textbf{b,} Molecular biomarker detection. \textbf{c,} Pan-cancer prognosis prediction. \textbf{d,} Generalizability of prognostic models in external validation. All results are reported as mean $\pm$ standard deviation.}
\end{figure}

\begin{figure}[hptb]
\centering
\phantomsection\label{edfig:2}
\includegraphics[width=\textwidth,height=0.72\textheight,keepaspectratio]{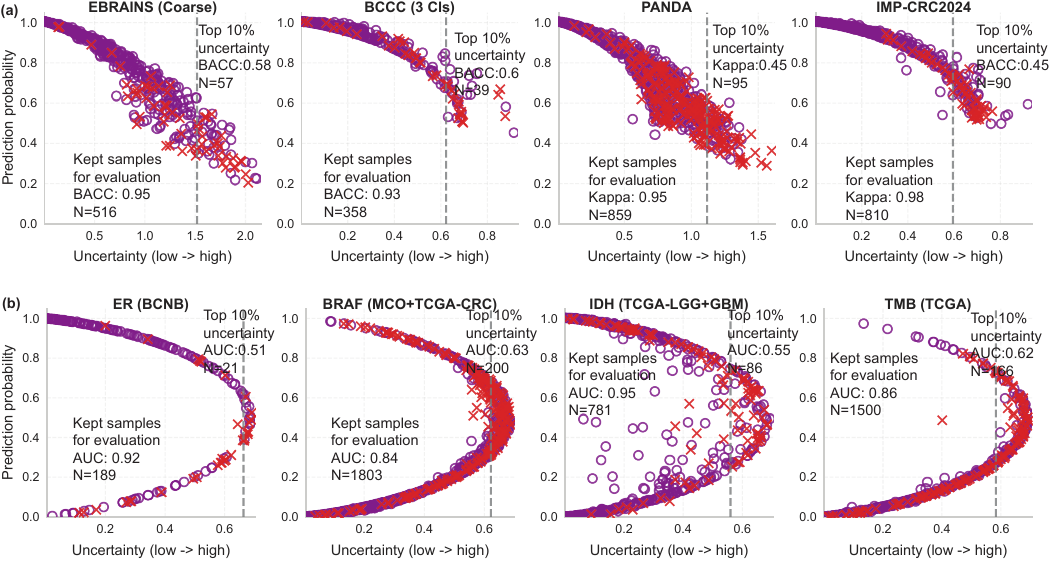}
\caption*{\textbf{Extended Data Figure 2 | Relationship between prediction score and estimated slide-level uncertainty.} \textbf{a,} Disease subtyping tasks, where uncertainty separates correct from incorrect predictions across EBRAINS, BCCC, PANDA and IMP-CRC2024. \textbf{b,} Molecular biomarker detection tasks, including ER, BRAF, IDH and TMB, showing that the relationship between uncertainty and prediction accuracy is preserved. Purple open circles indicate correctly classified cases; orange crosses denote misclassified samples.}
\end{figure}

\begin{figure}[hptb]
\centering
\phantomsection\label{edfig:3}
\includegraphics[width=\textwidth,height=0.72\textheight,keepaspectratio]{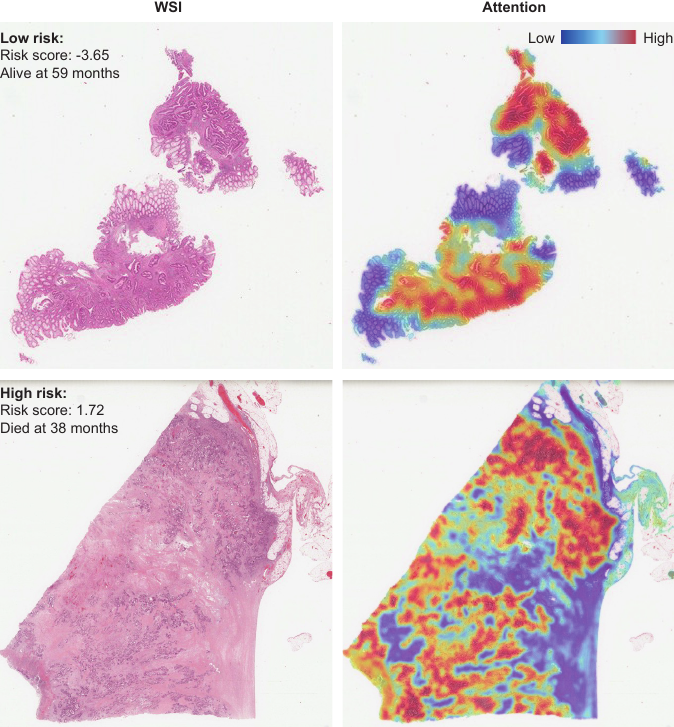}
\caption*{\textbf{Extended Data Figure 3 | Attention visualization of nnMIL aligned with pathological hallmarks.} Representative whole-slide images from the MCO cohort, with features extracted by Virchow2, and corresponding attention maps for low- and high-risk cases. High-attention regions predominantly localize to tumor-cell-rich areas and frequently co-localize with dense fibroblastic connective tissue, consistent with regions routinely emphasized by pathologists during diagnosis.}
\end{figure}

\begin{figure}[hptb]
\centering
\phantomsection\label{edfig:4}
\includegraphics[width=\textwidth,height=0.72\textheight,keepaspectratio]{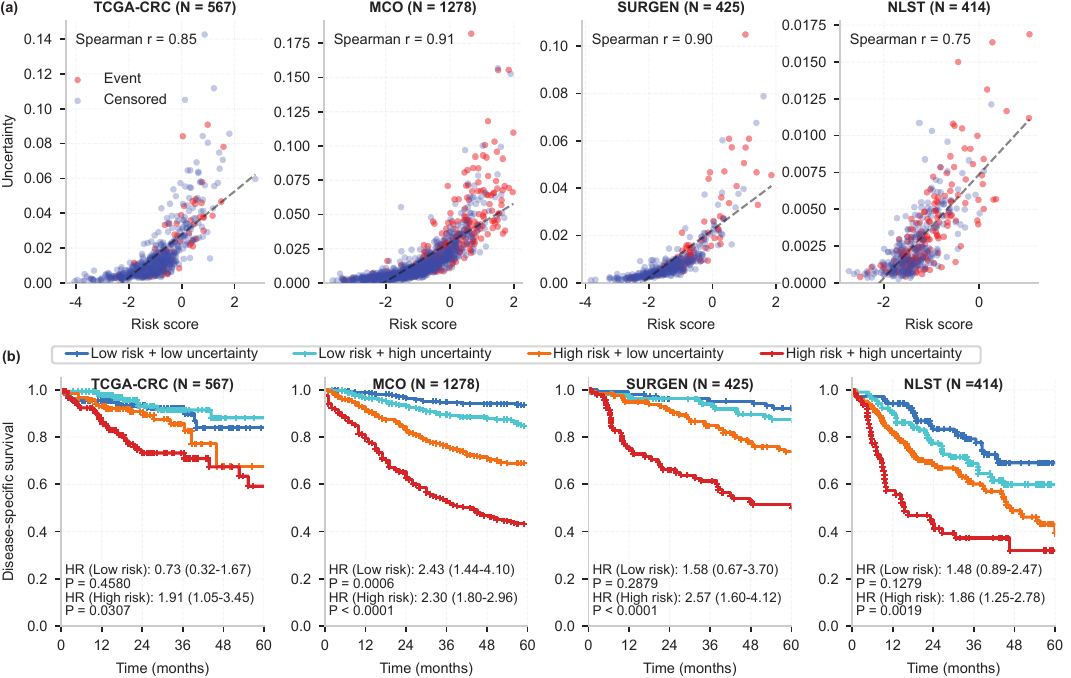}
\caption*{\textbf{Extended Data Figure 4 | Correlation between predicted risk and uncertainty in external cohorts and Kaplan--Meier survival analyses.} \textbf{a,} Correlation between predicted risk and estimated uncertainty across external cohorts. \textbf{b,} Kaplan--Meier survival analyses based on estimated uncertainty scores in both low- and high-risk groups in each external cohort. Survival differences were assessed using two-sided log-rank tests.}
\end{figure}

\begin{figure}[hptb]
\centering
\phantomsection\label{edfig:5}
\includegraphics[width=\textwidth,height=0.68\textheight,keepaspectratio]{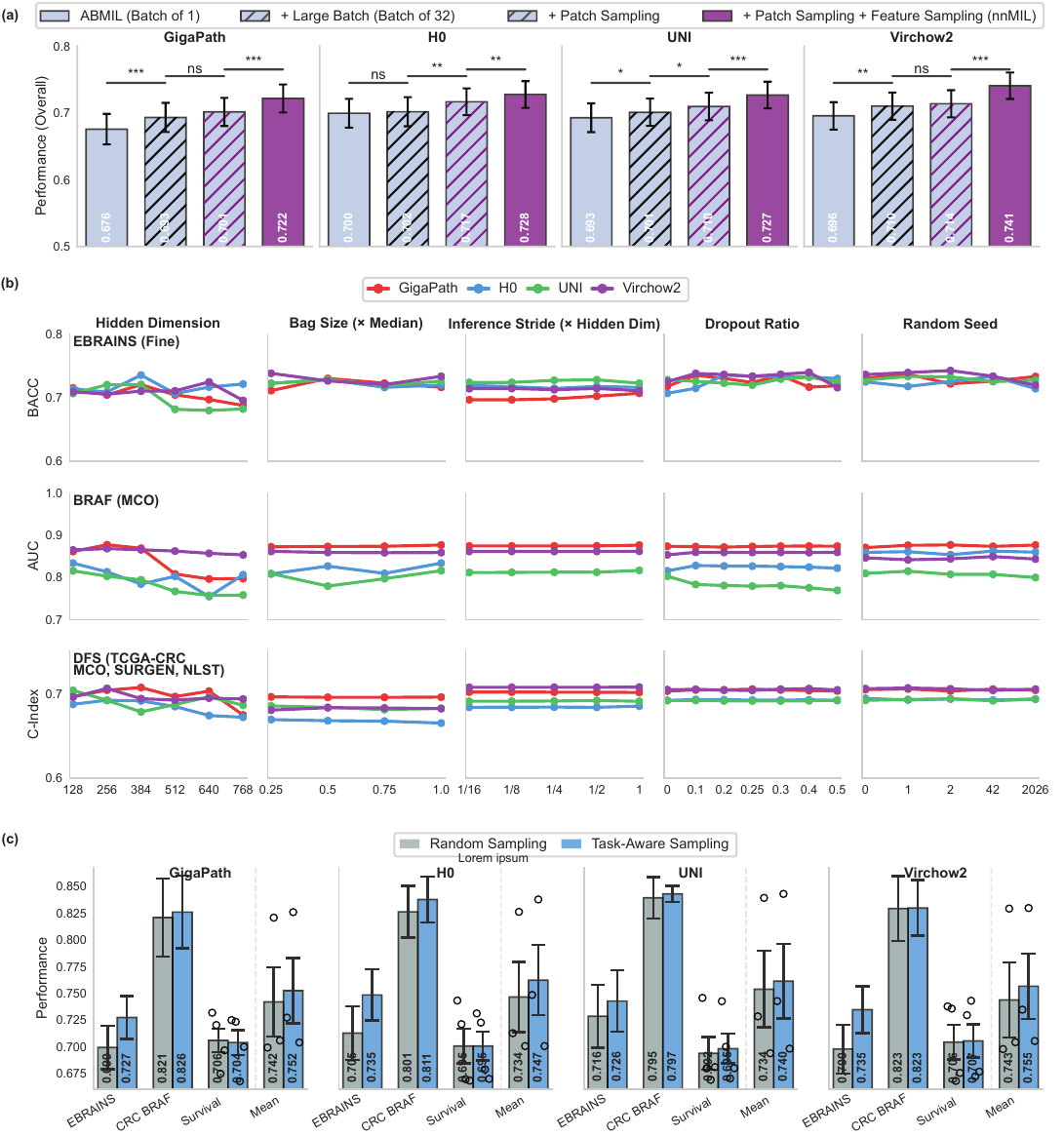}
\caption*{\textbf{Extended Data Figure 5 | Ablation study of the nnMIL framework.} \textbf{a,} Stepwise ablation of the nnMIL training strategy, progressively adding gradient accumulation (effective batch size 32), patch sampling and feature sampling to a baseline trained with a batch size of 1. Bars show means and error bars denote s.e.m. across 35 tasks (40 cohorts). Statistical significance was determined using a two-sided Wilcoxon signed-rank test: ns, not significant; *$P<0.05$; **$P<0.01$; ***$P<0.001$. \textbf{b,} Sensitivity analysis of nnMIL training hyperparameters. \textbf{c,} Ablation of two mini-batch sampling strategies. For EBRAINS and CRC BRAF, bars show means and error bars denote standard deviations from 1,000 bootstrap replicates. Survival and overall mean bars show means and s.e.m. across cohorts or tasks; each dot represents one task.}
\end{figure}

\clearpage
\Heading{Supplementary Information}

\setcounter{table}{0}
\renewcommand{\tablename}{Supplementary Table}
\begin{table}[htbp]
\centering
\caption{Performance comparison of different MIL methods on disease classification and subtyping tasks based on GigaPath. All results were evaluated using balanced accuracy (BACC), except for PANDA, which was assessed with Cohen's Kappa. {All results for individual datasets are reported as mean \ensuremath{\pm} standard deviation, estimated from 1,000 bootstrap replicates. Average results across datasets are reported as mean \ensuremath{\pm} standard error of the mean.} Statistical significance was determined using a two-sided Wilcoxon signed-rank test, where * indicates $P$ < 0.05, ** indicates $P$ < 0.01, and *** indicates $P$ < 0.001.}
\label{tab:diagnosis_gigapath}
\scalebox{0.72}{%
\begin{tabular}{lcccccccc}
\toprule
Dataset&CLAM&DTFD&DSMIL&ILRA&TransMIL&WIKG&ABMIL&nnMIL \\
\midrule
EBRAINS (Fine)&0.671$\pm$0.021&0.636$\pm$0.022&0.635$\pm$0.021&0.584$\pm$0.022&0.622$\pm$0.022&0.631$\pm$0.022&0.650$\pm$0.022&\textbf{0.709$\pm$0.021$^{***}$} \\
EBRAINS (Coarse)&0.793$\pm$0.021&0.800$\pm$0.026&0.840$\pm$0.027&0.724$\pm$0.028&0.803$\pm$0.027&0.790$\pm$0.026&0.855$\pm$0.024&\textbf{0.881$\pm$0.020$^{***}$} \\
PANDA&\textbf{0.942$\pm$0.009}&0.921$\pm$0.007&0.920$\pm$0.009&0.931$\pm$0.009&0.922$\pm$0.008&0.890$\pm$0.011&0.937$\pm$0.008&0.934$\pm$0.007 \\
IMP-CRC2024&\textbf{0.951$\pm$0.007}&0.947$\pm$0.008&0.862$\pm$0.013&0.944$\pm$0.008&0.941$\pm$0.008&0.948$\pm$0.008&0.947$\pm$0.008&0.950$\pm$0.007 \\
BCCC (2 Cls)&0.961$\pm$0.011&\textbf{0.976$\pm$0.008}&0.974$\pm$0.008&0.956$\pm$0.010&0.953$\pm$0.011&0.964$\pm$0.009&0.956$\pm$0.011&0.965$\pm$0.009 \\
BCCC (3 Cls)&0.765$\pm$0.019&0.897$\pm$0.016&0.837$\pm$0.018&0.881$\pm$0.016&\textbf{0.910$\pm$0.015}&0.885$\pm$0.016&0.883$\pm$0.017&0.872$\pm$0.017 \\
BCCC (5 Cls)&0.677$\pm$0.021&0.696$\pm$0.024&0.687$\pm$0.022&0.678$\pm$0.022&0.624$\pm$0.024&\textbf{0.734$\pm$0.024}&0.714$\pm$0.021&0.705$\pm$0.021 \\
BRACS&0.351$\pm$0.047&0.241$\pm$0.031&0.360$\pm$0.051&0.270$\pm$0.033&0.228$\pm$0.032&0.397$\pm$0.046&0.344$\pm$0.039&\textbf{0.441$\pm$0.055$^{***}$} \\
\midrule
Average&0.764$\pm$0.068&0.764$\pm$0.081&0.764$\pm$0.065&0.746$\pm$0.078&0.750$\pm$0.083&0.780$\pm$0.063&0.786$\pm$0.070&\textbf{0.807$\pm$0.059$^{***}$} \\
\bottomrule
\end{tabular}
}
\end{table}
\begin{table}[htbp]
\centering
\caption{Performance comparison of different MIL methods on disease classification and subtyping tasks based on H0. All results were evaluated using balanced accuracy (BACC), except for PANDA, which was assessed with Cohen's Kappa. {All results for individual datasets are reported as mean \ensuremath{\pm} standard deviation, estimated from 1,000 bootstrap replicates. Average results across datasets are reported as mean \ensuremath{\pm} standard error of the mean.} Statistical significance was determined using a two-sided Wilcoxon signed-rank test, where * indicates $P$ < 0.05, ** indicates $P$ < 0.01, and *** indicates $P$ < 0.001.}
\label{tab:diagnosis_h0}
\scalebox{0.72}{%
\begin{tabular}{lcccccccc}
\toprule
Dataset&CLAM&DTFD&DSMIL&ILRA&TransMIL&WIKG&ABMIL&nnMIL \\
\midrule
EBRAINS (Fine)&0.632$\pm$0.021&0.630$\pm$0.023&0.667$\pm$0.021&0.659$\pm$0.021&0.657$\pm$0.020&0.650$\pm$0.019&0.650$\pm$0.022&\textbf{0.718$\pm$0.020$^{***}$} \\
EBRAINS (Coarse)&0.672$\pm$0.029&0.791$\pm$0.025&0.806$\pm$0.025&0.795$\pm$0.028&0.735$\pm$0.023&0.839$\pm$0.023&0.841$\pm$0.024&\textbf{0.889$\pm$0.021$^{***}$} \\
PANDA&\textbf{0.947$\pm$0.006}&0.925$\pm$0.008&0.946$\pm$0.005&0.930$\pm$0.007&0.935$\pm$0.007&0.849$\pm$0.014&0.946$\pm$0.006&0.930$\pm$0.007 \\
IMP-CRC2024&0.957$\pm$0.007&0.947$\pm$0.008&0.896$\pm$0.012&0.920$\pm$0.010&0.859$\pm$0.013&0.899$\pm$0.010&0.942$\pm$0.009&\textbf{0.958$\pm$0.007$^{***}$} \\
BCCC (2 Cls)&0.962$\pm$0.010&0.974$\pm$0.008&0.972$\pm$0.008&0.949$\pm$0.010&0.968$\pm$0.008&\textbf{0.976$\pm$0.008}&0.976$\pm$0.008&0.973$\pm$0.008 \\
BCCC (3 Cls)&\textbf{0.896$\pm$0.016}&0.894$\pm$0.016&0.866$\pm$0.018&0.873$\pm$0.017&0.896$\pm$0.016&0.890$\pm$0.016&0.894$\pm$0.016&0.892$\pm$0.016 \\
BCCC (5 Cls)&0.742$\pm$0.020&0.722$\pm$0.023&0.685$\pm$0.024&0.665$\pm$0.025&0.737$\pm$0.020&\textbf{0.755$\pm$0.020}&0.742$\pm$0.022&0.752$\pm$0.020 \\
BRACS&0.324$\pm$0.044&0.407$\pm$0.043&\textbf{0.444$\pm$0.051}&0.330$\pm$0.046&0.392$\pm$0.050&0.384$\pm$0.045&0.394$\pm$0.049&0.434$\pm$0.055 \\
\midrule
Average&0.766$\pm$0.074&0.786$\pm$0.064&0.785$\pm$0.059&0.765$\pm$0.069&0.772$\pm$0.062&0.780$\pm$0.062&0.798$\pm$0.065&\textbf{0.818$\pm$0.060$^{***}$} \\
\bottomrule
\end{tabular}
}
\end{table}
\begin{table}[htbp]
\centering
\caption{Performance comparison of different MIL methods on disease classification and subtyping tasks based on UNI. All results were evaluated using balanced accuracy (BACC), except for PANDA, which was assessed with Cohen's Kappa. {All results for individual datasets are reported as mean \ensuremath{\pm} standard deviation, estimated from 1,000 bootstrap replicates. Average results across datasets are reported as mean \ensuremath{\pm} standard error of the mean.} Statistical significance was determined using a two-sided Wilcoxon signed-rank test, where * indicates $P$ < 0.05, ** indicates $P$ < 0.01, and *** indicates $P$ < 0.001.}
\label{tab:diagnosis_uni}
\scalebox{0.72}{%
\begin{tabular}{lcccccccc}
\toprule
Dataset&CLAM&DTFD&DSMIL&ILRA&TransMIL&WIKG&ABMIL&nnMIL \\
\midrule
EBRAINS (Fine)&0.667$\pm$0.021&0.644$\pm$0.022&0.627$\pm$0.020&0.605$\pm$0.022&0.635$\pm$0.022&0.614$\pm$0.022&0.656$\pm$0.020&\textbf{0.724$\pm$0.020$^{***}$} \\
EBRAINS (Coarse)&0.842$\pm$0.025&0.798$\pm$0.026&0.790$\pm$0.027&0.774$\pm$0.027&0.785$\pm$0.026&0.756$\pm$0.024&0.844$\pm$0.025&\textbf{0.913$\pm$0.019$^{***}$} \\
PANDA&0.937$\pm$0.008&0.928$\pm$0.007&0.922$\pm$0.009&0.925$\pm$0.010&0.920$\pm$0.008&0.902$\pm$0.010&\textbf{0.938$\pm$0.009}&0.924$\pm$0.008 \\
IMP-CRC2024&0.948$\pm$0.007&0.936$\pm$0.009&0.943$\pm$0.008&0.896$\pm$0.011&0.946$\pm$0.008&0.932$\pm$0.009&0.933$\pm$0.009&\textbf{0.950$\pm$0.007$^{***}$} \\
BCCC (2 Cls)&0.974$\pm$0.008&0.975$\pm$0.008&0.958$\pm$0.010&0.957$\pm$0.011&0.965$\pm$0.010&0.970$\pm$0.009&\textbf{0.976$\pm$0.008}&0.971$\pm$0.009 \\
BCCC (3 Cls)&0.885$\pm$0.016&0.893$\pm$0.016&0.869$\pm$0.017&0.796$\pm$0.019&\textbf{0.903$\pm$0.015}&0.874$\pm$0.016&0.872$\pm$0.017&0.896$\pm$0.016 \\
BCCC (5 Cls)&0.724$\pm$0.022&0.707$\pm$0.023&0.717$\pm$0.022&0.722$\pm$0.022&0.695$\pm$0.022&0.685$\pm$0.022&0.725$\pm$0.022&\textbf{0.736$\pm$0.021$^{***}$} \\
BRACS&0.382$\pm$0.046&0.414$\pm$0.053&0.407$\pm$0.049&0.303$\pm$0.036&0.336$\pm$0.043&0.363$\pm$0.048&0.366$\pm$0.041&\textbf{0.444$\pm$0.053$^{***}$} \\
\midrule
Average&0.795$\pm$0.066&0.787$\pm$0.063&0.779$\pm$0.063&0.747$\pm$0.071&0.773$\pm$0.071&0.762$\pm$0.068&0.789$\pm$0.067&\textbf{0.820$\pm$0.059$^{***}$} \\
\bottomrule
\end{tabular}
}
\end{table}
\begin{table}[htbp]
\centering
\caption{Performance comparison of different MIL methods on disease classification and subtyping tasks based on Virchow2. All results were evaluated using balanced accuracy (BACC), except for PANDA, which was assessed with Cohen's Kappa. {All results for individual datasets are reported as mean \ensuremath{\pm} standard deviation, estimated from 1,000 bootstrap replicates. Average results across datasets are reported as mean \ensuremath{\pm} standard error of the mean.} Statistical significance was determined using a two-sided Wilcoxon signed-rank test, where * indicates $P$ < 0.05, ** indicates $P$ < 0.01, and *** indicates $P$ < 0.001.}
\label{tab:diagnosis_virchow2}
\scalebox{0.72}{%
\begin{tabular}{lcccccccc}
\toprule
Dataset&CLAM&DTFD&DSMIL&ILRA&TransMIL&WIKG&ABMIL&nnMIL \\
\midrule
EBRAINS (Fine)&0.666$\pm$0.021&0.650$\pm$0.022&0.699$\pm$0.021&0.649$\pm$0.020&0.615$\pm$0.021&0.601$\pm$0.018&0.658$\pm$0.022&\textbf{0.720$\pm$0.021$^{***}$} \\
EBRAINS (Coarse)&0.777$\pm$0.028&0.804$\pm$0.025&0.808$\pm$0.022&0.860$\pm$0.023&0.821$\pm$0.024&0.746$\pm$0.028&0.846$\pm$0.021&\textbf{0.883$\pm$0.021$^{***}$} \\
PANDA&\textbf{0.956$\pm$0.006}&0.923$\pm$0.007&0.938$\pm$0.006&0.908$\pm$0.010&0.927$\pm$0.008&0.914$\pm$0.007&0.934$\pm$0.008&0.935$\pm$0.007 \\
IMP-CRC2024&0.932$\pm$0.009&0.945$\pm$0.008&0.944$\pm$0.008&0.930$\pm$0.009&0.952$\pm$0.007&0.929$\pm$0.008&0.941$\pm$0.008&\textbf{0.964$\pm$0.006$^{***}$} \\
BCCC (2 Cls)&0.974$\pm$0.008&0.969$\pm$0.009&0.941$\pm$0.011&\textbf{0.977$\pm$0.007}&0.969$\pm$0.009&0.964$\pm$0.010&0.967$\pm$0.009&0.975$\pm$0.008 \\
BCCC (3 Cls)&0.894$\pm$0.015&\textbf{0.909$\pm$0.016}&0.864$\pm$0.018&0.814$\pm$0.018&0.892$\pm$0.016&0.909$\pm$0.015&0.899$\pm$0.016&0.908$\pm$0.015 \\
BCCC (5 Cls)&\textbf{0.777$\pm$0.022}&0.747$\pm$0.023&0.722$\pm$0.022&0.712$\pm$0.024&0.777$\pm$0.022&0.652$\pm$0.023&0.748$\pm$0.022&0.754$\pm$0.019 \\
BRACS&0.332$\pm$0.048&0.400$\pm$0.046&0.341$\pm$0.042&0.255$\pm$0.029&0.368$\pm$0.047&0.383$\pm$0.038&0.399$\pm$0.048&\textbf{0.409$\pm$0.051$^{***}$} \\
\midrule
Average&0.788$\pm$0.070&0.793$\pm$0.064&0.782$\pm$0.067&0.763$\pm$0.077&0.790$\pm$0.068&0.762$\pm$0.068&0.799$\pm$0.064&\textbf{0.818$\pm$0.063$^{***}$} \\
\bottomrule
\end{tabular}
}
\end{table}
\begin{table}[htbp]
\centering
\caption{Performance comparison of different MIL methods on molecular biomarker prediction tasks based on GigaPath. All results were evaluated using the area under the curve (AUC), except for Aneuploidy, which was assessed with Pearson's correlation. {All results for individual datasets are reported as mean \ensuremath{\pm} standard deviation, estimated from 1,000 bootstrap replicates. Average results across datasets are reported as mean \ensuremath{\pm} standard error of the mean.} Statistical significance was determined using a two-sided Wilcoxon signed-rank test, where * indicates $P$ < 0.05, ** indicates $P$ < 0.01, and *** indicates $P$ < 0.001.}
\label{tab:biomarker_gigapath}
\scalebox{0.7}{%
\begin{tabular}{lcccccccc}
\toprule
Biomarker&CLAM&DTFD&DSMIL&ILRA&TransMIL&WIKG&ABMIL&nnMIL \\
\midrule
ER (BCNB)&0.862$\pm$0.032&0.819$\pm$0.038&0.854$\pm$0.031&0.833$\pm$0.033&0.847$\pm$0.033&0.859$\pm$0.029&0.889$\pm$0.026&\textbf{0.902$\pm$0.023$^{***}$} \\
HER2 (BCNB)&\textbf{0.749$\pm$0.041}&0.682$\pm$0.044&0.701$\pm$0.040&0.658$\pm$0.042&0.569$\pm$0.046&0.693$\pm$0.043&0.729$\pm$0.041&0.740$\pm$0.041 \\
PR (BCNB)&0.768$\pm$0.039&0.738$\pm$0.041&0.812$\pm$0.033&0.751$\pm$0.040&0.805$\pm$0.039&0.820$\pm$0.033&0.795$\pm$0.036&\textbf{0.827$\pm$0.035$^{***}$} \\
BRAF (MCO)&0.682$\pm$0.021&0.717$\pm$0.021&0.842$\pm$0.014&0.508$\pm$0.009&0.777$\pm$0.019&0.662$\pm$0.022&0.631$\pm$0.019&\textbf{0.875$\pm$0.013$^{***}$} \\
BRAF (TCGA-CRC)&0.732$\pm$0.040&0.645$\pm$0.044&0.775$\pm$0.036&0.501$\pm$0.022&0.755$\pm$0.040&0.772$\pm$0.031&0.743$\pm$0.037&\textbf{0.777$\pm$0.038} \\
KRAS (MCO)&\textbf{0.599$\pm$0.016}&0.575$\pm$0.016&0.595$\pm$0.016&0.500$\pm$0.000&0.533$\pm$0.015&0.497$\pm$0.016&0.558$\pm$0.017&0.583$\pm$0.016 \\
KRAS (TCGA-CRC)&0.620$\pm$0.024&0.606$\pm$0.026&0.612$\pm$0.025&0.502$\pm$0.005&0.563$\pm$0.025&0.468$\pm$0.026&0.603$\pm$0.025&\textbf{0.640$\pm$0.025$^{***}$} \\
IDH (TCGA-LGG)&0.831$\pm$0.027&0.797$\pm$0.029&0.804$\pm$0.027&0.539$\pm$0.027&\textbf{0.849$\pm$0.024}&0.786$\pm$0.027&0.824$\pm$0.026&0.840$\pm$0.026 \\
IDH (TCGA-GBM)&0.847$\pm$0.034&0.854$\pm$0.044&0.841$\pm$0.035&0.576$\pm$0.059&0.846$\pm$0.040&0.747$\pm$0.077&0.828$\pm$0.042&\textbf{0.902$\pm$0.028$^{***}$} \\
WGD (TCGA)&0.802$\pm$0.010&0.791$\pm$0.011&0.798$\pm$0.011&0.500$\pm$0.010&0.787$\pm$0.011&0.608$\pm$0.014&0.827$\pm$0.010&\textbf{0.831$\pm$0.010$^{***}$} \\
TMB (TCGA)&0.818$\pm$0.018&0.788$\pm$0.019&0.810$\pm$0.018&0.799$\pm$0.017&0.796$\pm$0.019&0.763$\pm$0.017&0.824$\pm$0.017&\textbf{0.844$\pm$0.016$^{***}$} \\
Aneuploidy (TCGA)&\textbf{0.585$\pm$0.017}&0.543$\pm$0.017&0.531$\pm$0.018&--&0.516$\pm$0.019&0.122$\pm$0.022&0.548$\pm$0.017&0.578$\pm$0.017 \\
\midrule
Average&0.741$\pm$0.027&0.713$\pm$0.028&0.748$\pm$0.031&0.606$\pm$0.038&0.720$\pm$0.037&0.650$\pm$0.057&0.733$\pm$0.033&\textbf{0.778$\pm$0.033$^{***}$} \\
\bottomrule
\end{tabular}
}
\end{table}
\begin{table}[htbp]
\centering
\caption{Performance comparison of different MIL methods on molecular biomarker prediction tasks based on H0. All results were evaluated using the area under the curve (AUC), except for Aneuploidy, which was assessed with Pearson's correlation. {All results for individual datasets are reported as mean \ensuremath{\pm} standard deviation, estimated from 1,000 bootstrap replicates. Average results across datasets are reported as mean \ensuremath{\pm} standard error of the mean.} Statistical significance was determined using a two-sided Wilcoxon signed-rank test, where * indicates $P$ < 0.05, ** indicates $P$ < 0.01, and *** indicates $P$ < 0.001.}
\label{tab:biomarker_h0}
\scalebox{0.7}{%
\begin{tabular}{lcccccccc}
\toprule
Biomarker&CLAM&DTFD&DSMIL&ILRA&TransMIL&WIKG&ABMIL&nnMIL \\
\midrule
ER (BCNB)&\textbf{0.901$\pm$0.028}&0.877$\pm$0.030&0.875$\pm$0.026&0.862$\pm$0.032&0.838$\pm$0.034&0.859$\pm$0.031&0.865$\pm$0.031&0.899$\pm$0.025 \\
HER2 (BCNB)&0.722$\pm$0.042&0.700$\pm$0.043&0.711$\pm$0.043&\textbf{0.767$\pm$0.037}&0.655$\pm$0.041&0.696$\pm$0.044&0.745$\pm$0.042&0.735$\pm$0.039 \\
PR (BCNB)&0.804$\pm$0.035&0.762$\pm$0.039&0.786$\pm$0.037&0.765$\pm$0.037&0.790$\pm$0.038&\textbf{0.831$\pm$0.035}&0.801$\pm$0.034&0.823$\pm$0.036 \\
BRAF (MCO)&\textbf{0.876$\pm$0.014}&0.771$\pm$0.017&0.750$\pm$0.018&0.520$\pm$0.015&0.832$\pm$0.016&0.708$\pm$0.021&0.876$\pm$0.014&0.804$\pm$0.016 \\
BRAF (TCGA-CRC)&\textbf{0.819$\pm$0.035}&0.736$\pm$0.038&0.780$\pm$0.033&0.527$\pm$0.031&0.790$\pm$0.033&0.686$\pm$0.037&0.807$\pm$0.034&0.769$\pm$0.038 \\
KRAS (MCO)&\textbf{0.657$\pm$0.015}&0.638$\pm$0.016&0.605$\pm$0.016&0.495$\pm$0.003&0.617$\pm$0.016&0.550$\pm$0.015&0.615$\pm$0.016&0.650$\pm$0.015 \\
KRAS (TCGA-CRC)&0.642$\pm$0.025&0.647$\pm$0.025&0.638$\pm$0.025&0.485$\pm$0.015&0.644$\pm$0.024&0.516$\pm$0.026&0.650$\pm$0.024&\textbf{0.668$\pm$0.024$^{***}$} \\
IDH (TCGA-LGG)&0.844$\pm$0.023&0.832$\pm$0.025&0.852$\pm$0.023&0.561$\pm$0.028&0.840$\pm$0.024&0.773$\pm$0.029&0.818$\pm$0.025&\textbf{0.868$\pm$0.023$^{***}$} \\
IDH (TCGA-GBM)&0.801$\pm$0.052&0.852$\pm$0.042&0.818$\pm$0.055&0.485$\pm$0.055&0.778$\pm$0.070&0.680$\pm$0.066&0.839$\pm$0.038&\textbf{0.869$\pm$0.031$^{***}$} \\
WGD (TCGA)&0.813$\pm$0.010&0.813$\pm$0.011&0.812$\pm$0.010&0.559$\pm$0.012&0.824$\pm$0.010&0.599$\pm$0.014&0.827$\pm$0.010&\textbf{0.836$\pm$0.010$^{***}$} \\
TMB (TCGA)&0.822$\pm$0.017&0.790$\pm$0.020&0.813$\pm$0.017&0.771$\pm$0.020&0.788$\pm$0.019&0.771$\pm$0.018&0.786$\pm$0.020&\textbf{0.851$\pm$0.015$^{***}$} \\
Aneuploidy (TCGA)&0.604$\pm$0.016&0.559$\pm$0.016&0.480$\pm$0.018&--&\textbf{0.608$\pm$0.016}&0.234$\pm$0.025&0.544$\pm$0.020&0.599$\pm$0.016 \\
\midrule
Average&0.775$\pm$0.027&0.748$\pm$0.027&0.743$\pm$0.032&0.618$\pm$0.041&0.750$\pm$0.025&0.658$\pm$0.047&0.764$\pm$0.029&\textbf{0.781$\pm$0.027$^{***}$} \\
\bottomrule
\end{tabular}
}
\end{table}
\begin{table}[htbp]
\centering
\caption{Performance comparison of different MIL methods on molecular biomarker prediction tasks based on UNI. All results were evaluated using the area under the curve (AUC), except for Aneuploidy, which was assessed with Pearson's correlation. {All results for individual datasets are reported as mean \ensuremath{\pm} standard deviation, estimated from 1,000 bootstrap replicates. Average results across datasets are reported as mean \ensuremath{\pm} standard error of the mean.} Statistical significance was determined using a two-sided Wilcoxon signed-rank test, where * indicates $P$ < 0.05, ** indicates $P$ < 0.01, and *** indicates $P$ < 0.001.}
\label{tab:biomarker_uni}
\scalebox{0.7}{%
\begin{tabular}{lcccccccc}
\toprule
Biomarker&CLAM&DTFD&DSMIL&ILRA&TransMIL&WIKG&ABMIL&nnMIL \\
\midrule
ER (BCNB)&0.874$\pm$0.026&0.859$\pm$0.033&0.876$\pm$0.026&0.884$\pm$0.029&0.842$\pm$0.034&0.867$\pm$0.026&0.868$\pm$0.030&\textbf{0.908$\pm$0.022$^{***}$} \\
HER2 (BCNB)&0.682$\pm$0.043&0.651$\pm$0.046&0.686$\pm$0.041&0.707$\pm$0.042&0.591$\pm$0.043&\textbf{0.721$\pm$0.043}&0.687$\pm$0.043&0.709$\pm$0.041 \\
PR (BCNB)&0.774$\pm$0.040&0.782$\pm$0.039&0.801$\pm$0.034&0.810$\pm$0.040&0.751$\pm$0.042&0.774$\pm$0.039&0.790$\pm$0.037&\textbf{0.830$\pm$0.033$^{***}$} \\
BRAF (MCO)&0.838$\pm$0.016&0.711$\pm$0.019&0.730$\pm$0.020&0.507$\pm$0.010&0.826$\pm$0.016&0.528$\pm$0.024&\textbf{0.847$\pm$0.015}&0.784$\pm$0.017 \\
BRAF (TCGA-CRC)&0.801$\pm$0.036&0.786$\pm$0.028&0.760$\pm$0.033&0.528$\pm$0.021&0.795$\pm$0.037&0.573$\pm$0.038&0.800$\pm$0.035&\textbf{0.823$\pm$0.031$^{***}$} \\
KRAS (MCO)&0.606$\pm$0.016&0.568$\pm$0.016&0.591$\pm$0.016&0.495$\pm$0.002&0.591$\pm$0.016&0.505$\pm$0.016&0.585$\pm$0.016&\textbf{0.620$\pm$0.016$^{***}$} \\
KRAS (TCGA-CRC)&\textbf{0.673$\pm$0.023}&0.586$\pm$0.025&0.629$\pm$0.024&0.495$\pm$0.009&0.598$\pm$0.024&0.534$\pm$0.025&0.623$\pm$0.026&0.661$\pm$0.024 \\
IDH (TCGA-LGG)&0.824$\pm$0.025&0.809$\pm$0.027&0.821$\pm$0.026&0.650$\pm$0.029&0.814$\pm$0.024&0.812$\pm$0.025&0.819$\pm$0.026&\textbf{0.846$\pm$0.024$^{***}$} \\
IDH (TCGA-GBM)&0.806$\pm$0.046&\textbf{0.885$\pm$0.028}&0.766$\pm$0.065&0.681$\pm$0.072&0.767$\pm$0.058&0.738$\pm$0.077&0.851$\pm$0.041&0.842$\pm$0.044 \\
WGD (TCGA)&0.794$\pm$0.012&0.795$\pm$0.011&0.804$\pm$0.011&0.515$\pm$0.010&0.796$\pm$0.011&0.687$\pm$0.013&0.818$\pm$0.010&\textbf{0.824$\pm$0.010$^{***}$} \\
TMB (TCGA)&0.812$\pm$0.017&0.780$\pm$0.019&0.795$\pm$0.019&0.798$\pm$0.018&0.762$\pm$0.021&0.810$\pm$0.016&\textbf{0.833$\pm$0.016}&0.827$\pm$0.017 \\
Aneuploidy (TCGA)&0.583$\pm$0.018&0.554$\pm$0.017&0.540$\pm$0.016&--&\textbf{0.605$\pm$0.016}&0.206$\pm$0.024&0.559$\pm$0.016&0.576$\pm$0.016 \\
\midrule
Average&0.756$\pm$0.026&0.730$\pm$0.032&0.733$\pm$0.028&0.643$\pm$0.041&0.728$\pm$0.028&0.646$\pm$0.051&0.757$\pm$0.031&\textbf{0.771$\pm$0.029$^{***}$} \\
\bottomrule
\end{tabular}
}
\end{table}
\begin{table}[htbp]
\centering
\caption{Performance comparison of different MIL methods on molecular biomarker prediction tasks based on Virchow2. All results were evaluated using the area under the curve (AUC), except for Aneuploidy, which was assessed with Pearson's correlation. {All results for individual datasets are reported as mean \ensuremath{\pm} standard deviation, estimated from 1,000 bootstrap replicates. Average results across datasets are reported as mean \ensuremath{\pm} standard error of the mean.} Statistical significance was determined using a two-sided Wilcoxon signed-rank test, where * indicates $P$ < 0.05, ** indicates $P$ < 0.01, and *** indicates $P$ < 0.001.}
\label{tab:biomarker_virchow2}
\scalebox{0.7}{%
\begin{tabular}{lcccccccc}
\toprule
Biomarker&CLAM&DTFD&DSMIL&ILRA&TransMIL&WIKG&ABMIL&nnMIL \\
\midrule
ER (BCNB)&0.863$\pm$0.027&0.887$\pm$0.028&0.881$\pm$0.024&0.876$\pm$0.026&0.886$\pm$0.026&0.882$\pm$0.026&0.885$\pm$0.029&\textbf{0.904$\pm$0.024$^{***}$} \\
HER2 (BCNB)&\textbf{0.781$\pm$0.036}&0.612$\pm$0.049&0.752$\pm$0.039&0.715$\pm$0.038&0.641$\pm$0.044&0.709$\pm$0.042&0.741$\pm$0.040&0.759$\pm$0.038 \\
PR (BCNB)&0.816$\pm$0.036&0.783$\pm$0.040&0.781$\pm$0.036&0.795$\pm$0.037&0.765$\pm$0.036&0.805$\pm$0.036&0.787$\pm$0.036&\textbf{0.846$\pm$0.032$^{***}$} \\
BRAF (MCO)&0.701$\pm$0.022&0.691$\pm$0.020&0.820$\pm$0.016&0.499$\pm$0.004&0.762$\pm$0.020&0.674$\pm$0.018&0.713$\pm$0.019&\textbf{0.866$\pm$0.014$^{***}$} \\
BRAF (TCGA-CRC)&0.781$\pm$0.037&0.691$\pm$0.042&\textbf{0.805$\pm$0.028}&0.507$\pm$0.021&0.729$\pm$0.036&0.697$\pm$0.037&0.782$\pm$0.032&0.797$\pm$0.039 \\
KRAS (MCO)&0.612$\pm$0.016&0.623$\pm$0.016&0.624$\pm$0.016&0.489$\pm$0.012&0.598$\pm$0.016&0.539$\pm$0.016&0.623$\pm$0.016&\textbf{0.646$\pm$0.015$^{***}$} \\
KRAS (TCGA-CRC)&0.616$\pm$0.025&\textbf{0.656$\pm$0.023}&0.636$\pm$0.025&0.488$\pm$0.022&0.558$\pm$0.026&0.514$\pm$0.026&0.634$\pm$0.024&0.623$\pm$0.025 \\
IDH (TCGA-LGG)&0.811$\pm$0.026&0.814$\pm$0.026&0.820$\pm$0.026&0.653$\pm$0.030&0.835$\pm$0.024&0.716$\pm$0.032&0.821$\pm$0.025&\textbf{0.849$\pm$0.023$^{***}$} \\
IDH (TCGA-GBM)&0.791$\pm$0.064&0.857$\pm$0.045&0.841$\pm$0.041&0.678$\pm$0.067&0.837$\pm$0.056&0.869$\pm$0.036&0.810$\pm$0.053&\textbf{0.882$\pm$0.033$^{***}$} \\
WGD (TCGA)&0.831$\pm$0.010&0.823$\pm$0.010&0.792$\pm$0.011&0.522$\pm$0.012&0.802$\pm$0.011&0.603$\pm$0.015&0.841$\pm$0.010&\textbf{0.845$\pm$0.009$^{***}$} \\
TMB (TCGA)&0.853$\pm$0.017&0.836$\pm$0.017&0.854$\pm$0.015&0.799$\pm$0.019&0.836$\pm$0.016&0.841$\pm$0.015&0.832$\pm$0.017&\textbf{0.876$\pm$0.013$^{***}$} \\
Aneuploidy (TCGA)&0.629$\pm$0.015&0.561$\pm$0.018&0.544$\pm$0.017&--&\textbf{0.637$\pm$0.016}&0.129$\pm$0.020&0.561$\pm$0.018&0.630$\pm$0.015 \\
\midrule
Average&0.757$\pm$0.026&0.736$\pm$0.030&0.762$\pm$0.029&0.638$\pm$0.042&0.741$\pm$0.030&0.665$\pm$0.057&0.753$\pm$0.028&\textbf{0.794$\pm$0.029$^{***}$} \\
\bottomrule
\end{tabular}
}
\end{table}
\begin{table}[htbp]
\centering
\caption{Performance comparison of different MIL methods on pan-cancer prognosis tasks based on GigaPath. All results were evaluated using the C-Index. {All results for each dataset are reported as mean \ensuremath{\pm} standard error of the mean based on five-fold cross-validation. Average results across datasets are reported as mean \ensuremath{\pm} standard error of the mean.} Statistical significance was determined using a two-sided Wilcoxon signed-rank test, where * indicates $P$ < 0.05, ** indicates $P$ < 0.01, and *** indicates $P$ < 0.001.}
\label{tab:prognosis_gigapath}
\scalebox{0.7}{%
\begin{tabular}{lcccccccc}
\toprule
Dataset&CLAM&DTFD&DSMIL&ILRA&TransMIL&WIKG&ABMIL&nnMIL \\
\midrule
BLCA&0.547$\pm$0.031&\textbf{0.571$\pm$0.024}&0.532$\pm$0.032&0.569$\pm$0.027&0.511$\pm$0.042&0.558$\pm$0.007&0.483$\pm$0.031&0.571$\pm$0.024 \\
BRCA&0.620$\pm$0.027&0.629$\pm$0.024&0.629$\pm$0.013&0.498$\pm$0.001&0.584$\pm$0.008&0.605$\pm$0.023&0.667$\pm$0.020&\textbf{0.719$\pm$0.015} \\
CESC&0.538$\pm$0.055&0.557$\pm$0.023&0.537$\pm$0.068&0.519$\pm$0.041&0.489$\pm$0.038&0.413$\pm$0.053&0.507$\pm$0.050&\textbf{0.591$\pm$0.042} \\
COADREAD&0.644$\pm$0.040&0.641$\pm$0.029&0.667$\pm$0.023&0.577$\pm$0.032&0.537$\pm$0.029&0.616$\pm$0.035&0.640$\pm$0.035&\textbf{0.692$\pm$0.024} \\
ESCA&0.656$\pm$0.052&0.649$\pm$0.054&0.634$\pm$0.035&0.636$\pm$0.046&0.556$\pm$0.046&0.576$\pm$0.033&0.527$\pm$0.056&\textbf{0.674$\pm$0.019} \\
GBM&0.560$\pm$0.010&0.560$\pm$0.017&0.540$\pm$0.024&0.558$\pm$0.021&0.527$\pm$0.024&0.548$\pm$0.021&0.527$\pm$0.015&\textbf{0.573$\pm$0.021} \\
HNSC&0.544$\pm$0.033&\textbf{0.594$\pm$0.022}&0.523$\pm$0.037&0.541$\pm$0.040&0.564$\pm$0.048&0.565$\pm$0.035&0.550$\pm$0.048&0.547$\pm$0.031 \\
LGG&0.712$\pm$0.022&0.689$\pm$0.017&0.709$\pm$0.029&0.557$\pm$0.036&0.643$\pm$0.033&0.687$\pm$0.025&0.720$\pm$0.018&\textbf{0.771$\pm$0.028} \\
LIHC&0.650$\pm$0.014&0.651$\pm$0.029&0.689$\pm$0.020&0.653$\pm$0.034&0.635$\pm$0.039&0.593$\pm$0.057&0.644$\pm$0.029&\textbf{0.726$\pm$0.020} \\
LUAD&0.491$\pm$0.031&0.591$\pm$0.027&0.534$\pm$0.018&0.549$\pm$0.032&0.579$\pm$0.035&\textbf{0.617$\pm$0.012}&0.536$\pm$0.020&0.565$\pm$0.026 \\
LUSC&0.480$\pm$0.056&0.505$\pm$0.028&\textbf{0.563$\pm$0.057}&0.498$\pm$0.020&0.509$\pm$0.045&0.527$\pm$0.050&0.498$\pm$0.030&0.559$\pm$0.040 \\
PAAD&0.549$\pm$0.048&\textbf{0.591$\pm$0.022}&0.548$\pm$0.046&0.542$\pm$0.025&0.569$\pm$0.025&0.554$\pm$0.027&0.534$\pm$0.018&0.536$\pm$0.024 \\
RCC&0.772$\pm$0.021&0.781$\pm$0.009&0.763$\pm$0.013&0.512$\pm$0.008&0.768$\pm$0.016&0.743$\pm$0.019&0.752$\pm$0.014&\textbf{0.813$\pm$0.010} \\
SKCM&0.569$\pm$0.024&0.544$\pm$0.024&0.586$\pm$0.034&0.563$\pm$0.020&0.591$\pm$0.031&0.569$\pm$0.043&0.596$\pm$0.027&\textbf{0.603$\pm$0.021} \\
STAD&0.554$\pm$0.022&0.575$\pm$0.010&\textbf{0.613$\pm$0.015}&0.587$\pm$0.032&0.605$\pm$0.033&0.606$\pm$0.038&0.554$\pm$0.030&0.597$\pm$0.024 \\
UCEC&0.656$\pm$0.047&0.715$\pm$0.046&0.685$\pm$0.033&0.534$\pm$0.022&0.643$\pm$0.060&0.616$\pm$0.036&0.639$\pm$0.061&\textbf{0.717$\pm$0.026} \\
\midrule
Average&0.596$\pm$0.019&0.615$\pm$0.017&0.609$\pm$0.018&0.556$\pm$0.011&0.582$\pm$0.017&0.587$\pm$0.017&0.586$\pm$0.020&\textbf{0.641$\pm$0.021$^{***}$} \\
\bottomrule
\end{tabular}
}
\end{table}
\begin{table}[htbp]
\centering
\caption{Performance comparison of different MIL methods on pan-cancer prognosis tasks based on H0. All results were evaluated using the C-Index. {All results for each dataset are reported as mean \ensuremath{\pm} standard error of the mean based on five-fold cross-validation. Average results across datasets are reported as mean \ensuremath{\pm} standard error of the mean.} Statistical significance was determined using a two-sided Wilcoxon signed-rank test, where * indicates $P$ < 0.05, ** indicates $P$ < 0.01, and *** indicates $P$ < 0.001.}
\label{tab:prognosis_h0}
\scalebox{0.7}{%
\begin{tabular}{lcccccccc}
\toprule
Dataset&CLAM&DTFD&DSMIL&ILRA&TransMIL&WIKG&ABMIL&nnMIL \\
\midrule
BLCA&0.537$\pm$0.024&\textbf{0.583$\pm$0.030}&0.525$\pm$0.019&0.527$\pm$0.023&0.519$\pm$0.039&0.528$\pm$0.036&0.535$\pm$0.030&0.552$\pm$0.021 \\
BRCA&0.627$\pm$0.034&0.666$\pm$0.034&0.659$\pm$0.031&0.495$\pm$0.010&0.597$\pm$0.034&0.627$\pm$0.044&0.653$\pm$0.038&\textbf{0.756$\pm$0.009} \\
CESC&0.532$\pm$0.060&0.585$\pm$0.022&0.575$\pm$0.069&0.468$\pm$0.023&0.538$\pm$0.034&0.461$\pm$0.064&0.548$\pm$0.059&\textbf{0.643$\pm$0.034} \\
COADREAD&\textbf{0.665$\pm$0.043}&0.619$\pm$0.017&0.626$\pm$0.038&0.574$\pm$0.037&0.565$\pm$0.033&0.613$\pm$0.032&0.649$\pm$0.048&0.661$\pm$0.056 \\
ESCA&0.679$\pm$0.030&0.632$\pm$0.056&\textbf{0.697$\pm$0.015}&0.649$\pm$0.028&0.605$\pm$0.053&0.655$\pm$0.051&0.629$\pm$0.039&0.644$\pm$0.038 \\
GBM&0.547$\pm$0.016&0.541$\pm$0.021&0.563$\pm$0.019&0.553$\pm$0.013&0.533$\pm$0.018&0.560$\pm$0.015&0.538$\pm$0.021&\textbf{0.589$\pm$0.018} \\
HNSC&0.601$\pm$0.025&0.589$\pm$0.050&0.580$\pm$0.034&0.551$\pm$0.018&0.614$\pm$0.040&0.587$\pm$0.030&0.571$\pm$0.030&\textbf{0.623$\pm$0.037} \\
LGG&0.756$\pm$0.026&0.653$\pm$0.011&0.751$\pm$0.040&0.588$\pm$0.050&0.676$\pm$0.032&0.708$\pm$0.014&0.746$\pm$0.026&\textbf{0.765$\pm$0.029} \\
LIHC&0.650$\pm$0.032&0.650$\pm$0.033&0.654$\pm$0.045&0.535$\pm$0.041&0.630$\pm$0.046&0.572$\pm$0.050&0.648$\pm$0.023&\textbf{0.727$\pm$0.031} \\
LUAD&0.561$\pm$0.027&0.558$\pm$0.020&\textbf{0.572$\pm$0.029}&0.497$\pm$0.042&0.491$\pm$0.027&0.570$\pm$0.042&0.557$\pm$0.023&0.534$\pm$0.015 \\
LUSC&0.488$\pm$0.032&\textbf{0.565$\pm$0.036}&0.544$\pm$0.048&0.548$\pm$0.018&0.524$\pm$0.043&0.509$\pm$0.058&0.509$\pm$0.047&0.552$\pm$0.047 \\
PAAD&0.587$\pm$0.008&\textbf{0.610$\pm$0.027}&0.564$\pm$0.014&0.503$\pm$0.018&0.566$\pm$0.024&0.532$\pm$0.022&0.584$\pm$0.030&0.593$\pm$0.022 \\
RCC&0.736$\pm$0.011&0.776$\pm$0.012&0.776$\pm$0.016&0.624$\pm$0.035&0.761$\pm$0.014&0.741$\pm$0.018&0.752$\pm$0.014&\textbf{0.809$\pm$0.015} \\
SKCM&0.580$\pm$0.024&0.620$\pm$0.030&0.570$\pm$0.024&0.578$\pm$0.028&0.591$\pm$0.020&0.570$\pm$0.018&0.568$\pm$0.029&\textbf{0.622$\pm$0.024} \\
STAD&0.616$\pm$0.029&0.518$\pm$0.006&0.634$\pm$0.018&0.636$\pm$0.035&\textbf{0.673$\pm$0.016}&0.604$\pm$0.019&0.585$\pm$0.033&0.614$\pm$0.023 \\
UCEC&0.643$\pm$0.037&0.714$\pm$0.031&0.692$\pm$0.027&0.539$\pm$0.025&0.718$\pm$0.028&0.631$\pm$0.037&0.708$\pm$0.044&\textbf{0.725$\pm$0.037} \\
\midrule
Average&0.613$\pm$0.018&0.617$\pm$0.016&0.624$\pm$0.018&0.554$\pm$0.013&0.600$\pm$0.018&0.592$\pm$0.017&0.611$\pm$0.018&\textbf{0.651$\pm$0.020$^{***}$} \\
\bottomrule
\end{tabular}
}
\end{table}
\begin{table}[htbp]
\centering
\caption{Performance comparison of different MIL methods on pan-cancer prognosis tasks based on UNI. All results were evaluated using the C-Index. {All results for each dataset are reported as mean \ensuremath{\pm} standard error of the mean based on five-fold cross-validation. Average results across datasets are reported as mean \ensuremath{\pm} standard error of the mean.} Statistical significance was determined using a two-sided Wilcoxon signed-rank test, where * indicates $P$ < 0.05, ** indicates $P$ < 0.01, and *** indicates $P$ < 0.001.}
\label{tab:prognosis_uni}
\scalebox{0.7}{%
\begin{tabular}{lcccccccc}
\toprule
Dataset&CLAM&DTFD&DSMIL&ILRA&TransMIL&WIKG&ABMIL&nnMIL \\
\midrule
BLCA&0.551$\pm$0.032&0.503$\pm$0.023&0.537$\pm$0.022&0.546$\pm$0.034&0.507$\pm$0.035&\textbf{0.563$\pm$0.043}&0.558$\pm$0.040&0.558$\pm$0.025 \\
BRCA&0.650$\pm$0.038&0.622$\pm$0.019&0.677$\pm$0.018&0.520$\pm$0.012&0.582$\pm$0.036&0.632$\pm$0.022&0.666$\pm$0.019&\textbf{0.792$\pm$0.015} \\
CESC&0.540$\pm$0.069&0.602$\pm$0.078&0.553$\pm$0.058&0.509$\pm$0.026&0.544$\pm$0.041&0.513$\pm$0.042&0.570$\pm$0.075&\textbf{0.655$\pm$0.027} \\
COADREAD&0.644$\pm$0.021&0.685$\pm$0.030&0.645$\pm$0.027&0.523$\pm$0.021&0.597$\pm$0.044&0.632$\pm$0.034&0.638$\pm$0.031&\textbf{0.691$\pm$0.015} \\
ESCA&0.630$\pm$0.042&0.576$\pm$0.038&0.665$\pm$0.055&0.656$\pm$0.036&0.639$\pm$0.036&0.642$\pm$0.063&0.602$\pm$0.033&\textbf{0.685$\pm$0.038} \\
GBM&0.547$\pm$0.025&0.519$\pm$0.015&\textbf{0.597$\pm$0.032}&0.567$\pm$0.016&0.530$\pm$0.025&0.589$\pm$0.023&0.547$\pm$0.016&0.577$\pm$0.024 \\
HNSC&0.601$\pm$0.037&0.605$\pm$0.035&0.581$\pm$0.028&0.521$\pm$0.016&0.608$\pm$0.042&0.576$\pm$0.040&0.588$\pm$0.035&\textbf{0.613$\pm$0.038} \\
LGG&0.724$\pm$0.022&0.663$\pm$0.016&0.727$\pm$0.018&0.550$\pm$0.037&0.697$\pm$0.020&0.719$\pm$0.028&0.700$\pm$0.038&\textbf{0.763$\pm$0.022} \\
LIHC&0.618$\pm$0.034&0.638$\pm$0.043&0.633$\pm$0.032&0.570$\pm$0.040&0.652$\pm$0.026&0.615$\pm$0.015&0.607$\pm$0.038&\textbf{0.742$\pm$0.020} \\
LUAD&\textbf{0.586$\pm$0.019}&0.541$\pm$0.031&0.577$\pm$0.021&0.542$\pm$0.024&0.552$\pm$0.020&0.585$\pm$0.023&0.538$\pm$0.031&0.575$\pm$0.022 \\
LUSC&0.553$\pm$0.036&0.479$\pm$0.042&0.513$\pm$0.034&0.522$\pm$0.032&0.525$\pm$0.042&0.479$\pm$0.024&0.495$\pm$0.040&\textbf{0.564$\pm$0.037} \\
PAAD&0.536$\pm$0.025&0.566$\pm$0.035&0.519$\pm$0.040&0.547$\pm$0.018&\textbf{0.575$\pm$0.007}&0.557$\pm$0.039&0.563$\pm$0.026&0.526$\pm$0.033 \\
RCC&0.780$\pm$0.016&0.791$\pm$0.010&0.784$\pm$0.015&0.587$\pm$0.030&0.708$\pm$0.019&0.743$\pm$0.013&0.763$\pm$0.008&\textbf{0.805$\pm$0.012} \\
SKCM&0.580$\pm$0.038&0.584$\pm$0.041&0.555$\pm$0.030&0.548$\pm$0.036&\textbf{0.602$\pm$0.032}&0.582$\pm$0.024&0.580$\pm$0.020&0.582$\pm$0.008 \\
STAD&0.595$\pm$0.029&0.583$\pm$0.025&0.606$\pm$0.033&0.618$\pm$0.031&\textbf{0.639$\pm$0.033}&0.626$\pm$0.025&0.598$\pm$0.013&0.626$\pm$0.026 \\
UCEC&0.628$\pm$0.034&0.652$\pm$0.057&0.651$\pm$0.036&0.551$\pm$0.019&0.673$\pm$0.049&0.695$\pm$0.050&0.648$\pm$0.042&\textbf{0.733$\pm$0.020} \\
\midrule
Average&0.610$\pm$0.016&0.601$\pm$0.019&0.614$\pm$0.018&0.555$\pm$0.009&0.602$\pm$0.015&0.609$\pm$0.017&0.604$\pm$0.016&\textbf{0.656$\pm$0.022$^{***}$} \\
\bottomrule
\end{tabular}
}
\end{table}
\begin{table}[htbp]
\centering
\caption{Performance comparison of different MIL methods on pan-cancer prognosis tasks based on Virchow2. All results were evaluated using the C-Index. {All results for each dataset are reported as mean \ensuremath{\pm} standard error of the mean based on five-fold cross-validation. Average results across datasets are reported as mean \ensuremath{\pm} standard error of the mean.} Statistical significance was determined using a two-sided Wilcoxon signed-rank test, where * indicates $P$ < 0.05, ** indicates $P$ < 0.01, and *** indicates $P$ < 0.001.}
\label{tab:prognosis_virchow2}
\scalebox{0.7}{%
\begin{tabular}{lcccccccc}
\toprule
Dataset&CLAM&DTFD&DSMIL&ILRA&TransMIL&WIKG&ABMIL&nnMIL \\
\midrule
BLCA&\textbf{0.602$\pm$0.022}&0.527$\pm$0.026&0.540$\pm$0.015&0.531$\pm$0.023&0.529$\pm$0.040&0.553$\pm$0.041&0.572$\pm$0.018&0.585$\pm$0.031 \\
BRCA&0.689$\pm$0.022&0.616$\pm$0.035&0.673$\pm$0.041&0.499$\pm$0.001&0.611$\pm$0.031&0.636$\pm$0.042&0.636$\pm$0.025&\textbf{0.757$\pm$0.009} \\
CESC&0.585$\pm$0.069&0.607$\pm$0.056&0.617$\pm$0.045&0.613$\pm$0.036&0.548$\pm$0.023&0.526$\pm$0.049&0.621$\pm$0.049&\textbf{0.679$\pm$0.028} \\
COADREAD&0.654$\pm$0.033&0.645$\pm$0.044&0.706$\pm$0.014&0.506$\pm$0.007&0.602$\pm$0.027&0.647$\pm$0.041&0.657$\pm$0.025&\textbf{0.726$\pm$0.022} \\
ESCA&0.570$\pm$0.041&0.632$\pm$0.058&0.591$\pm$0.049&0.601$\pm$0.039&0.595$\pm$0.034&0.607$\pm$0.072&0.593$\pm$0.030&\textbf{0.687$\pm$0.033} \\
GBM&0.543$\pm$0.023&0.522$\pm$0.019&0.553$\pm$0.024&0.503$\pm$0.012&0.562$\pm$0.020&0.585$\pm$0.012&0.547$\pm$0.015&\textbf{0.590$\pm$0.019} \\
HNSC&0.603$\pm$0.040&\textbf{0.633$\pm$0.035}&0.565$\pm$0.019&0.502$\pm$0.028&0.599$\pm$0.027&0.587$\pm$0.024&0.600$\pm$0.031&0.619$\pm$0.035 \\
LGG&0.694$\pm$0.023&0.701$\pm$0.018&0.701$\pm$0.018&0.507$\pm$0.006&0.613$\pm$0.045&0.711$\pm$0.028&0.699$\pm$0.028&\textbf{0.746$\pm$0.014} \\
LIHC&0.672$\pm$0.018&0.713$\pm$0.034&0.667$\pm$0.017&0.611$\pm$0.051&0.680$\pm$0.015&0.654$\pm$0.034&0.643$\pm$0.020&\textbf{0.727$\pm$0.022} \\
LUAD&0.585$\pm$0.013&0.511$\pm$0.017&0.532$\pm$0.026&0.505$\pm$0.021&0.539$\pm$0.023&0.562$\pm$0.029&0.550$\pm$0.023&\textbf{0.589$\pm$0.017} \\
LUSC&0.540$\pm$0.038&0.496$\pm$0.046&0.555$\pm$0.030&0.494$\pm$0.021&0.525$\pm$0.023&0.509$\pm$0.036&0.553$\pm$0.027&\textbf{0.591$\pm$0.041} \\
PAAD&0.581$\pm$0.045&0.534$\pm$0.026&0.565$\pm$0.030&0.576$\pm$0.024&\textbf{0.657$\pm$0.038}&0.578$\pm$0.016&0.568$\pm$0.020&0.580$\pm$0.027 \\
RCC&0.785$\pm$0.015&\textbf{0.816$\pm$0.009}&0.793$\pm$0.010&0.513$\pm$0.005&0.756$\pm$0.012&0.768$\pm$0.010&0.786$\pm$0.005&0.814$\pm$0.007 \\
SKCM&0.591$\pm$0.023&0.557$\pm$0.022&0.561$\pm$0.014&0.553$\pm$0.017&0.546$\pm$0.023&0.566$\pm$0.033&0.598$\pm$0.032&\textbf{0.619$\pm$0.016} \\
STAD&0.619$\pm$0.023&0.603$\pm$0.029&\textbf{0.661$\pm$0.020}&0.550$\pm$0.026&0.635$\pm$0.042&0.614$\pm$0.013&0.617$\pm$0.017&0.656$\pm$0.029 \\
UCEC&0.645$\pm$0.059&0.671$\pm$0.045&0.729$\pm$0.023&0.508$\pm$0.007&0.695$\pm$0.040&0.705$\pm$0.035&0.595$\pm$0.023&\textbf{0.758$\pm$0.037} \\
\midrule
Average&0.622$\pm$0.016&0.611$\pm$0.021&0.626$\pm$0.019&0.536$\pm$0.010&0.606$\pm$0.016&0.613$\pm$0.017&0.615$\pm$0.015&\textbf{0.670$\pm$0.019$^{***}$} \\
\bottomrule
\end{tabular}
}
\end{table}

\begin{table}[htbp]
    \centering
    \caption{Performance comparison of different MIL methods on generalization of breast cancer metastases detection (CAMELYON16 and 17 for development and evaluation, respectively). All results were evaluated using the official employed accuracy. {All results for individual datasets are reported as mean \ensuremath{\pm} standard deviation, estimated from 1,000 bootstrap replicates. Average results across datasets are reported as mean \ensuremath{\pm} standard error of the mean.}}
    \label{tab:camelyon_results_acc}
    \footnotesize 
    \scalebox{0.8}{%
    \begin{tabular}{lcccccccc}
    \toprule
    FM&CLAM&DTFD&DSMIL&ILRA&TransMIL&WIKG&ABMIL&nnMIL \\
     \midrule{GigaPath}&$0.821\pm0.228$&$0.825\pm0.211$&$0.817\pm0.238$&$0.910\pm0.055$&$0.837\pm0.230$&$0.786\pm0.239$&$0.843\pm0.181$&$\bm{0.916\pm0.040}$ \\
{H0}&$0.869\pm0.123$&$0.860\pm0.074$&$\bm{0.938\pm0.032}$&$0.846\pm0.136$&$0.880\pm0.117$&$0.812\pm0.241$&$0.822\pm0.175$&$0.871\pm0.114$ \\
{UNI}&$\bm{0.921\pm0.027}$&$0.820\pm0.211$&$0.903\pm0.055$&$0.597\pm0.200$&$0.907\pm0.039$&$0.813\pm0.228$&$0.912\pm0.033$&$0.906\pm0.039$ \\
    {Virchow2} &$0.940\pm0.032$&$0.928\pm0.025$&$0.931\pm0.026$&$\bm{0.944\pm0.022}$&$0.931\pm0.021$&$0.915\pm0.045$&$0.893\pm0.023$&$0.924\pm0.034$ \\
     \midrule
     \textbf{Average}&$0.887\pm0.046$&$0.858\pm0.043$&$0.897\pm0.048$&$0.824\pm0.135$&$0.888\pm0.035$&$0.831\pm0.049$&$0.867\pm0.036$&$\bm{0.904\pm0.021}$ \\
    \bottomrule
    \end{tabular}
    }
    \end{table}

\begin{table}[htbp]
    \centering
    \caption{Performance comparison of different MIL methods on generalization of breast cancer metastases detection across different institutions based on the GigaPath. All results were evaluated using the official employed accuracy. {All results for individual datasets are reported as mean \ensuremath{\pm} standard deviation, estimated from 1,000 bootstrap replicates. Average results across datasets are reported as mean \ensuremath{\pm} standard error of the mean.} RUMC: Radboud University Medical Center
in Nijmegen, CWZ: Canisius-Wilhelmina
Hospital in Nijmegen, UMCU: University Medical
Center Utrecht, RST: Rijnstate Hospital in
Arnhem (RST), and LPON: Laboratory of Pathology
East-Netherlands in Hengelo.}
    \label{tab:camelyon_results_acc_detailed}
    \footnotesize 
    \scalebox{0.8}{%
    \begin{tabular}{lcccccccc}
    \toprule
    Institution&CLAM&DTFD&DSMIL&ILRA&TransMIL&WIKG&ABMIL&nnMIL \\
     \midrule
    CWZ &$0.956\pm0.022$&$0.943\pm0.025$&$0.956\pm0.022$&$\bm{0.967\pm0.019}$&$0.954\pm0.022$&$0.878\pm0.034$&$0.944\pm0.025$&$0.944\pm0.024$ \\
    RST &$0.367\pm0.051$&$0.406\pm0.053$&$0.343\pm0.048$&$0.807\pm0.041$&$0.378\pm0.051$&$0.311\pm0.049$&$0.483\pm0.050$&$\bm{0.849\pm0.037}$ \\
    UMCU&$0.937\pm0.025$&$0.929\pm0.026$&$0.928\pm0.025$&$0.908\pm0.029$&$\bm{0.960\pm0.020}$&$0.939\pm0.024$&$0.919\pm0.028$&$0.939\pm0.024$ \\
    RUMC&$0.903\pm0.031$&$0.891\pm0.033$&$0.902\pm0.031$&$\bm{0.935\pm0.026}$&$0.934\pm0.025$&$0.880\pm0.034$&$0.913\pm0.028$&$0.890\pm0.032$ \\
    LPON&$0.944\pm0.025$&$0.956\pm0.020$&$0.957\pm0.022$&$0.935\pm0.025$&$0.957\pm0.021$&$0.924\pm0.027$&$\bm{0.957\pm0.021}$&$0.956\pm0.022$ \\
    \midrule
    \textbf{Average} &$0.821\pm0.102$&$0.825\pm0.094$&$0.817\pm0.106$&$0.910\pm0.025$&$0.837\pm0.103$&$0.786\pm0.107$&$0.843\pm0.081$&$\bm{0.916\pm0.018}$\\
    \bottomrule
    \end{tabular}
    }
    \end{table}
\begin{table}[htbp]
\centering
\caption{Performance comparison of different MIL methods on generalization prognosis tasks based on GigaPath. All results were evaluated using the C-Index. {All results for individual datasets are reported as mean \ensuremath{\pm} standard deviation, estimated from 1,000 bootstrap replicates. Average results across datasets are reported as mean \ensuremath{\pm} standard error of the mean.} Statistical significance was determined using a two-sided Wilcoxon signed-rank test, where * indicates $P$ < 0.05, ** indicates $P$ < 0.01, and *** indicates $P$ < 0.001.}
\label{tab:generalization_prognosis_gigapath}
\scalebox{0.68}{%
\begin{tabular}{lcccccccc}
\toprule
Dataset&CLAM&DTFD&DSMIL&ILRA&TransMIL&WIKG&ABMIL&nnMIL \\
\midrule
TCGA-CRC&0.637$\pm$0.034&0.631$\pm$0.040&0.610$\pm$0.031&0.600$\pm$0.030&0.654$\pm$0.031&0.633$\pm$0.032&0.647$\pm$0.032&\textbf{0.675$\pm$0.030$^{***}$} \\
MCO&0.670$\pm$0.016&0.597$\pm$0.016&0.587$\pm$0.016&0.522$\pm$0.007&0.543$\pm$0.017&0.535$\pm$0.017&0.647$\pm$0.016&\textbf{0.728$\pm$0.014$^{***}$} \\
SURGEN&0.678$\pm$0.028&0.578$\pm$0.030&0.668$\pm$0.031&0.606$\pm$0.028&0.606$\pm$0.031&0.695$\pm$0.028&0.632$\pm$0.031&\textbf{0.721$\pm$0.026$^{***}$} \\
NLST&0.661$\pm$0.023&0.658$\pm$0.024&0.641$\pm$0.021&0.615$\pm$0.022&0.590$\pm$0.023&0.637$\pm$0.022&0.647$\pm$0.024&\textbf{0.694$\pm$0.022$^{***}$} \\
\midrule
Average&0.662$\pm$0.008&0.616$\pm$0.015&0.627$\pm$0.015&0.586$\pm$0.019&0.598$\pm$0.020&0.625$\pm$0.029&0.643$\pm$0.003&\textbf{0.705$\pm$0.011$^{***}$} \\
\bottomrule
\end{tabular}
}
\end{table}
\begin{table}[htbp]
\centering
\caption{Performance comparison of different MIL methods on generalization prognosis tasks based on H0. All results were evaluated using the C-Index. {All results for individual datasets are reported as mean \ensuremath{\pm} standard deviation, estimated from 1,000 bootstrap replicates. Average results across datasets are reported as mean \ensuremath{\pm} standard error of the mean.} Statistical significance was determined using a two-sided Wilcoxon signed-rank test, where * indicates $P$ < 0.05, ** indicates $P$ < 0.01, and *** indicates $P$ < 0.001.}
\label{tab:generalization_prognosis_h0}
\scalebox{0.68}{%
\begin{tabular}{lcccccccc}
\toprule
Dataset&CLAM&DTFD&DSMIL&ILRA&TransMIL&WIKG&ABMIL&nnMIL \\
\midrule
TCGA-CRC&0.653$\pm$0.033&0.665$\pm$0.033&0.652$\pm$0.028&0.499$\pm$0.001&0.656$\pm$0.038&0.628$\pm$0.032&0.643$\pm$0.031&\textbf{0.667$\pm$0.032$^{***}$} \\
MCO&0.668$\pm$0.015&0.638$\pm$0.017&0.618$\pm$0.016&0.500$\pm$0.000&0.657$\pm$0.016&0.608$\pm$0.016&0.695$\pm$0.014&\textbf{0.728$\pm$0.013$^{***}$} \\
SURGEN&0.679$\pm$0.030&0.586$\pm$0.032&0.633$\pm$0.030&0.500$\pm$0.000&0.638$\pm$0.029&0.665$\pm$0.028&0.689$\pm$0.029&\textbf{0.716$\pm$0.026$^{***}$} \\
NLST&0.638$\pm$0.023&0.661$\pm$0.023&0.597$\pm$0.024&0.664$\pm$0.024&0.613$\pm$0.024&0.617$\pm$0.024&0.618$\pm$0.026&\textbf{0.668$\pm$0.022$^{***}$} \\
\midrule
Average&0.660$\pm$0.008&0.638$\pm$0.016&0.625$\pm$0.010&0.541$\pm$0.036&0.641$\pm$0.009&0.630$\pm$0.011&0.661$\pm$0.016&\textbf{0.695$\pm$0.014$^{***}$} \\
\bottomrule
\end{tabular}
}
\end{table}
\begin{table}[htbp]
\centering
\caption{Performance comparison of different MIL methods on generalization prognosis tasks based on UNI. All results were evaluated using the C-Index. {All results for individual datasets are reported as mean \ensuremath{\pm} standard deviation, estimated from 1,000 bootstrap replicates. Average results across datasets are reported as mean \ensuremath{\pm} standard error of the mean.} Statistical significance was determined using a two-sided Wilcoxon signed-rank test, where * indicates $P$ < 0.05, ** indicates $P$ < 0.01, and *** indicates $P$ < 0.001.}
\label{tab:generalization_prognosis_uni}
\scalebox{0.68}{%
\begin{tabular}{lcccccccc}
\toprule
Dataset&CLAM&DTFD&DSMIL&ILRA&TransMIL&WIKG&ABMIL&nnMIL \\
\midrule
TCGA-CRC&0.643$\pm$0.032&0.674$\pm$0.030&0.655$\pm$0.032&0.555$\pm$0.034&0.621$\pm$0.035&0.640$\pm$0.036&0.662$\pm$0.030&\textbf{0.680$\pm$0.031$^{***}$} \\
MCO&0.676$\pm$0.015&0.704$\pm$0.014&0.630$\pm$0.016&0.615$\pm$0.015&0.630$\pm$0.016&0.625$\pm$0.016&0.667$\pm$0.015&\textbf{0.727$\pm$0.014$^{***}$} \\
SURGEN&0.652$\pm$0.029&0.668$\pm$0.028&0.619$\pm$0.030&0.630$\pm$0.030&0.662$\pm$0.027&0.647$\pm$0.031&\textbf{0.690$\pm$0.027}&0.681$\pm$0.028 \\
NLST&0.613$\pm$0.023&0.670$\pm$0.022&0.608$\pm$0.023&0.505$\pm$0.003&0.614$\pm$0.023&0.624$\pm$0.024&0.648$\pm$0.023&\textbf{0.684$\pm$0.021$^{***}$} \\
\midrule
Average&0.646$\pm$0.011&0.679$\pm$0.007&0.628$\pm$0.009&0.576$\pm$0.025&0.631$\pm$0.009&0.634$\pm$0.005&0.667$\pm$0.007&\textbf{0.693$\pm$0.010$^{***}$} \\
\bottomrule
\end{tabular}
}
\end{table}
\begin{table}[htbp]
\centering
\caption{Performance comparison of different MIL methods on generalization prognosis tasks based on Virchow2. All results were evaluated using the C-Index. {All results for individual datasets are reported as mean \ensuremath{\pm} standard deviation, estimated from 1,000 bootstrap replicates. Average results across datasets are reported as mean \ensuremath{\pm} standard error of the mean.} Statistical significance was determined using a two-sided Wilcoxon signed-rank test, where * indicates $P$ < 0.05, ** indicates $P$ < 0.01, and *** indicates $P$ < 0.001.}
\label{tab:generalization_prognosis_virchow2}
\scalebox{0.68}{%
\begin{tabular}{lcccccccc}
\toprule
Dataset&CLAM&DTFD&DSMIL&ILRA&TransMIL&WIKG&ABMIL&nnMIL \\
\midrule
TCGA-CRC&0.657$\pm$0.033&0.612$\pm$0.033&0.620$\pm$0.035&0.498$\pm$0.001&0.607$\pm$0.041&0.638$\pm$0.035&0.635$\pm$0.033&\textbf{0.679$\pm$0.032$^{***}$} \\
MCO&0.661$\pm$0.016&0.703$\pm$0.015&0.655$\pm$0.016&0.503$\pm$0.003&0.671$\pm$0.016&0.642$\pm$0.016&0.639$\pm$0.016&\textbf{0.739$\pm$0.013$^{***}$} \\
SURGEN&0.661$\pm$0.028&0.689$\pm$0.030&0.667$\pm$0.028&0.500$\pm$0.000&0.649$\pm$0.030&0.694$\pm$0.026&0.674$\pm$0.026&\textbf{0.734$\pm$0.025$^{***}$} \\
NLST&0.652$\pm$0.022&0.646$\pm$0.023&0.634$\pm$0.023&0.650$\pm$0.022&0.630$\pm$0.023&0.620$\pm$0.024&0.627$\pm$0.024&\textbf{0.683$\pm$0.021$^{***}$} \\
\midrule
Average&0.658$\pm$0.002&0.663$\pm$0.018&0.644$\pm$0.009&0.538$\pm$0.032&0.639$\pm$0.012&0.649$\pm$0.014&0.644$\pm$0.009&\textbf{0.709$\pm$0.014$^{***}$} \\
\bottomrule
\end{tabular}
}
\end{table}

\begin{table}[htbp]
\centering
\caption{Patient characteristics in the non-small lung cancer cohorts. {Cohort comparisons used Pearson $\chi^2$ tests for categorical variables.}}
\label{tab:lung_patient_characteristics}
\begin{tabular}{lccr}
\toprule
Variable&PLCO (n = 470), No. (\%)&NLST (n = 414), No. (\%)&\textit{P} \\
\midrule
Age, years&&&.45 \\
  ~~~~~~~~$\leq$65&277 (59)&254 (61)&\\
  ~~~~~~~~>65&193 (41)&160 (39)&\\
Sex&&&.34 \\
  ~~~~~~~~Male&269 (57)&251 (61)&\\
  ~~~~~~~~Female&201 (43)&163 (39)&\\
Stage&&&.15 \\
  ~~~~~~~~I&305 (65)&273 (66)&\\
  ~~~~~~~~II&59 (13)&63 (15)&\\
  ~~~~~~~~III&67 (14)&57 (14)&\\
  ~~~~~~~~IV&38 (8)&19 (5)&\\
  ~~~~~~~~Missing&1 (0)&2 (0)&\\
Grade&&&<.0001 \\
  ~~~~~~~~1&61 (13)&24 (6)&\\
  ~~~~~~~~2&183 (39)&58 (14)&\\
  ~~~~~~~~3&167 (36)&168 (41)&\\
  ~~~~~~~~4&29 (6)&135 (33)&\\
  ~~~~~~~~Missing&30 (6)&29 (7)&\\
Survival status&&&.02 \\
  ~~~~~~~~Dead&239 (51)&245 (59)&\\
  ~~~~~~~~Alive or censored&231 (49)&169 (41)&\\
\bottomrule
\end{tabular}
\end{table}

\begin{table}[htbp]
\centering
\caption{Patient characteristics in the colorectal cancer cohorts. {Cohort comparisons used Pearson $\chi^2$ tests for categorical variables.}}
\label{tab:crc_patient_characteristics}
\scalebox{0.7}{
\begin{tabular}{lccccr}
\toprule
Variable&MCO (n = 1278), No. (\%)&PLCO (n = 661), No. (\%)&Surgen (n = 425), No. (\%)&TCGA (n = 567), No. (\%)&\textit{P} \\
\midrule
Age, years&&&&&<.0001 \\
  ~~~~~~~~$\leq$65&506 (40)&354 (54)&172 (40)&253 (45)&\\
  ~~~~~~~~>65&772 (60)&307 (46)&253 (60)&314 (55)&\\
Sex&&&&&<.001 \\
  ~~~~~~~~Male&710 (56)&383 (58)&196 (46)&295 (52)&\\
  ~~~~~~~~Female&568 (44)&278 (42)&229 (54)&272 (48)&\\
Stage&&&&&<.0001 \\
  ~~~~~~~~I&255 (20)&211 (32)&59 (14)&101 (18)&\\
  ~~~~~~~~II&450 (35)&181 (27)&152 (36)&202 (36)&\\
  ~~~~~~~~III&408 (32)&195 (30)&193 (45)&165 (29)&\\
  ~~~~~~~~IV&165 (13)&64 (10)&20 (5)&80 (14)&\\
  ~~~~~~~~Missing&0 (0)&10 (1)&1 (0)&19 (3)&\\
Grade&&&&&<.0001 \\
  ~~~~~~~~1&5 (0)&65 (10)&21 (5)&31 (5)&\\
  ~~~~~~~~2&1086 (85)&444 (67)&355 (84)&392 (69)&\\
  ~~~~~~~~3&184 (14)&90 (14)&47 (11)&100 (18)&\\
  ~~~~~~~~Missing&3 (0)&62 (9)&2 (0)&44 (8)&\\
Survival status&&&&&<.0001 \\
  ~~~~~~~~Dead&318 (25)&151 (23)&95 (22)&75 (13)&\\
  ~~~~~~~~Alive or censored&960 (75)&510 (77)&330 (78)&492 (87)&\\
\bottomrule
\end{tabular}}
\end{table}

\begin{table*}[htbp]
\centering
\caption{Comparison of single-task (ST, train a model for each task individually) and multiple-task (MT, train a model for all tasks jointly) performance of nnMIL across four foundation models (GigaPath, H0, UNI, Virchow2). Tasks include tumor grading (three classes), molecular biomarker detection (BRAF, KRAS), prognosis prediction on two external MCO and TCGA-CRC cohorts (the development cohort is SURGEN). {All results for individual datasets are reported as mean \ensuremath{\pm} standard deviation, estimated from 1,000 bootstrap replicates. Average results across datasets are reported as mean \ensuremath{\pm} standard error of the mean.}}
\label{tab:singletask_vs_multitask}
\scalebox{0.5}{%
\begin{tabular}{lcccccccccccc}
\toprule
\multirow{2}{*}{Cohort}&\multicolumn{3}{c}{GigaPath}&\multicolumn{3}{c}{H0}&\multicolumn{3}{c}{UNI}&\multicolumn{3}{c}{Virchow2} \\
\cmidrule(lr){2-4} \cmidrule(lr){5-7} \cmidrule(lr){8-10} \cmidrule(lr){11-13}
&ABMIL (MT)&nnMIL (ST)&nnMIL (MT)&ABMIL (MT)&nnMIL (ST)&nnMIL (MT)&ABMIL (MT)&nnMIL (ST)&nnMIL (MT)&ABMIL (MT)&nnMIL (ST)&nnMIL (MT) \\
\midrule
BRAF (MCO)&0.730$\pm$0.022&\textbf{0.859$\pm$0.016}&0.803$\pm$0.017&0.719$\pm$0.023&0.737$\pm$0.020&\textbf{0.738$\pm$0.023}&0.709$\pm$0.024&0.758$\pm$0.019&\textbf{0.768$\pm$0.021}&0.677$\pm$0.023&0.824$\pm$0.016&\textbf{0.830$\pm$0.016} \\
BRAF (TCGA-CRC)&0.751$\pm$0.038&\textbf{0.814$\pm$0.037}&0.806$\pm$0.033&0.759$\pm$0.037&0.756$\pm$0.038&\textbf{0.779$\pm$0.033}&\textbf{0.800$\pm$0.034}&0.799$\pm$0.039&0.778$\pm$0.034&0.703$\pm$0.038&0.773$\pm$0.041&\textbf{0.786$\pm$0.037} \\
KRAS (MCO)&0.577$\pm$0.017&0.563$\pm$0.017&\textbf{0.587$\pm$0.017}&0.599$\pm$0.017&\textbf{0.628$\pm$0.017}&0.624$\pm$0.017&0.620$\pm$0.017&0.611$\pm$0.017&\textbf{0.622$\pm$0.017}&0.577$\pm$0.017&\textbf{0.624$\pm$0.017}&0.619$\pm$0.016 \\
KRAS (TCGA-CRC)&0.610$\pm$0.026&0.638$\pm$0.027&\textbf{0.642$\pm$0.026}&0.606$\pm$0.027&\textbf{0.650$\pm$0.026}&0.612$\pm$0.027&\textbf{0.642$\pm$0.027}&0.621$\pm$0.027&0.630$\pm$0.026&\textbf{0.633$\pm$0.026}&0.590$\pm$0.028&0.618$\pm$0.026 \\
Grade (MCO)&0.776$\pm$0.011&\textbf{0.868$\pm$0.009}&0.861$\pm$0.010&\textbf{0.886$\pm$0.009}&0.794$\pm$0.011&0.873$\pm$0.010&0.639$\pm$0.014&0.818$\pm$0.011&\textbf{0.892$\pm$0.009}&\textbf{0.890$\pm$0.009}&0.875$\pm$0.009&0.887$\pm$0.009 \\
Grade (TCGA-CRC)&0.463$\pm$0.021&0.618$\pm$0.020&\textbf{0.679$\pm$0.021}&\textbf{0.773$\pm$0.018}&0.590$\pm$0.021&0.763$\pm$0.017&0.502$\pm$0.022&0.506$\pm$0.020&\textbf{0.709$\pm$0.019}&0.760$\pm$0.018&0.755$\pm$0.019&\textbf{0.779$\pm$0.017} \\
Survival (MCO)&0.588$\pm$0.013&\textbf{0.703$\pm$0.012}&0.699$\pm$0.011&0.671$\pm$0.013&\textbf{0.727$\pm$0.011}&0.719$\pm$0.012&0.688$\pm$0.012&\textbf{0.715$\pm$0.012}&0.706$\pm$0.012&0.679$\pm$0.012&\textbf{0.745$\pm$0.011}&0.740$\pm$0.010 \\
Survival (TCGA-CRC)&0.625$\pm$0.028&\textbf{0.682$\pm$0.025}&0.669$\pm$0.024&0.683$\pm$0.028&0.695$\pm$0.026&\textbf{0.695$\pm$0.027}&\textbf{0.707$\pm$0.023}&0.684$\pm$0.025&0.664$\pm$0.027&0.618$\pm$0.032&0.616$\pm$0.038&\textbf{0.682$\pm$0.030} \\
\midrule
\textbf{Overall}&0.640$\pm$0.037&0.718$\pm$0.041&\textbf{0.718$\pm$0.034}&0.712$\pm$0.034&0.697$\pm$0.024&\textbf{0.725$\pm$0.030}&0.663$\pm$0.030&0.689$\pm$0.037&\textbf{0.721$\pm$0.032}&0.692$\pm$0.035&0.725$\pm$0.037&\textbf{0.743$\pm$0.034} \\
\bottomrule
\end{tabular}}
\end{table*}

\begin{table}[htbp]
\centering
\caption{Performance comparison of different MIL methods with different training strategies (default setting with batch size of 1, a simulated batch size of 32 by gradient accumulation and our proposed nnMIL) across all 35 tasks (totally 40 cohorts) based on the GigaPath. {The results across multiple cohorts are reported as mean \ensuremath{\pm} standard error of the mean.} Statistical significance was determined using a two-sided Wilcoxon signed-rank test, where * indicates $P$ < 0.05, ** indicates $P$~<~0.01, *** indicates $P$ < 0.001.}
\label{tab:mil_batch1_vs_grad_accum_gigapath}
\scalebox{0.75}{%
\begin{tabular}{ccccccccc}
\toprule
\multirow{2}{*}{MIL Method}&\multicolumn{2}{c}{Diagnosis (8 cohorts)}&\multicolumn{2}{c}{Biomarker (12 cohorts)}&\multicolumn{2}{c}{Prognosis (20 cohorts)}&\multicolumn{2}{c}{Average (40 cohorts)} \\
\cmidrule(lr){2-3}\cmidrule(lr){4-5}\cmidrule(lr){6-7}\cmidrule(lr){8-9}
&Batch of 1&Batch of 32&Batch of 1&Batch of 32&Batch of 1&Batch of 32&Batch of 1&Batch of 32 \\
\midrule
CLAM&$0.764\pm0.068$&$0.795\pm0.071$&$0.741\pm0.027$&$0.758\pm0.030$&{$0.609\pm0.017$}&{$0.633\pm0.018$}&{$0.680\pm0.021$}&{$0.703\pm0.022$} \\
DTFD&$0.764\pm0.081$&$0.780\pm0.071$&$0.713\pm0.028$&$0.712\pm0.031$&$0.615\pm0.014$&$0.618\pm0.018$&$0.674\pm0.022$&$0.679\pm0.022$ \\
DSMIL&$0.764\pm0.065$&$0.790\pm0.065$&$0.748\pm0.031$&$0.752\pm0.032$&$0.613\pm0.015$&$0.638\pm0.017$&$0.684\pm0.021$&$0.703\pm0.021$ \\
ILRA&$0.746\pm0.078$&$0.781\pm0.068$&$0.606\pm0.036$&$0.716\pm0.033$&$0.562\pm0.010$&$0.598\pm0.016$&$0.612\pm0.023$&$0.670\pm0.022$ \\
TransMIL&$0.750\pm0.083$&$0.767\pm0.078$&$0.720\pm0.037$&$0.739\pm0.033$&$0.585\pm0.014$&$0.612\pm0.016$&$0.659\pm0.024$&$0.681\pm0.023$ \\
WIKG&$0.780\pm0.063$&$0.773\pm0.075$&$0.650\pm0.057$&$0.716\pm0.042$&$0.595\pm0.015$&$0.602\pm0.015$&$0.648\pm0.025$&$0.671\pm0.024$ \\
ABMIL&$0.786\pm0.070$&$0.806\pm0.061$&$0.733\pm0.033$&$0.739\pm0.035$&$0.597\pm0.017$&$0.621\pm0.017$&$0.676\pm0.023$&$0.693\pm0.022$ \\
nnMIL&\multicolumn{2}{c}{$\mathbf{0.807}\pm\mathbf{0.059}$}&\multicolumn{2}{c}{$\mathbf{0.778}\pm\mathbf{0.033}^{**}$}&\multicolumn{2}{c}{$\mathbf{0.654}\pm\mathbf{0.018}^{*}$}&\multicolumn{2}{c}{$\mathbf{0.722}\pm\mathbf{0.021}^{***}$} \\
\bottomrule
\end{tabular}
}
\end{table}

\begin{table}[htbp]
\centering
\caption{Performance comparison of different MIL methods with different training strategies (default setting with batch size of 1, a simulated batch size of 32 by gradient accumulation and our proposed nnMIL) across all 35 tasks (totally 40 cohorts) based on the H0. {The results across multiple cohorts are reported as mean \ensuremath{\pm} standard error of the mean.} Statistical significance was determined using a two-sided Wilcoxon signed-rank test, where * indicates $P$ < 0.05, ** indicates $P$~<~0.01, *** indicates $P$ < 0.001.}
\label{tab:mil_batch1_vs_grad_accum_h0}
\scalebox{0.75}{%
\begin{tabular}{ccccccccc}
\toprule
\multirow{2}{*}{MIL Method}&\multicolumn{2}{c}{Diagnosis (8 cohorts)}&\multicolumn{2}{c}{Biomarker (12 cohorts)}&\multicolumn{2}{c}{Prognosis (20 cohorts)}&\multicolumn{2}{c}{Average (40 cohorts)} \\
\cmidrule(lr){2-3}\cmidrule(lr){4-5}\cmidrule(lr){6-7}\cmidrule(lr){8-9}
&Batch of 1&Batch of 32&Batch of 1&Batch of 32&Batch of 1&Batch of 32&Batch of 1&Batch of 32 \\
\midrule
CLAM&{$0.767\pm0.074$}&{$0.807\pm0.064$}&{$0.775\pm0.027$}&{$\mathbf{0.784}\pm\mathbf{0.026}$}&{$0.622\pm0.015$}&{$0.639\pm0.017$}&{$0.697\pm0.022$}&{$0.715\pm0.021$} \\
DTFD&$0.786\pm0.064$&$0.774\pm0.072$&$0.748\pm0.027$&$0.725\pm0.029$&$0.621\pm0.013$&$0.638\pm0.015$&$0.692\pm0.020$&$0.691\pm0.020$ \\
DSMIL&$0.785\pm0.059$&$0.789\pm0.066$&$0.743\pm0.032$&$0.747\pm0.031$&$0.624\pm0.015$&$0.638\pm0.017$&$0.692\pm0.020$&$0.701\pm0.021$ \\
ILRA&$0.765\pm0.069$&$0.777\pm0.071$&$0.618\pm0.041$&$0.703\pm0.033$&$0.551\pm0.013$&$0.611\pm0.013$&$0.614\pm0.023$&$0.672\pm0.021$ \\
TransMIL&$0.772\pm0.062$&$0.773\pm0.079$&$0.750\pm0.025$&$0.737\pm0.031$&$0.608\pm0.015$&$0.619\pm0.016$&$0.684\pm0.020$&$0.685\pm0.023$ \\
WIKG&$0.780\pm0.062$&$0.780\pm0.073$&$0.658\pm0.047$&$0.741\pm0.033$&$0.599\pm0.015$&$0.619\pm0.016$&$0.653\pm0.023$&$0.688\pm0.022$ \\
ABMIL&$0.798\pm0.065$&$0.803\pm0.065$&$0.764\pm0.029$&$0.753\pm0.031$&$0.621\pm0.016$&$0.630\pm0.018$&$0.699\pm0.022$&$0.702\pm0.022$ \\
nnMIL&\multicolumn{2}{c}{$\mathbf{0.818}\pm\mathbf{0.060}$}&\multicolumn{2}{c}{$\mathbf{0.781}\pm\mathbf{0.027}$}&\multicolumn{2}{c}{$\mathbf{0.660}\pm\mathbf{0.017}^{*}$}&\multicolumn{2}{c}{$\mathbf{0.728}\pm\mathbf{0.020}^{***}$} \\
\bottomrule
\end{tabular}
}
\end{table}

\begin{table}[htbp]
\centering
\caption{Performance comparison of different MIL methods with different training strategies (default setting with batch size of 1, a simulated batch size of 32 by gradient accumulation and our proposed nnMIL) across all 35 tasks (totally 40 cohorts) based on the UNI. {The results across multiple cohorts are reported as mean \ensuremath{\pm} standard error of the mean.} Statistical significance was determined using a two-sided Wilcoxon signed-rank test, where * indicates $P$ < 0.05, ** indicates $P$~<~0.01, *** indicates $P$ < 0.001.}
\label{tab:mil_batch1_vs_grad_accum_uni}
\scalebox{0.75}{%
\begin{tabular}{ccccccccc}
\toprule
\multirow{2}{*}{MIL Method}&\multicolumn{2}{c}{Diagnosis (8 cohorts)}&\multicolumn{2}{c}{Biomarker (12 cohorts)}&\multicolumn{2}{c}{Prognosis (20 cohorts)}&\multicolumn{2}{c}{Average (40 cohorts)} \\
\cmidrule(lr){2-3}\cmidrule(lr){4-5}\cmidrule(lr){6-7}\cmidrule(lr){8-9}
&Batch of 1&Batch of 32&Batch of 1&Batch of 32&Batch of 1&Batch of 32&Batch of 1&Batch of 32 \\
\midrule
CLAM&{$0.795\pm0.066$}&{$0.805\pm0.060$}&{$0.756\pm0.026$}&{$0.760\pm0.027$}&{$0.617\pm0.014$}&{$0.643\pm0.015$}&{$0.694\pm0.021$}&{$0.710\pm0.019$}\\
DTFD&$0.787\pm0.063$&$0.779\pm0.069$&$0.731\pm0.032$&$0.744\pm0.028$&$0.616\pm0.017$&$0.637\pm0.018$&$0.685\pm0.021$&$0.697\pm0.021$ \\
DSMIL&$0.779\pm0.063$&$0.777\pm0.073$&$0.733\pm0.028$&$0.738\pm0.029$&$0.617\pm0.015$&$0.642\pm0.015$&$0.684\pm0.020$&$0.698\pm0.021$ \\
ILRA&$0.747\pm0.071$&$0.767\pm0.077$&$0.643\pm0.041$&$0.701\pm0.029$&$0.559\pm0.009$&$0.604\pm0.015$&$0.621\pm0.022$&$0.666\pm0.022$ \\
TransMIL&$0.773\pm0.071$&$0.791\pm0.069$&$0.728\pm0.028$&$0.722\pm0.027$&$0.608\pm0.012$&$0.614\pm0.014$&$0.677\pm0.021$&$0.682\pm0.021$ \\
WIKG&$0.762\pm0.068$&$0.781\pm0.074$&$0.646\pm0.051$&$0.732\pm0.029$&$0.614\pm0.014$&$0.623\pm0.014$&$0.653\pm0.023$&$0.687\pm0.021$ \\
ABMIL&$0.789\pm0.067$&$0.801\pm0.063$&$0.757\pm0.031$&$0.743\pm0.031$&$0.616\pm0.014$&$0.636\pm0.015$&$0.693\pm0.022$&$0.701\pm0.020$ \\
nnMIL&\multicolumn{2}{c}{$\mathbf{0.820}\pm\mathbf{0.059}$}&\multicolumn{2}{c}{$\mathbf{0.771}\pm\mathbf{0.029}$}&\multicolumn{2}{c}{$\mathbf{0.663}\pm\mathbf{0.018}^{*}$}&\multicolumn{2}{c}{$\mathbf{0.727}\pm\mathbf{0.020}^{**}$} \\
\bottomrule
\end{tabular}
}
\end{table}

\begin{table}[htbp]
\centering
\caption{Performance comparison of different MIL methods with different training strategies (default setting with batch size of 1, a simulated batch size of 32 by gradient accumulation and our proposed nnMIL) across all 35 tasks (totally 40 cohorts) based on the Virchow2. {The results across multiple cohorts are reported as mean \ensuremath{\pm} standard error of the mean.} Statistical significance was determined using a two-sided Wilcoxon signed-rank test, where * indicates $P$ < 0.05, ** indicates $P$~<~0.01, *** indicates $P$ < 0.001.}
\label{tab:mil_batch1_vs_grad_accum_virchow2}
\scalebox{0.75}{%
\begin{tabular}{ccccccccc}
\toprule
\multirow{2}{*}{MIL Method}&\multicolumn{2}{c}{Diagnosis (8 cohorts)}&\multicolumn{2}{c}{Biomarker (12 cohorts)}&\multicolumn{2}{c}{Prognosis (20 cohorts)}&\multicolumn{2}{c}{Average (40 cohorts)} \\
\cmidrule(lr){2-3}\cmidrule(lr){4-5}\cmidrule(lr){6-7}\cmidrule(lr){8-9}
&Batch of 1&Batch of 32&Batch of 1&Batch of 32&Batch of 1&Batch of 32&Batch of 1&Batch of 32 \\
\midrule
CLAM&{$0.788\pm0.070$}&{$0.809\pm0.064$}&{$0.757\pm0.026$}&{$0.772\pm0.026$}&{$0.629\pm0.013$}&{$0.649\pm0.016$}&{$0.700\pm0.021$}&{$0.718\pm0.020$} \\
DTFD&$0.793\pm0.064$&$0.805\pm0.065$&$0.736\pm0.030$&$0.743\pm0.028$&$0.622\pm0.018$&$0.643\pm0.015$&$0.690\pm0.021$&$0.706\pm0.020$ \\
DSMIL&$0.782\pm0.067$&$0.807\pm0.062$&$0.762\pm0.029$&$0.751\pm0.032$&$0.629\pm0.016$&$0.647\pm0.014$&$0.700\pm0.021$&$0.710\pm0.020$ \\
ILRA&$0.763\pm0.077$&$0.796\pm0.072$&$0.638\pm0.042$&$0.737\pm0.026$&$0.536\pm0.011$&$0.618\pm0.013$&$0.612\pm0.025$&$0.689\pm0.021$ \\
TransMIL&$0.790\pm0.068$&$0.788\pm0.072$&$0.741\pm0.030$&$0.738\pm0.030$&$0.612\pm0.013$&$0.629\pm0.013$&$0.686\pm0.021$&$0.694\pm0.021$ \\
WIKG&$0.762\pm0.068$&$0.800\pm0.068$&$0.665\pm0.057$&$0.751\pm0.031$&$0.620\pm0.014$&$0.629\pm0.013$&$0.662\pm0.025$&$0.699\pm0.021$ \\
ABMIL&$0.799\pm0.064$&$0.803\pm0.062$&$0.753\pm0.028$&$0.767\pm0.027$&$0.620\pm0.012$&$0.639\pm0.016$&$0.696\pm0.021$&$0.710\pm0.020$ \\
nnMIL&\multicolumn{2}{c}{$\mathbf{0.818}\pm\mathbf{0.063}$}&\multicolumn{2}{c}{$\mathbf{0.794}\pm\mathbf{0.029}^{**}$}&\multicolumn{2}{c}{$\mathbf{0.678}\pm\mathbf{0.015}^{***}$}&\multicolumn{2}{c}{$\mathbf{0.741}\pm\mathbf{0.020}^{***}$} \\
\bottomrule
\end{tabular}
}
\end{table}

\begin{table}[htbp]
\centering
\caption{Performance comparison of different MIL methods with different training strategies (default setting with batch size of 1 $vs$ nnMIL's training strategies) across all 35 tasks (totally 40 cohorts) based on GigaPath, H0, UNI and Virchow2, respectively. {The results across multiple cohorts are reported as mean \ensuremath{\pm} standard error of the mean.} Statistical significance was determined using a two-sided Wilcoxon signed-rank test, where * indicates $P$ < 0.05, ** indicates $P$~<~0.01, *** indicates $P$ < 0.001, \textit{Num} means the total numbers of cohorts, and \textbf{$\Delta$} means the performance gain between two different training strategies.}
\label{tab:mil_comparison_combined}
\scalebox{0.75}{%
\begin{tabular}{ll|ccc|ccc|c}
\toprule
FM&Task (\textit{Num})&ABMIL&ABMIL (with nnMIL)&\textbf{$\Delta$}&DS-MIL&DS-MIL (with nnMIL)&\textbf{$\Delta$}&nnMIL \\
\midrule
\multirow{4}{*}{GigaPath}&Diagnosis (8)&0.786$\pm$0.070&0.803$\pm$0.065&+0.017&0.765$\pm$0.065&0.778$\pm$0.077&+0.013&\textbf{0.807$\pm$0.059} \\
&Biomarker (12)&0.733$\pm$0.033&0.744$\pm$0.035&+0.012&0.748$\pm$0.031&0.765$\pm$0.033&+0.018&\textbf{0.778$\pm$0.033} \\
&Prognosis (20)&0.597$\pm$0.017&0.635$\pm$0.018&+0.038&0.613$\pm$0.015&0.637$\pm$0.015&+0.024&\textbf{0.654$\pm$0.018} \\
&Average (40)&0.676$\pm$0.023&0.701$\pm$0.021&+0.026&0.684$\pm$0.021&0.704$\pm$0.022&+0.020&\textbf{0.722$\pm$0.021$^{***}$} \\
\midrule
\multirow{4}{*}{H0}&Diagnosis (8)&0.798$\pm$0.065&0.813$\pm$0.063&+0.015&0.785$\pm$0.059&0.786$\pm$0.075&+0.000&\textbf{0.818$\pm$0.060} \\
&Biomarker (12)&0.764$\pm$0.029&0.769$\pm$0.030&+0.005&0.743$\pm$0.032&0.769$\pm$0.029&+0.026&\textbf{0.781$\pm$0.027} \\
&Prognosis (20)&0.621$\pm$0.016&0.646$\pm$0.016&+0.025&0.624$\pm$0.015&0.627$\pm$0.015&+0.003&\textbf{0.660$\pm$0.017} \\
&Average (40)&0.699$\pm$0.022&0.717$\pm$0.020&+0.017&0.692$\pm$0.020&0.701$\pm$0.022&+0.009&\textbf{0.728$\pm$0.020$^{**}$} \\
\midrule
\multirow{4}{*}{UNI}&Diagnosis (8)&0.789$\pm$0.067&0.800$\pm$0.071&+0.011&0.779$\pm$0.063&0.787$\pm$0.080&+0.007&\textbf{0.820$\pm$0.059} \\
&Biomarker (12)&0.757$\pm$0.031&0.752$\pm$0.031&-0.005&0.733$\pm$0.028&0.762$\pm$0.032&+0.029&\textbf{0.771$\pm$0.029} \\
&Prognosis (20)&0.616$\pm$0.014&0.648$\pm$0.017&+0.032&0.617$\pm$0.015&0.649$\pm$0.017&+0.032&\textbf{0.663$\pm$0.018} \\
&Average (40)&0.693$\pm$0.022&0.710$\pm$0.021&+0.017&0.684$\pm$0.020&0.710$\pm$0.022&+0.026&\textbf{0.727$\pm$0.020$^{**}$} \\
\midrule
\multirow{4}{*}{Virchow2}&Diagnosis (8)&0.799$\pm$0.064&0.813$\pm$0.065&+0.014&0.782$\pm$0.067&0.800$\pm$0.071&+0.018&\textbf{0.818$\pm$0.063} \\
&Biomarker (12)&0.753$\pm$0.028&0.753$\pm$0.034&+0.000&0.762$\pm$0.029&0.774$\pm$0.032&+0.012&\textbf{0.794$\pm$0.029} \\
&Prognosis (20)&0.620$\pm$0.012&0.651$\pm$0.015&+0.031&0.629$\pm$0.016&0.650$\pm$0.015&+0.021&\textbf{0.678$\pm$0.015} \\
&Average (40)&0.696$\pm$0.021&0.714$\pm$0.021&+0.018&0.700$\pm$0.021&0.717$\pm$0.021&+0.017&\textbf{0.741$\pm$0.020$^{***}$} \\
\bottomrule
\end{tabular}}
\end{table}

\begin{table}[htbp]
\centering
\caption{Performance comparison of different MIL methods across all 35 tasks (totally 40 cohorts) based on the CONCHV1.5\cite{ding2024multimodal}. {The results across multiple cohorts are reported as mean \ensuremath{\pm} standard error of the mean.} Statistical significance was determined using a two-sided Wilcoxon signed-rank test, where * indicates $P$ < 0.05, ** indicates $P$~<~0.01, *** indicates $P$ < 0.001.}
\label{tab:conch_v1_5_mil_comparison}
\scalebox{0.95}{%
\begin{tabular}{lcccc}
\toprule
MIL Method&Diagnosis (8 cohorts)&Biomarker (12 cohorts)&Prognosis (20 cohorts)&Average (40 cohorts) \\
\midrule
CLAM&$0.789 \pm 0.072$&$0.740 \pm 0.030$&$0.628 \pm 0.014$&$0.693 \pm 0.021$ \\
DTFD&$0.795 \pm 0.076$&$0.748 \pm 0.029$&$0.630 \pm 0.015$&$0.698 \pm 0.022$ \\
DSMIL&$0.775 \pm 0.064$&$0.748 \pm 0.034$&$0.632 \pm 0.015$&$0.696 \pm 0.020$ \\
ILRA&$0.775 \pm 0.073$&$0.646 \pm 0.044$&$0.519 \pm 0.038$&$0.607 \pm 0.031$ \\
TransMIL&$0.776 \pm 0.073$&$0.729 \pm 0.035$&$0.607 \pm 0.014$&$0.677 \pm 0.022$ \\
WIKG&$0.763 \pm 0.067$&$0.637 \pm 0.051$&$0.632 \pm 0.014$&$0.659 \pm 0.022$ \\
ABMIL&$0.795 \pm 0.068$&$0.724 \pm 0.033$&$0.618 \pm 0.015$&$0.685 \pm 0.021$ \\
nnMIL&$\mathbf{0.825} \pm \mathbf{0.059}^{***}$&$\mathbf{0.761} \pm \mathbf{0.031^{***}}$&$\mathbf{0.655} \pm \mathbf{0.018}^{***}$&$\mathbf{0.721} \pm \mathbf{0.020}^{***}$ \\
\bottomrule
\end{tabular}
}
\end{table}

\begin{table}[htbp]
\centering
\caption{Total inference time comparison across multiple MIL methods and foundation models based on the EBRAINS test set (573 WSIs). Total inference time (in seconds, \textit{s}) per whole-slide image (WSI), computed as the sum of pathology foundation model feature extraction time (Feat Extra) and slide-level MIL inference time (MIL Infer) (on an NVIDIA L40 S GPU with solid-state drive storage). Lower values indicate faster end-to-end inference.}
\label{tab:total_inference_time}
\scalebox{0.85}{
\begin{tabular}{llcccccccc}
\toprule
FM&Stage
& CLAM&DTFD&DSMIL&ILRA&TransMIL&WIKG&ABMIL&nnMIL \\
\midrule
\multirow{3}{*}{GigaPath}
& Feat Extra
& \multicolumn{8}{c}{152.4} \\
& MIL Infer
& 0.0008&0.0009&0.0024&0.0033&0.0050&0.0065&0.0006&0.0078 \\
& Total
& 152.4008&152.4009&152.4024&152.4033&152.4050&152.4065&\textbf{152.4006}&152.4078 \\
\midrule
\multirow{3}{*}{H0}
& Feat Extra
& \multicolumn{8}{c}{177.6} \\
& MIL Infer
& 0.0009&0.0006&0.0026&0.0034&0.0080&0.0055&0.0006&0.0082 \\
& Total
& 177.6009&177.6006&177.6026&177.6034&177.6080&177.6055&\textbf{177.6006}&177.6082 \\
\midrule
\multirow{3}{*}{UNI}
& Feat Extra
& \multicolumn{8}{c}{133.8} \\
& MIL Infer
& 0.0005&0.0005&0.0017&0.0026&0.0041&0.0051&0.0004&0.0033 \\
& Total
& 133.8005&133.8005&133.8017&133.8026&133.8041&133.8051&\textbf{133.8004}&133.8033 \\
\midrule
\multirow{3}{*}{Virchow2}
& Feat Extra
& \multicolumn{8}{c}{183.6} \\
& MIL Infer
& 0.0016&0.0008&0.0021&0.0040&0.0056&0.0055&0.0005&0.0145 \\
& Total
& 183.6016&183.6008&183.6021&183.6040&183.6056&183.6055&\textbf{183.6005}&183.6145 \\
\bottomrule
\end{tabular}}
\end{table}


\end{document}